\documentclass{article}

\usepackage{arxiv}

\usepackage[utf8]{inputenc} %
\usepackage[T1]{fontenc}    %
\usepackage{hyperref}       %
\usepackage{url}            %
\usepackage{booktabs}       %
\usepackage{amsfonts}       %
\usepackage{nicefrac}       %
\usepackage{microtype}      %
\usepackage{lipsum}       %
\usepackage{graphicx}
\usepackage{natbib}
\usepackage{doi}
\usepackage{enumitem}
\usepackage{wrapfig}
\usepackage{multirow}

\usepackage{graphicx}
\usepackage{subcaption}
\usepackage{amsmath}
\usepackage{cleveref}
\DeclareMathOperator*{\argmax}{argmax}

\title{Raising the Bar on the Evaluation of Out-of-Distribution Detection}

\date{}

\author{ \hspace{1mm}Jishnu Mukhoti\thanks{Preliminary work. Contact: jishnu.mukhoti@eng.ox.ac.uk}  \\
	University of Oxford \& \\
	Meta AI\\
	\And
	\hspace{1mm}Tsung-Yu Lin \\
	Meta AI\\
	\And
	\hspace{1mm}Bor-Chun Chen \\
	Meta AI\\
    \And
	\hspace{1mm}Ashish Shah \\
	Meta AI\\
	\AND
	\hspace{1mm}Philip H.S. Torr \\
	University of Oxford\\
    \And
	\hspace{1mm}Puneet K. Dokania \thanks{Primary mentors, alphabetical order.} \\
	University of Oxford \\
	\And
	\hspace{1mm} Ser-Nam Lim $^\dagger$ \\
	Meta AI
}

\renewcommand{\shorttitle}{Raising the Bar on the Evaluation of Out-of-Distribution Detection}

\begin{document}
\maketitle
\begin{abstract}
In image classification, a lot of development has happened in detecting out-of-distribution (OoD) data. However, most OoD detection methods are evaluated on a standard set of datasets, arbitrarily different from training data. There is no clear definition of what forms a ``good" OoD dataset. Furthermore, the state-of-the-art OoD detection methods already achieve near perfect results on these standard benchmarks. In this paper, we define 2 categories of OoD data using the subtly different concepts of perceptual/visual and semantic similarity to in-distribution (iD) data. We define \textit{Near OoD} samples as perceptually similar but semantically different from iD samples, and \textit{Shifted} samples as points which are visually different but semantically akin to iD data. We then propose a GAN based framework for generating OoD samples from each of these 2 categories, given an iD dataset. Through extensive experiments on MNIST, CIFAR-10/100 and ImageNet, we show that \textbf{a)} state-of-the-art OoD detection methods which perform exceedingly well on conventional benchmarks are significantly less robust to our proposed benchmark. Moreover, \textbf{b)} models performing well on our setup also perform well on conventional real-world OoD detection benchmarks and vice versa, thereby indicating that one might not even need a separate OoD set, to reliably evaluate performance in OoD detection.
\end{abstract}


\section{Introduction}
\label{sec:intro}

With the wide-spread deployment of deep learning models in real-life applications like autonomous driving \cite{filos2020can} and medical diagnosis \cite{roy2021does}, it is imperative to ensure that in addition to being accurate, such models are also able to reliably quantify their uncertainty and identify inputs which they “don’t know”. One of the major applications of such uncertainty quantification methods is the detection of inputs sampled from a distribution different from the model’s training distribution (i.e, Out-of-Distribution or OoD inputs). A lot of work has been done in this direction from the perspective of uncertainty quantification \cite{liu2020simple, van2020uncertainty, mukhoti2021deterministic, lakshminarayanan2016simple}, OoD Detection \cite{hendrycks2016baseline, lee2018simple, fort2021exploring, winkens2020contrastive, liang2017enhancing, PintoRegMixup2022}, open-set recognition \cite{liu2019large, mundt2019unified} and the like.

\begin{figure}[!t]
    \centering
    \begin{subfigure}{0.25\linewidth}
        \centering
        \includegraphics[width=\linewidth]{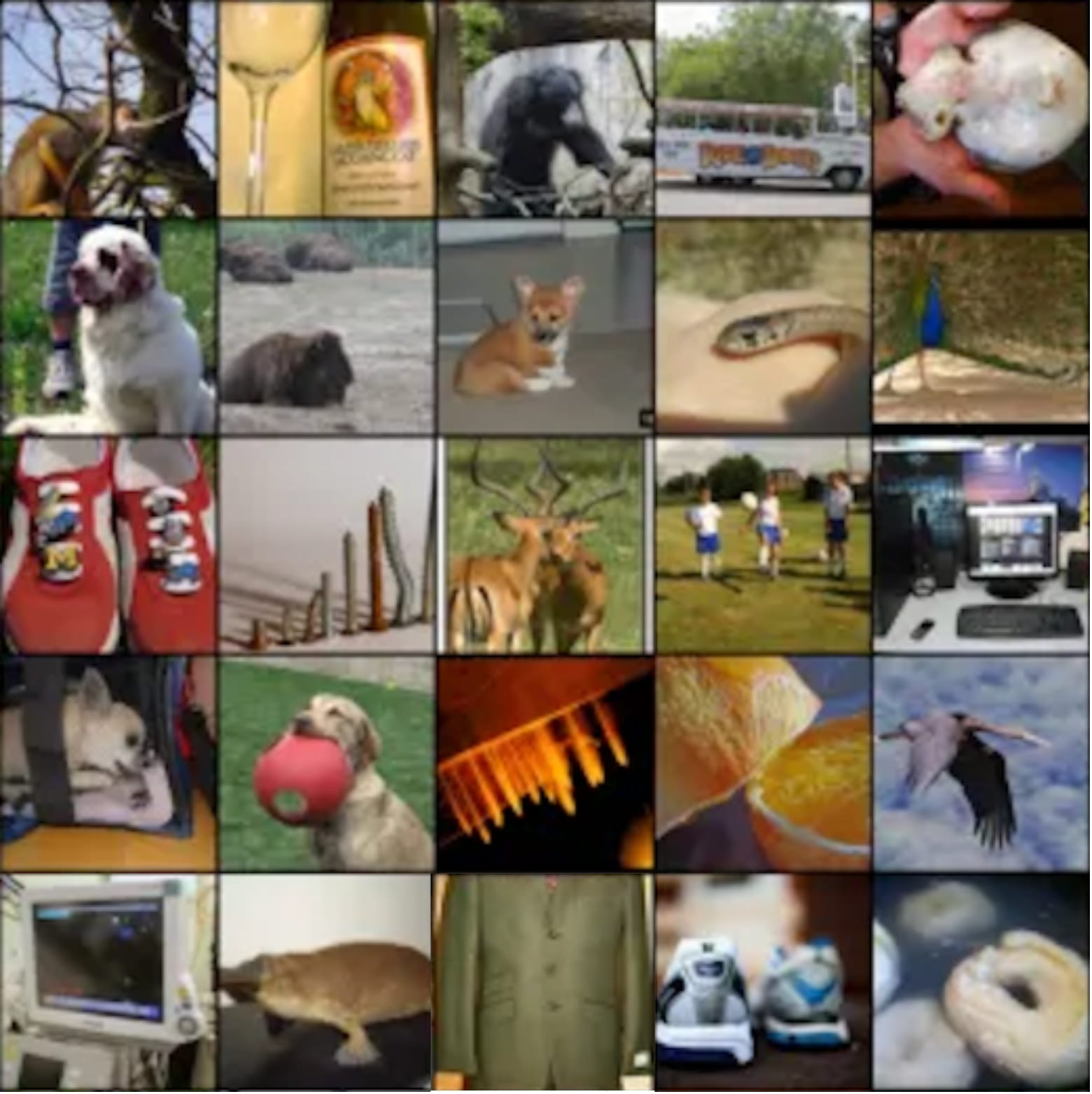}
        \caption{iD}
        \label{subfig:intro_fig_id}
    \end{subfigure}
    \begin{subfigure}{0.25\linewidth}
        \centering
        \includegraphics[width=\linewidth]{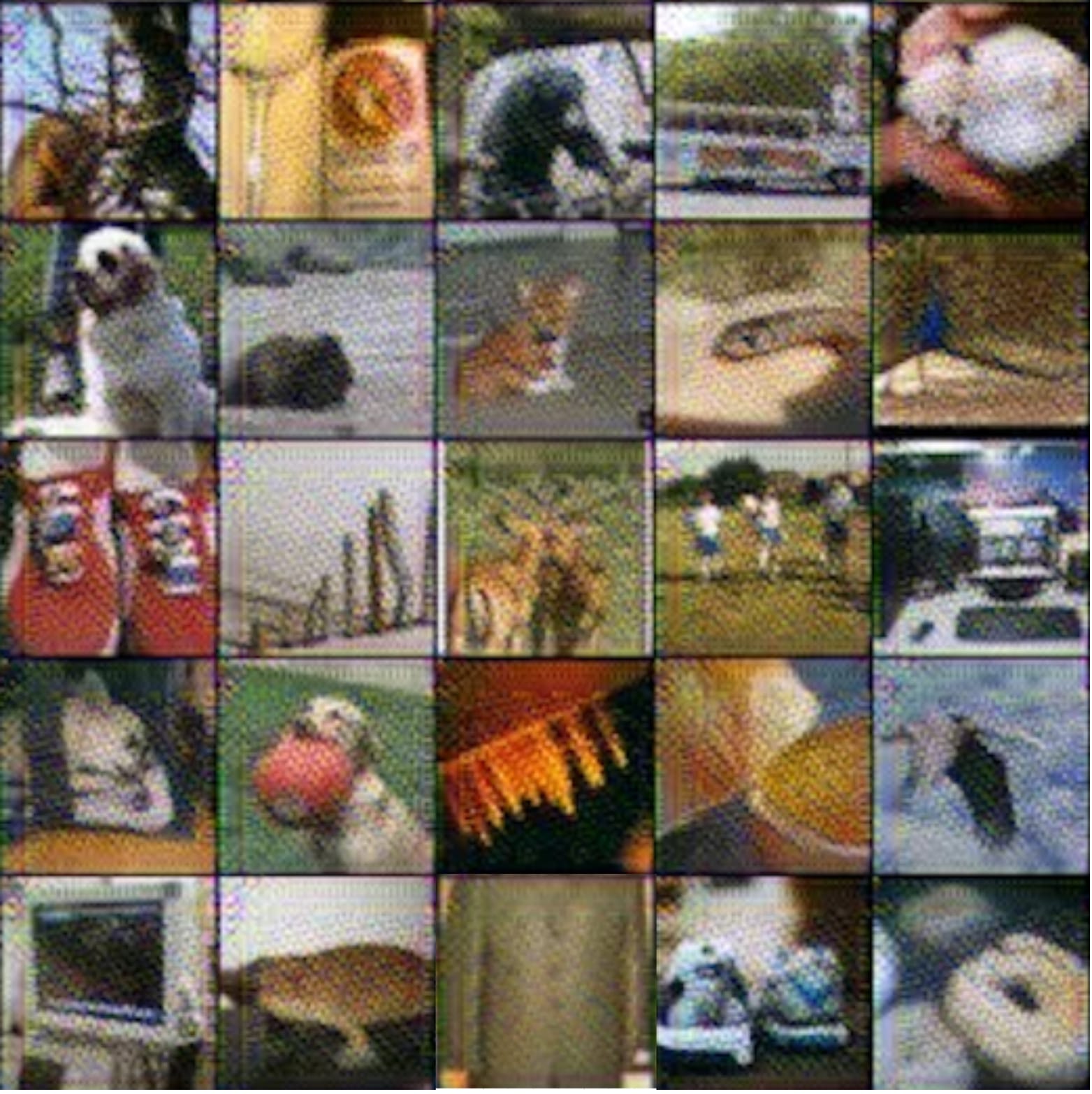}
        \caption{Shifted}
        \label{subfig:intro_fig_shifted}
    \end{subfigure}
    \begin{subfigure}{0.25\linewidth}
        \centering
        \includegraphics[width=\linewidth]{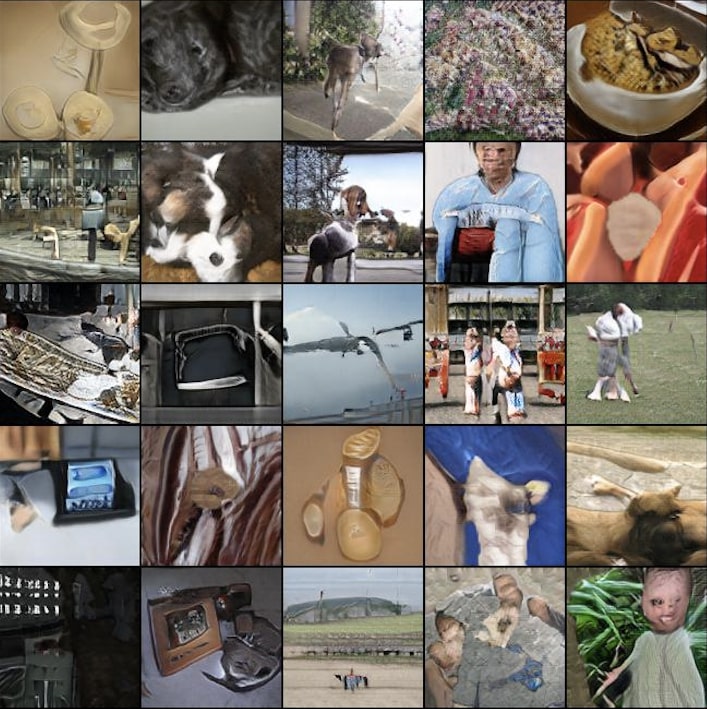}
        \caption{Near OoD}
        \label{subfig:intro_fig_morphed}
    \end{subfigure}
    \caption{\emph{Shifted} and \emph{Near OoD} samples obtained from ImageNet. \emph{Shifted samples} are visually different but semantically similar to the iD data. \emph{Near OoD images} are perceptually similar to iD data but are semantically dissimilar.
    }
    \label{fig:intro_fig}
    \vspace{-6mm}
\end{figure}

Since any point outside the training distribution can be considered OoD, the set of potential OoD inputs is infinite. This makes evaluating OoD detection a particularly challenging problem. The general evaluation practice involves using a proxy OoD dataset which is different from the training distribution (or in-distribution (iD) samples) to simulate an out-of-distribution scenario. The OoD detection algorithm is then evaluated on how well it can separate the iD samples from the OoD points. For the purposes of evaluation and benchmarking, it is then natural to ask which proxy OoD dataset is best suited for measuring model performance. To answer this question, we need to consider the different types of OoD inputs that can arise in a real-world scenario. 

In image classification, we model the conditional categorical distribution $p(y|\mathbf{x})$ over classes, given an input image $\mathbf{x}$. Under the i.i.d assumption, both the training and test images are assumed to be sampled from the same continuous distribution in image space, i.e., $p_\mathrm{train}(\mathbf{x}) = p_\mathrm{test}(\mathbf{x})$. In case of OoD samples, this assumption is broken, i.e., $p_\mathrm{train}(\mathbf{x}) \neq p_\mathrm{ood}(\mathbf{x})$. Based on the conditional distribution $p(y|\mathbf{x})$, we can then define two kinds of OoD samples.

\textbf{Distribution Shift:} Although the distribution in image space is different, the conditional distribution over class labels remains the same, i.e., $p_\mathrm{train}(y|\mathbf{x}) = p_\mathrm{ood}(y|\mathbf{x})$, and $p_\mathrm{train}(\mathbf{x}) \neq p_\mathrm{ood}(\mathbf{x})$. Such samples are generally derived from the training set by applying transformations like corruptions \cite{hendrycks2019benchmarking} and semantic shifts \cite{xiao2020noise, koh2021wilds}, where the transformed images have the same labels as the originals from the training set. For example, ImageNet-C/P \cite{hendrycks2019benchmarking} contain synthetic corruptions and perturbations applied to ImageNet \cite{deng2009imagenet}. Such datasets provide a controlled environment to study models under specific synthetic and real-world distribution shifts. 

\textbf{Unseen Categories:} The second category of OoD comprise images of classes which the model has not been trained on, i.e., $p_\mathrm{train}(y|\mathbf{x}) \neq p_\mathrm{ood}(y|\mathbf{x})$, and $p_\mathrm{train}(\mathbf{x}) \neq p_\mathrm{ood}(\mathbf{x})$. For a given training set, any dataset having a disjoint set of labels qualifies as OoD with unseen categories. How do we then decide which OoD dataset is good for evaluation? The convention is to use a well-known set of (iD vs OoD) dataset pairs like MNIST \cite{mnist} vs Fashion-MNIST \cite{xiao2017fashion}, CIFAR-10 \cite{krizhevsky2009learning} vs SVHN \cite{netzer2011reading} etc. However, \textbf{firstly,} the choice of these dataset pairs is relatively arbitrary and there is no guarantee that performance on these benchmarks will generalise to the real-world. \textbf{Secondly}, in recent literature \cite{fort2021exploring, winkens2020contrastive}, the terms ``Near OoD" and ``Far OoD" have been used to indicate the difficulty of an OoD detection task with Near OoD datasets (CIFAR-10 vs CIFAR-100) being more difficult than Far OoD (CIFAR-10 vs SVHN). With no model-agnostic metric quantifying the ``nearness" of an OoD dataset, these terms are also not well-defined. \textbf{Finally,} the current state-of-the-art OoD detection baseline, Vision Transformer \cite{fort2021exploring}, obtains around $96\%$ AUROC on CIFAR-100 vs CIFAR-10 and over $99.5\%$ AUROC on CIFAR-10 vs SVHN. Hence, the most popular OoD detection benchmarks are saturated and might give us the impression that state-of-the-art baselines are robust to OoD. The near perfect AUROC scores also indicate that these benchmarks might be rendered redundant in future for evaluating the performance of even better OoD detection methods which outperform the Vision Transformer \cite{fort2021exploring}.

In this work, we thus aim to take a step towards improving the conventional evaluation process for OoD detection in image classification. We first look at the two types of OoD mentioned above through the lens of \emph{perceptual/visual similarity} and \emph{semantic similarity} \cite{deselaers2011visual, brust2018not} between images. Perceptual similarity between two images denotes how visually similar they are and semantic similarity captures the similarity of concepts that they represent. With this in mind, we define:

\vspace{-1mm}
\begin{enumerate}[leftmargin=*]
    \itemsep0em
    \item \emph{\textbf{Shifted sets} as perceptually dissimilar but semantically similar to the training distribution}.
    \item \emph{\textbf{Near OoD sets} as perceptually similar but semantically dissimilar to the training distribution}.
    \item \emph{\textbf{Far OoD sets} as both perceptually and semantically dissimilar to the training distribution}.
\end{enumerate}
\vspace{-1mm}

\begin{figure}[!t]
    \centering
    \includegraphics[width=8cm]{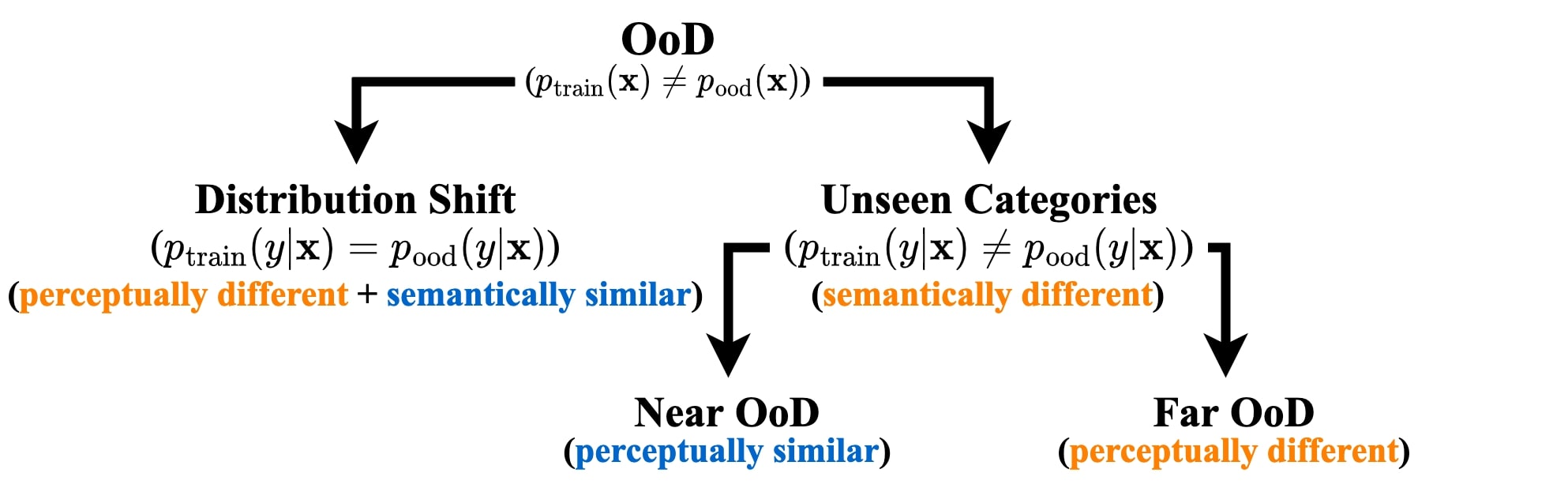}
    \vspace{-2mm}
    \caption{Categories of OoD samples}
    \label{fig:ood_categories}
    \vspace{-6mm}
\end{figure}

Clearly, images which are both perceptually and semantically similar to the training distribution would be iD. We show the hierarchy of OoD samples in \cref{fig:ood_categories}. In this work, we particularly focus on generating Shifted and Near OoD sets. Given the training set, it is difficult to define a single distance measure in the image space which can capture both perceptual and semantic similarity. Hence, we propose using a sampling based generative model, a Generative Adversarial Network (GAN) \cite{goodfellow2014generative} and design regularisers for the GAN objective using the definitions above to generate OoD samples.

For a training set $\mathcal{D} = (\mathbf{x}_i, y_i)_{i=1}^N$, where $y_i \in \mathcal{S}$, $\forall i$, in order to generate \emph{shifted samples}, we learn a transformation $t_{\mathrm{shift}}: \mathbf{x} \rightarrow \hat{\mathbf{x}}$ in the image space, $\mathbf{x}, \hat{\mathbf{x}} \in \mathbb{R}^{H, W, C}$, such that $\mathbf{x}$ and $\hat{\mathbf{x}}$ are perceptually different and semantically similar, i.e., have the same label: $\argmax_c p(y_c|\mathbf{x}) = \argmax_c p(y_c|\hat{\mathbf{x}})$. This is an Image-to-Image translation problem and hence, we use a Pix-2-Pix \cite{isola2017image} model to learn a distribution shift. In case of \emph{Near OoD}, we want to learn a distribution in the close perceptual vicinity of $\mathcal{D}$. This can be seen as a transformation $t_{\text{near ood}}: \mathbf{z} \rightarrow \hat{\mathbf{x}}$, $\mathbf{z} \sim \mathcal{N}(0, \mathbf{I})$ where the generated image $\hat{\mathbf{x}}$ is perceptually similar to the training distribution but does not belong to any of the iD classes: $\argmax_c p(y_c|\hat{\mathbf{x}}) \notin \mathcal{S}$. This is an Image generation problem and hence, we use a GAN for generating Near OoD samples. 
In \cref{fig:intro_fig}, we show examples from ImageNet of shifted as well as Near OoD samples. Through extensive experiments using several OoD detection baselines comparing our benchmarks with conventional ones, we make the following observations and contributions.

\textbf{Firstly,} the performance of state-of-the-art OoD detection baselines (Deep Ensembles \cite{lakshminarayanan2016simple} and Vision Transformers \cite{fort2021exploring}), established to be relatively robust on standard benchmarks, is consistently worse on our proposed benchmarks for both distribution shift and unseen category scenarios. This is true across datasets of all sizes: ImageNet, CIFAR-10/100 and MNIST, thereby showing that \emph{there's still plenty of room for improvement in OoD detection research}. \textbf{Secondly,} we observe a consistent trend where models which perform better on our benchmarks also perform well on standard real-world benchmarks and vice versa. Assuming that standard benchmarks are indicative of real-world OoD detection performance, the fact that our benchmarks have been created without the use of any external OoD dataset then indicates that \emph{one might not need an OoD dataset to measure OoD detection performance}. \textbf{Finally,} we propose a novel way to generate benchmarks for the evaluation of any OoD detection method.

\section{Related Work}
\label{sec:related_work}

\textbf{State-of-the-art on OoD Detection:} As mentioned in \Cref{sec:intro}, the task of uncertainty quantification naturally serves as a solution for OoD detection as OoD inputs should intuitively be assigned higher uncertainty. From this perspective, a lot of work has been done to model scalable deep neural networks to quantify their uncertainty. A popular thread of work in this regard uses the softmax distribution from a neural network to capture uncertainty. This starts off from \cite{hendrycks2016baseline} where the authors simply use the softmax probability as uncertainty and continues on to several methods including augmentations \cite{liang2017enhancing, hsu2020generalized, lee2017training},  calibration \cite{guo2017calibration, mukhoti2020calibrating,JoyAdaTemp2022} and energy based models \cite{liu2020energy}. The advantage of these methods lies in the fact that they can quantify uncertainty using a single deterministic forward pass. However, the softmax distribution often fails to capture epistemic uncertainty \cite{kendall2017uncertainties} and is overconfident on incorrect predictions for OoD inputs \cite{Gal2016Uncertainty}. A more principled approach of quantifying uncertainty uses the Bayesian formalism \cite{neal2012bayesian} and is applied to deep neural nets using approximate Bayesian inference methods \cite{gal2016dropout, blundell2015weight, mandt2017stochastic}. Yet another popular uncertainty quantification method is deep ensembles \cite{lakshminarayanan2016simple} which uses an ensemble of neural networks and averages the softmax distributions to compute uncertainty. Deep ensembles and its modifications \cite{wen2020batchensemble} have been widely accepted as one of the the state-of-the-art methods for uncertainty quantification. However, both approximate Bayesian inference and deep ensembles require either multiple forward passes at test time or multiple models to make predictions, thereby adding significant computational overhead. Attempting to achieve ensemble level performance from a single deterministic model, DUQ \cite{van2020uncertainty} and SNGP \cite{liu2020simple, liu2022simple} develop distance-aware deterministic models which can quantify uncertainty. These methods perform competitively with deep ensembles without the additional computational overhead. More recently, RegMixup~\cite{PintoRegMixup2022} showed that simply using the widely known mixup loss~\cite{zhang2018mixup} as a regularizer not only provides state-of-the-art OoD performance but also significantly improves the accuracy of a model. Finally, \cite{fort2021exploring} show that pre-trained Vision Transformers, when fine-tuned on a downstream dataset, achieve state-of-the-art AUROC scores on various conventional OoD benchmarks.

\textbf{Conventional OoD evaluation:} OoD samples are generally one of two types: i) \emph{distribution shifted} samples and ii) samples which belong to an \emph{unseen category} which the model hasn't been trained on. For evaluation, the general practice is to use separate OoD datasets for testing. For shifted samples, some of the well-known datasets include ImageNet-C (corrupted) and ImageNet-P (perturbed) \cite{hendrycks2019benchmarking} which use synthetic corruptions and perturbations as well as stylised versions of ImageNet like ImageNet-R \cite{hendrycks2020many}, ImageNet-Sketch \cite{wang2019learning} etc. There are also datasets containing specific real-world shifts like WILDS \cite{koh2021wilds}, Backgrounds \cite{xiao2020noise}, colored MNIST \cite{gulrajani2020search} etc. As shifted datasets retain the label information of the original dataset, models are evaluated on their calibration error \cite{ovadia2019can}, which compares the model confidence with its accuracy on the provided test set. On the other hand, to test a model's performance on unseen categories, a separate OoD dataset is normally used. In the current literature, MNIST \cite{mnist} vs Fashion-MNIST \cite{xiao2017fashion}, CIFAR-10/100 vs SVHN \& CIFAR-100/10 and ImageNet \cite{deng2009imagenet} vs ImageNet-O \cite{hendrycks2021natural} are the standard benchmarks for unseen distributions. Recently, \cite{hendrycks2019scaling} released the Species dataset as an OoD dataset for ImageNet-21K. However, the choice of these datasets is relatively arbitrary and a lot of the current OoD benchmarks including CIFAR-10 vs SVHN/CIFAR-100 and MNIST vs Fashion-MNIST are saturated \cite{fort2021exploring}. In this work, we thus show that generating OoD samples given a training set can produce significantly more challenging benchmarks for even the state-of-the-art OoD detection methods.

\textbf{GAN for OoD:} GANs have been previously used to generate samples on the boundary of image classes \cite{lee2017training, dionelis2021omasgan, zaheer2020old} as well as train the discriminator for anomaly detection \cite{wang2018anomaly, ngo2019fence}. The purpose is to either use the GAN itself for anomaly detection or use generated samples during training to improve the performance for OoD detection \cite{kong2021opengan, chen2021adversarial}. As mentioned before, with the lack of a clear definition of distance in image space, it is difficult to encode different types of OoD in a GAN and this is where one of our primary contributions lies. Secondly, our motivation is also orthogonal to these works. We use a GAN to improve the evaluation of OoD detection methods rather than improve the methods themselves.

\section{Method}
\label{sec:method}

In this section, we formalise our approach to generate shifted and near OoD samples given a training set. First, we encode perceptual and semantic similarity as quantifiable loss functions for generative models. Then we discuss the GAN architectures and objective functions to generate shifted and Near OoD samples respectively.

\textbf{Perceptual similarity as a loss function} As mentioned in \cref{sec:intro}, perceptual similarity between images represents how visually similar they are. Since we have target images from the training set, we can use a Full-Reference Image Quality Assessment (FR-IQA) \cite{8063957} metric to encode perceptual loss. There exist several FR-IQA metrics in the literature like SSIM \cite{wang2004image}, FSIM \cite{zhang2011fsim} and LPIPS \cite{zhang2018unreasonable} but we use the \emph{Learned Perceptual Image Patch Similarity} (LPIPS) \cite{zhang2018unreasonable} as it is known to correlate with human judgement well. Let $f_\theta$ represent a pre-trained convolutional network. Given two images, $\mathbf{x}_1$ and $\mathbf{x}_2$,  the LPIPS computes the cosine distance between feature space activations of $\mathbf{x}_1$ and $\mathbf{x}_2$ across different layers of the network $f_\theta$ as shown below:
\begin{equation}
    \mathcal{L}_{\mathrm{LPIPS}}(\mathbf{x}_1, \mathbf{x}_2) = \sum_{l} \left( \frac{1}{H_l W_l} ||f_{\theta}^l(\mathbf{x}_1) - f_{\theta}^l(\mathbf{x}_2)||_2^2 \right)
\end{equation}
where $f_{\theta}^l(\mathbf{x}_1)$ and $f_{\theta}^l(\mathbf{x}_2) \in \mathbb{R}^{H_l, W_l, C_l}$ are the feature space representations from inputs $\mathbf{x}_1$ and $\mathbf{x}_2$ in layer $l$ of the network. In this work, we use LPIPS with a VGG, although other architectures can be used as well.
From here on out, we represent perceptual loss as $\mathcal{L}_{\mathrm{LPIPS}}$. Minimizing $\mathcal{L}_{\mathrm{LPIPS}}(\mathbf{x}_1, \mathbf{x}_2)$ encourages images $\mathbf{x}_1$ and $\mathbf{x}_2$ to be perceptually similar and vice versa.

\textbf{Semantic similarity as a loss function} In image classification, the semantic meaning of an image is encoded in its class label. To identify if an image is semantically similar to the training distribution, we need a model which understands the semantic meaning of the training distribution. However, a single classifier can make incorrect and confident predictions on inputs \cite{guo2017calibration}, especially when they are not from any of the training classes \cite{ovadia2019can}. In this work, we take inspiration from Bayesian literature \cite{neal2012bayesian} and quantify semantic similarity, using the \emph{mutual information} (MI) $\mathbb{I}[y, \theta|\mathcal{D}, \mathbf{x}]$, (also known as \emph{information gain}) \cite{Gal2016Uncertainty} between the posterior distribution over parameters of a Bayesian model $p(\theta|\mathcal{D})$ and its predicted distribution over classes $p(y|\mathbf{x}, \theta)$. Let $(\mathbf{x}_g, y_g)$ be an input and $\mathcal{S}$ be the set of training classes. If $y_g \in \mathcal{S}$, seeing the sample $(\mathbf{x}_g, y_g)$ won't cause much information gain about the posterior $p(\theta|\mathcal{D})$. On the other hand, $y_g \notin \mathcal{S}$ will cause a high information gain about the posterior and  $\mathbb{I}[y, \theta|\mathcal{D}, \mathbf{x}]$ will be high. Thus, in order to generate semantically similar/dissimilar images, we want $\mathbb{I}[y, \theta|\mathcal{D}, \mathbf{x}]$ to be low/high. 

Due to the computational intractability of Bayesian inference in deep learning, we use a pre-trained deep ensemble which can be seen as a way to perform Bayesian Model Averaging \cite{wilson2020bayesian} with each model in the ensemble being a sample from the posterior. Following \cite{Gal2016Uncertainty}, we approximate MI as the difference between the entropy of the average softmax distribution of the ensemble and the average entropy of the softmax distributions of each individual network in the ensemble. Let $p(y|\mathbf{x}, \theta_t)$ represent the softmax distribution produced by the $t^{th}$ network in an ensemble of $T$ networks, on an input $\mathbf{x}$. The average softmax distribution for the ensemble is $p(y|\mathbf{x}, \theta) = \frac{1}{T}\sum_{t=1}^T p(y|\mathbf{x}, \theta_t)$. Then, the MI for the ensemble on input $\mathbf{x}$ can be approximated as:
\begin{equation}
\label{eq:mi}
    \mathcal{L}_\mathrm{MI}(\mathbf{x}) = \hat{\mathbb{I}}[y, \theta|\mathcal{D}, \mathbf{x}] \approx \mathbb{H}[p(y|\mathbf{x}, \theta)] - \frac{1}{T}\sum_{t=1}^T \mathbb{H}[p(y|\mathbf{x}, \theta_t)]
\end{equation}
where $\mathbb{H}[.]$ represents the entropy of a distribution. We use $\mathcal{L}_\mathrm{MI}$ of a pre-trained ensemble as the semantic loss to quantify semantic similarity of an image to the training distribution. Semantically similar images (eg., images belonging to training classes) should have low $\mathcal{L}_\mathrm{MI}$ \footnote{Interestingly, an alternative explanation of $\mathcal{L}_{\mathrm{MI}}$ can be found in uncertainty quantification literature \cite{Gal2016Uncertainty}. MI is well-known to capture \emph{epistemic uncertainty}, the type of uncertainty which occurs due to lack of data and reduces on observing more data. Hence, MI is high for OoD inputs with previously unseen categories. It is also used in active learning (the BALD metric) \cite{gal2017deep, kirsch2019batchbald} as an acquisition function to obtain informative samples from the pool set. Here, we maximise MI of a pretrained ensemble as an objective function to generate OoD samples from unseen categories. i.e., semantically dissimilar samples with high epistemic uncertainty.} and vice versa.

\begin{figure}[!t]
  \begin{center}
    \includegraphics[width=0.5\linewidth]{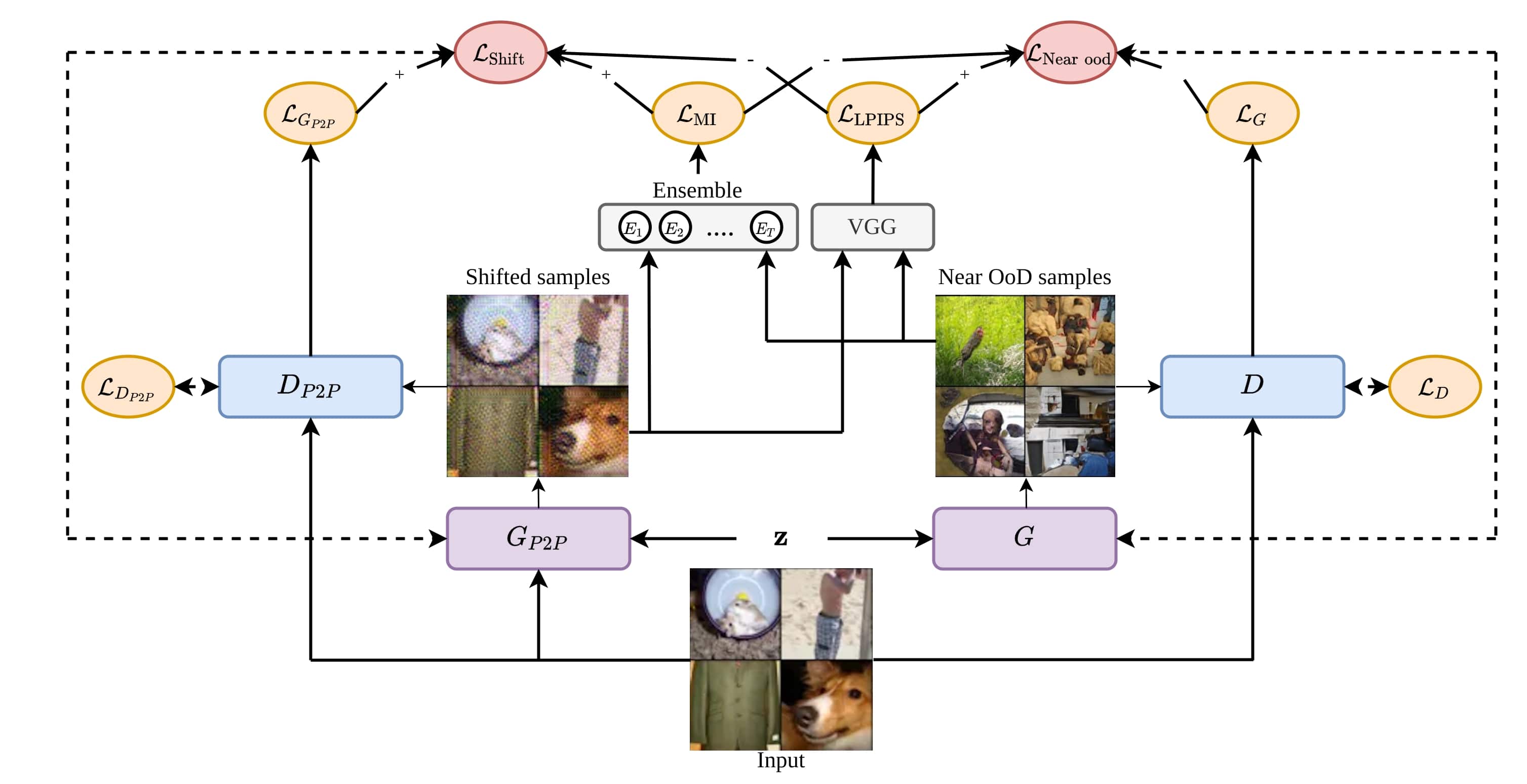}
  \end{center}
  \vspace{-2mm}
  \caption{Schematic of the proposed method to generate Shifted and Near OoD samples. $G_{P2P}$ and $D_{P2P}$ represent the Pix-2-Pix to generate shifted samples and $G$ and $D$ represent the GAN to generate Near OoD images. The dotted lines show paths for gradient propagation.}
  \label{fig:schematic_full}
  \vspace{-4mm}  
\end{figure}

\textbf{Generative Model} Having defined perceptual and semantic similarity as quantifiable loss functions, we discuss two different GAN architectures for the two different types of OoD data. For distribution shift, we intend to learn a transformation on iD data and hence, use a conditional GAN architecture, Pix-2-Pix \cite{isola2017image} in particular. For Near OoD, we intend to learn a distribution instead and hence, use a standard GAN. 
We do not change the loss for the discriminator in any of the GANs and only regularise the loss for the generator to produce the desired OoD type.

\textbf{\textit{Distribution Shift}}: For distribution shift, we want to maximize the perceptual loss $\mathcal{L}_\mathrm{LPIPS}$ (to generate perceptually different images) and minimize the semantic loss $\mathcal{L}_\mathrm{MI}$ (to preserve the semantic meaning of generated images). Thus, the regularised objective for the Pix-2-Pix generator is:
\begin{equation}
\begin{split}
\label{eq:shift_loss}
    \mathcal{L}_{\mathrm{Shift}} & = \underbrace{\mathbb{E}_{\mathbf{x}, \mathbf{z}} [\log (1-D_{P2P}(\mathbf{x}, G_{P2P}(\mathbf{x}, \mathbf{z})))]}_\text{Pix-2-Pix Generator Loss} \\
    & - \underbrace{\lambda_p \mathbb{E}_{\mathbf{x}, \mathbf{z}} [\mathcal{L}_{\mathrm{LPIPS}}(\mathbf{x}, G_{P2P}(\mathbf{x}, \mathbf{z}))]}_\text{maximize perceptual loss} \\
    & + \underbrace{\lambda_s \mathbb{E}_{\mathbf{x}, \mathbf{z}} [\mathcal{L}_{\mathrm{MI}}(G_{P2P}(\mathbf{x}, \mathbf{z}))]}_\text{minimize semantic loss},
\end{split}
\end{equation}
where $G_{P2P}$ and $D_{P2P}$ represent the generator and discriminator of the Pix-2-Pix GAN and $\lambda_p$ and $\lambda_s$ are the regularisation coefficients for the perceptual and semantic losses.

\textbf{\textit{Near OoD}}: Similarly, for Near OoD, we want to minimize the perceptual loss $\mathcal{L}_\mathrm{LPIPS}$ (to encourage perceptually similar images) and maximize the semantic loss $\mathcal{L}_\mathrm{MI}$ (to generate semantically different images which don't belong to training set classes). We use a GAN for Near OoD distributions with the generator objective as shown below:
\begin{equation}
\begin{split}
\label{eq:morph_loss}
    \mathcal{L}_{\text{Near ood}} & = \underbrace{\mathbb{E}_{\mathbf{z}} [\log (1-D(G(\mathbf{z})))]}_\text{GAN Generator Loss} \\
    & + \underbrace{\lambda_p \mathbb{E}_{\mathbf{x}, \mathbf{z}} [\mathcal{L}_{\mathrm{LPIPS}}(\mathbf{x}, G(\mathbf{z}))]}_\text{minimize perceptual loss} \\
    & - \underbrace{\lambda_s \mathbb{E}_{\mathbf{z}} [\mathcal{L}_{\mathrm{MI}}(G(\mathbf{z}))]}_\text{maximize semantic loss},
\end{split}
\end{equation}
where $G$ and $D$ denote the generator and discriminator of the GAN respectively. See \cref{fig:schematic_full} for a schematic.

\section {Experiments}
\label{sec:experiments}

\subsection{Implementation Details}
\label{sec:morphed_shifted_benchmarks}

\textbf{Setting $\lambda_p$ and $\lambda_s$ and MI thresholding}: In \cref{eq:shift_loss} and \cref{eq:morph_loss}, we introduce the regularisation coefficients $\lambda_p$ and $\lambda_s$. To set these, we generate samples using different combinations of $\lambda_s$ and $\lambda_p$. 
We then measure the MI on the training ensemble for all generated samples and filter them such that MI is neither too high (avoid samples which are too dissimilar), nor too low (samples which coincide with iD). The lower bound is selected to be the lowest value which minimizes MI overlap between val and near OoD sets and the upper bound is chosen to be lower bound + 0.4 (see \cref{fig:cifar10_mi} in appendix). In particular, we use $[0.1, 0.5]$ for MNIST, $[0.2, 0.6]$ for CIFAR-10 and $[0.4, 0.8]$ for CIFAR-100 and ImageNet. We found these settings to empirically produce good OoD samples across datasets.

\textbf{Ensemble for $\mathcal{L}_{\mathrm{MI}}$}: We implement the semantic loss $\mathcal{L}_{\mathrm{MI}}$ on MNIST using an ensemble of 4 different networks: LeNet \cite{mnist}, AlexNet \cite{krizhevsky2012imagenet}, VGG-11 \cite{simonyan2014very} and ResNet-18 \cite{he2016deep}. On CIFAR-10/100, we use 6 networks: DenseNet-121 \cite{huang2017densely}, ResNet-50/110, VGG-16, Wide-ResNet-28-10 and Inception-v3 \cite{szegedy2016rethinking} and on ImageNet, we use 3 networks: ResNet-18, MobileNet-v3-Large \cite{howard2019searching} and EfficientNet-B0 \cite{tan2019efficientnet}. All the ensemble models are trained on their respective training sets.

\textbf{Generating Near OoD Datasets} We use a GAN to generate Near OoD samples from the training set using the $\mathcal{L}_{\text{Near ood}}$ loss. In particular, for MNIST, we use DCGAN for its simplicity and on CIFAR-10/100 and ImageNet, we use BigGAN due to its superior performance in terms of FID scores.  For all the GANs, we set $\lambda_p$ to $1$ and perform a grid search for $\lambda_s$ over the interval $[0.5, 3]$ at steps of $0.25$. As mentioned above, we use all the resulting trained GANs and filter samples out by thresholding on MI for the ensemble.

\textbf{Generating Shifted Datasets}  We generate shifted samples using a Pix-2-Pix model trained on $\mathcal{L}_{\mathrm{shift}}$ loss. After a grid search over different values, we found $\lambda_s = 1$ and $\lambda_p = 2$ to produce the best results on CIFAR-10 and ImageNet. Further training details can be found in \cref{app:additional_training_details}. We show qualitative examples of both near OoD and shifted samples in \cref{fig:intro_fig} and additional samples in \cref{app:qualitative}.

\subsection{Sanity Check on Benchmarks}
\label{sec:benchmark_sanity_checks}
We perform a sanity check using the CIFAR-10 dataset to verify that the Near OoD samples are indeed semantically dissimilar and perceptually similar while the Shifted samples are semantically similar and perceptually dissimilar.

\begin{figure}[!t]
    \centering
    \begin{subfigure}{0.12\linewidth}
        \centering
        \includegraphics[width=\linewidth]{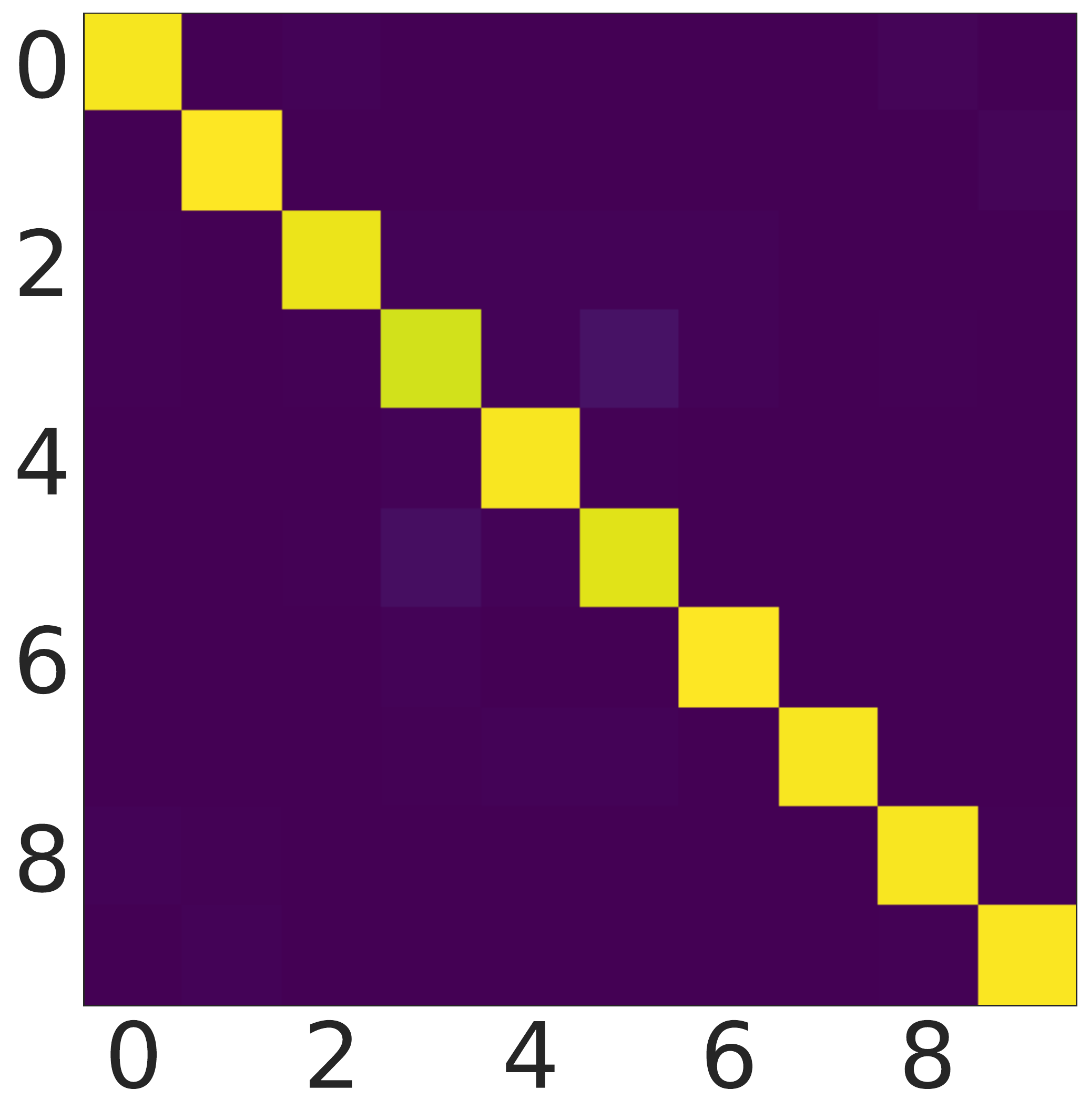}
        \caption{iD}
    \end{subfigure}
    \begin{subfigure}{0.12\linewidth}
        \centering
        \includegraphics[width=\linewidth]{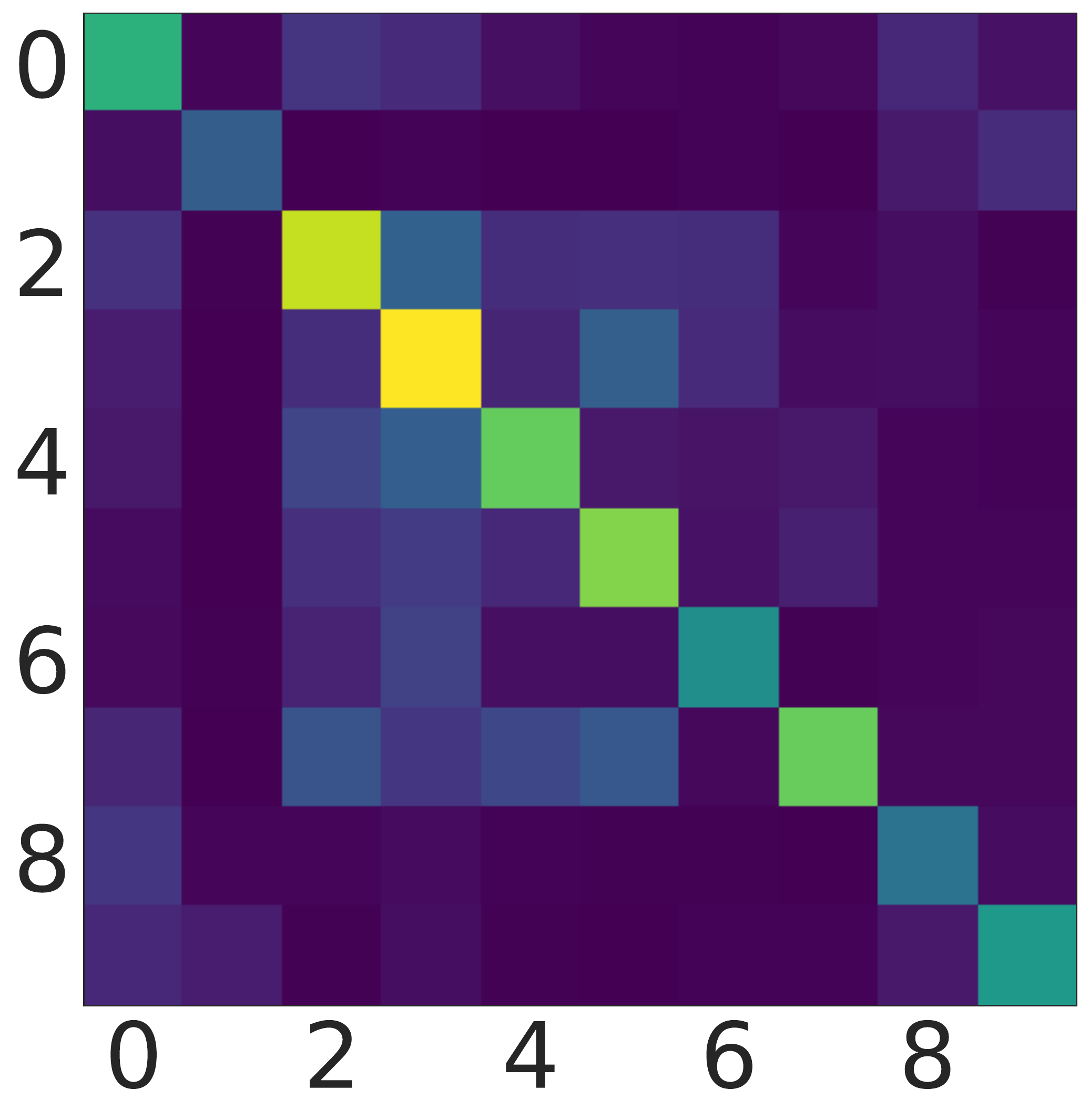}
        \caption{Shifted}
    \end{subfigure}
    \begin{subfigure}{0.12\linewidth}
        \centering
        \includegraphics[width=\linewidth]{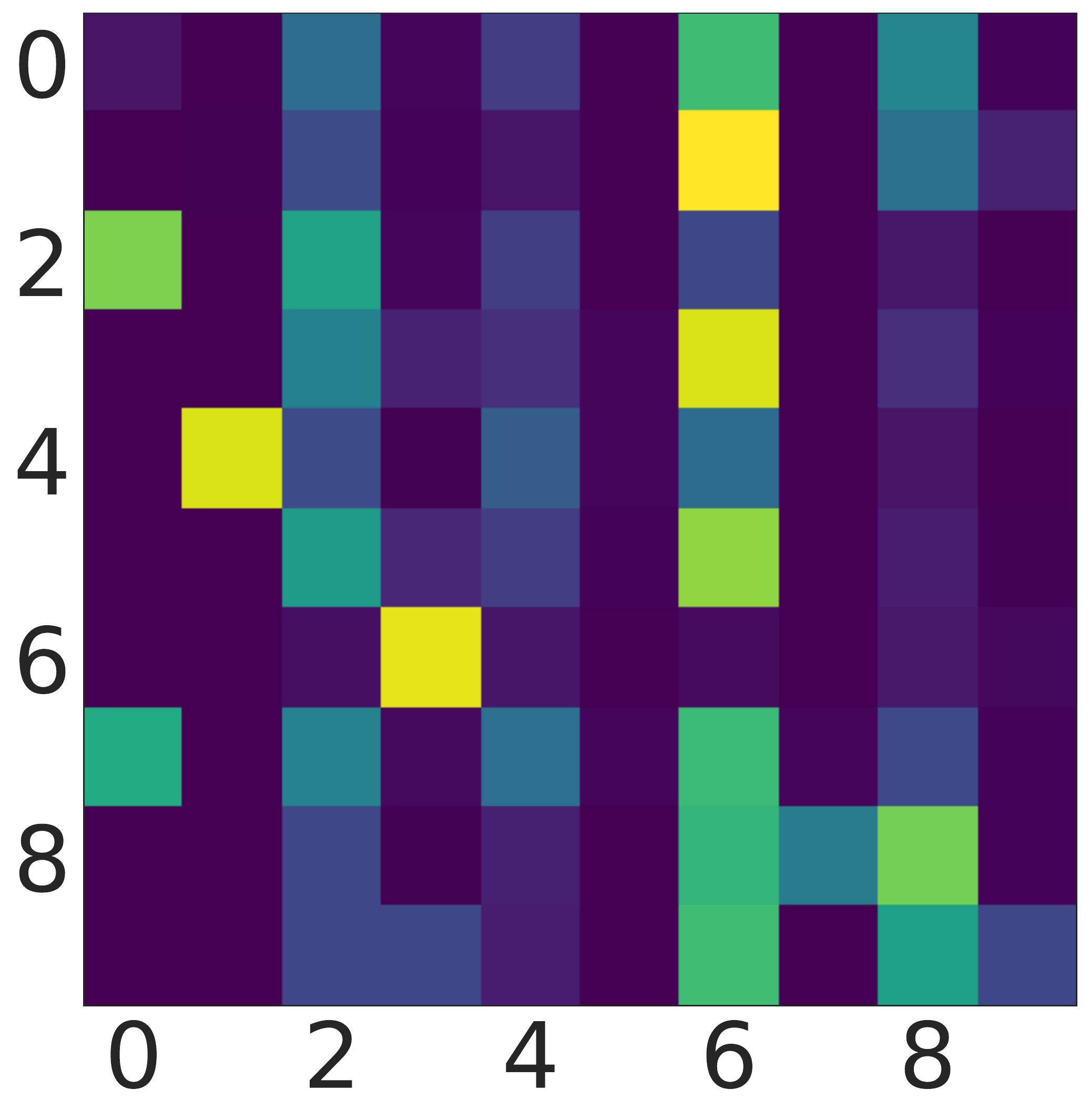}
        \caption{Near OoD}
    \end{subfigure}
    \caption{Confusion matrices for ensemble predictions on versions of CIFAR-10.}
    \vspace{-1mm}
    \label{fig:shifted_morphed_semantic_meaning}
\end{figure}

\textbf{Semantic similarity to the training distribution} To verify that Near OoD samples are semantically dissimilar and Shifted samples are semantically similar to iD, we compare the predictions of an ensemble of 6 models: DenseNet-121, ResNet-50/110, VGG-16, Wide-ResNet-28-10 and Inception-v3 on the CIFAR-10 test set with both near OoD and shifted versions of CIFAR-10. Note that the ensembles are different from the ones used in training. We present the corresponding confusion matrices in \cref{fig:shifted_morphed_semantic_meaning}. Clearly, in case of shifted samples, the label information is preserved as the ensemble's predicted classes broadly match the correct labels. However, such is not the case for near OoD samples where the predictions for each class are mostly incorrect indicating that the dataset has not preserved the label information from the training distribution.

\begin{figure}[!t]
    \centering
    \begin{subfigure}{0.35\linewidth}
        \centering
        \includegraphics[width=\linewidth]{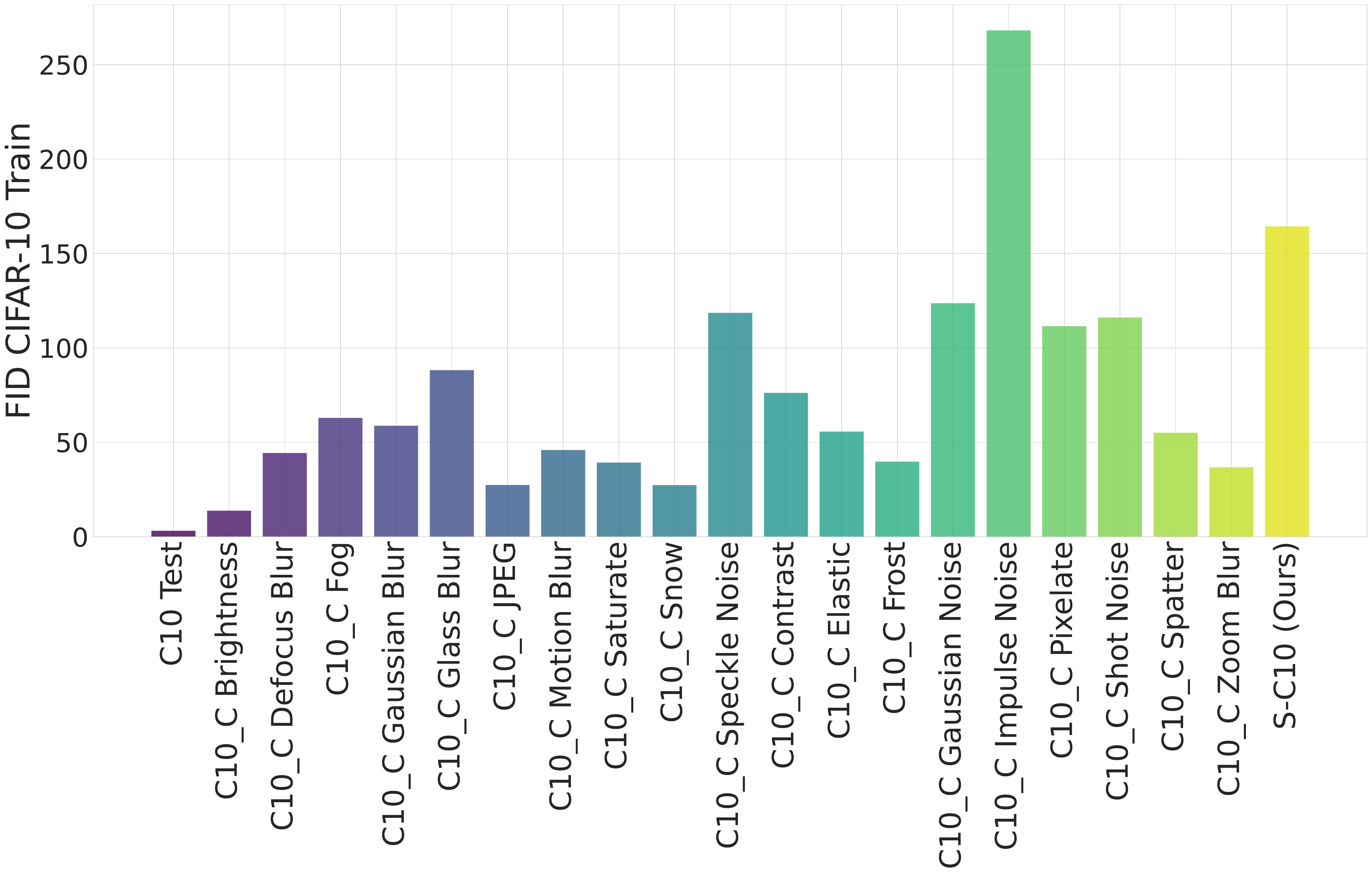}
    \end{subfigure}
    \begin{subfigure}{0.25\linewidth}
        \centering
        \includegraphics[width=\linewidth]{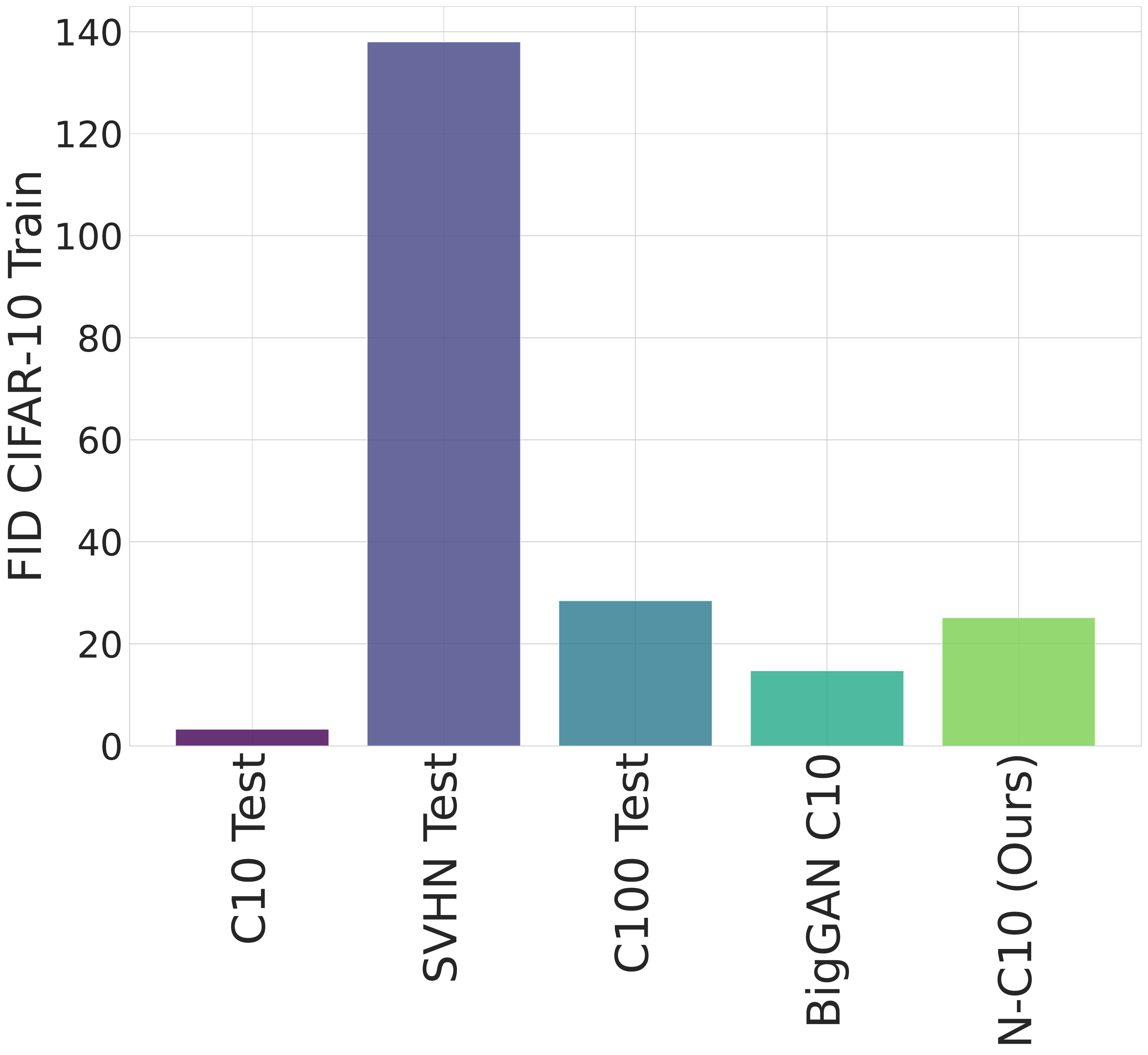}
    \end{subfigure}
    \caption{FID scores with CIFAR-10 training set on various OoD sets comparing with CIFAR-10 test set for reference.}
    \vspace{-3mm}
    \label{fig:shifted_morphed_perceptual_meaning}
\end{figure}

\textbf{Perceptual similarity to the training distribution} To measure perceptual similarity between images, we use the well-known FID score \cite{heusel2017gans}. We show the FID between the CIFAR-10 training set and Shifted (S) CIFAR-10 comparing with all the corruption types at intensity 5 of CIFAR-10-C \cite{hendrycks2019benchmarking}. Simlarly, we compare the FID of Near OoD (N) CIFAR-10 with SVHN, CIFAR-100 and samples generated by a BigGAN trained on CIFAR-10. We present the results in \cref{fig:shifted_morphed_perceptual_meaning}. It is evident that S CIFAR-10 has a very high FID score indicating perceptual dissimilarity from the training set. Note however, that it is not the highest among all the corruption types in CIFAR-10-C. On the other hand, N CIFAR-10 has a significantly lower FID than SVHN and slightly lower than CIFAR-100, providing evidence in favour of the fact that N CIFAR-10 is perceptually more similar to the training set as compared to CIFAR-100 or SVHN.

\subsection{Evaluating Near OoD Datasets}
\label{sec:exp_ood_detection}

Next, we evaluate the Near OoD samples using two experiments. Firstly, we use state-of-the-art OoD detection methods \cite{lakshminarayanan2016simple, fort2021exploring} on generated Near OoD datasets obtained from MNIST, CIFAR-10/100 and ImageNet and compare them with conventionally used OoD benchmarks. Secondly, we compare performance on models trained using outlier exposure \cite{hendrycks2018deep} on our datasets with conventional ones.

\textbf{OoD Detection Baselines} For evaluation, we use state-of-the-art OoD detection methods. First, we use softmax entropy and softmax confidence \cite{hendrycks2016baseline} as well as the Mahalanobis distance \cite{lee2018simple} for models trained on the respective training sets. Second, we use 5-member deep ensembles, with predictive entropy \cite{lakshminarayanan2016simple} as the measure of uncertainty. The predictive entropy of a deep ensemble is the entropy of the average softmax distribution (first term in \cref{eq:mi}). Note that the ensemble used to compute $\mathcal{L}_\mathrm{MI}$ during training contains models with different architectures to encourage variability in predictions. During evaluation however, the models in each ensemble have the same architecture.
Finally, we use Vision Transformers \cite{fort2021exploring} with softmax confidence, entropy and Mahalanobis as the strongest baseline. On ImageNet, we only provide results using Vision Transformers.

\begin{figure*}[!t]
    \centering
    \begin{subfigure}{0.24\linewidth}
        \centering
        \includegraphics[width=\linewidth]{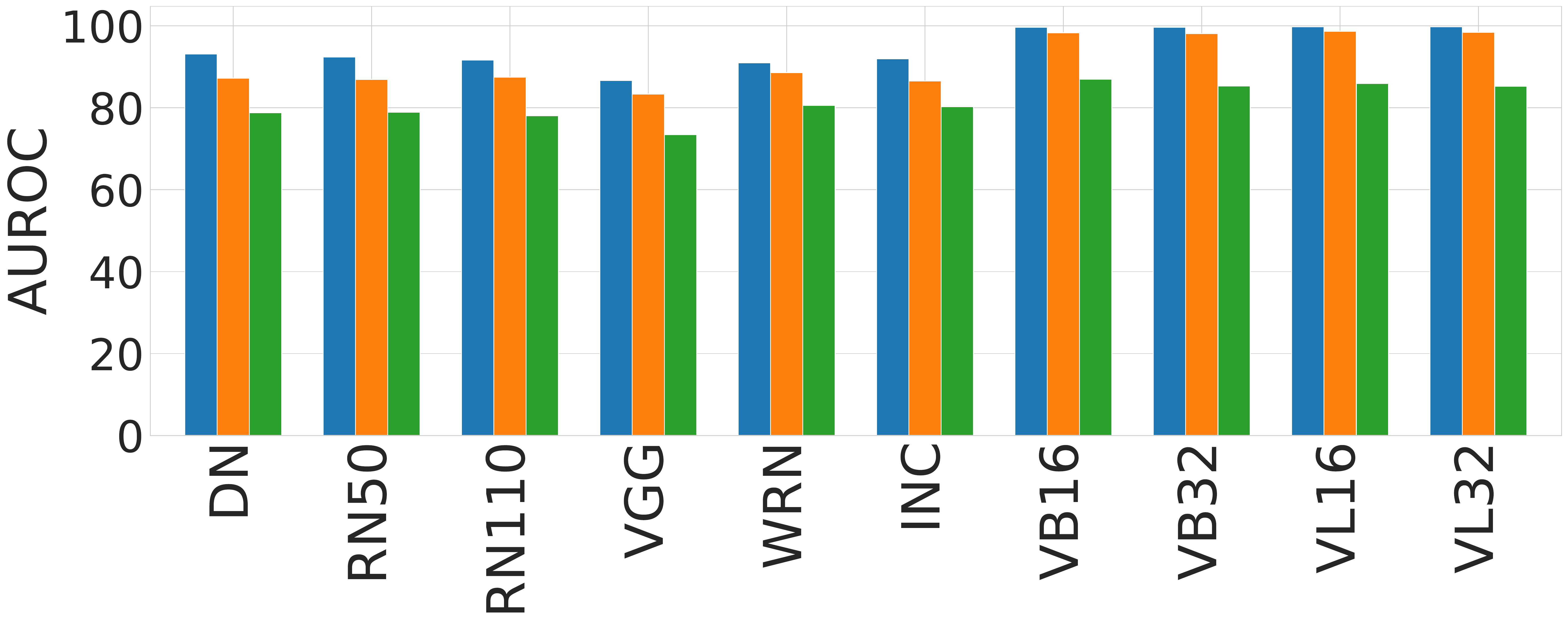}
        \vspace{-3mm}
        \label{subfig:auroc_c10_softmax_entropy}
    \end{subfigure}
    \begin{subfigure}{0.24\linewidth}
        \centering
        \includegraphics[width=\linewidth]{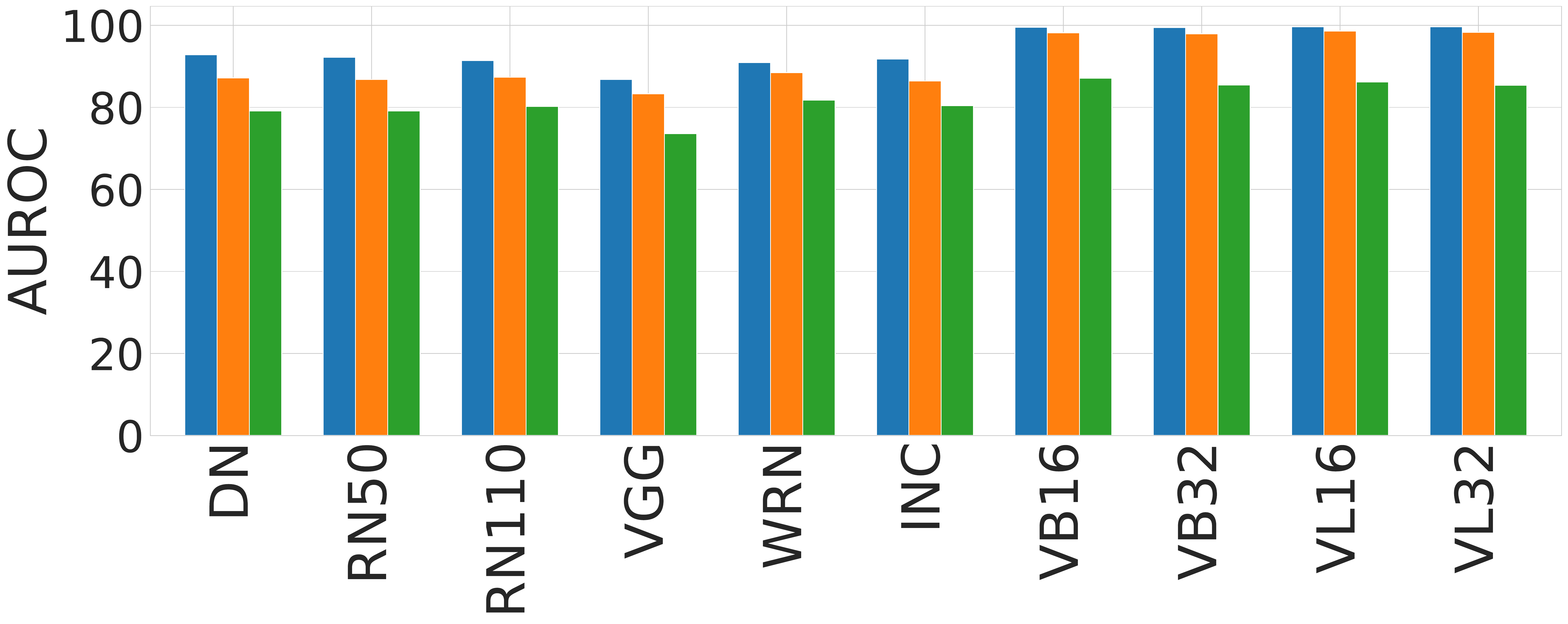}
        \vspace{-3mm}
        \label{subfig:auroc_c10_softmax_confidence}
    \end{subfigure}
    \begin{subfigure}{0.24\linewidth}
        \centering
        \includegraphics[width=\linewidth]{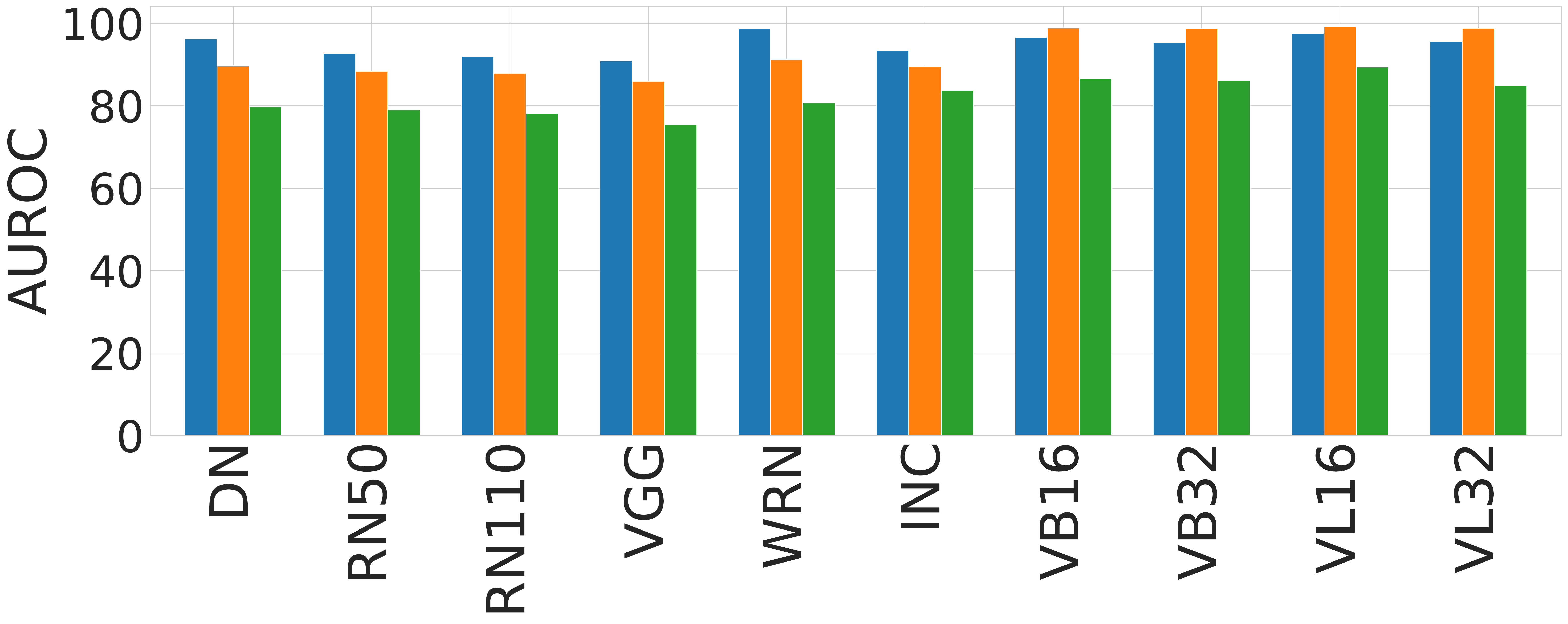}
        \vspace{-3mm}
        \label{subfig:auroc_c10_mahalanobis}
    \end{subfigure}
    \begin{subfigure}{0.24\linewidth}
        \centering
        \includegraphics[width=\linewidth]{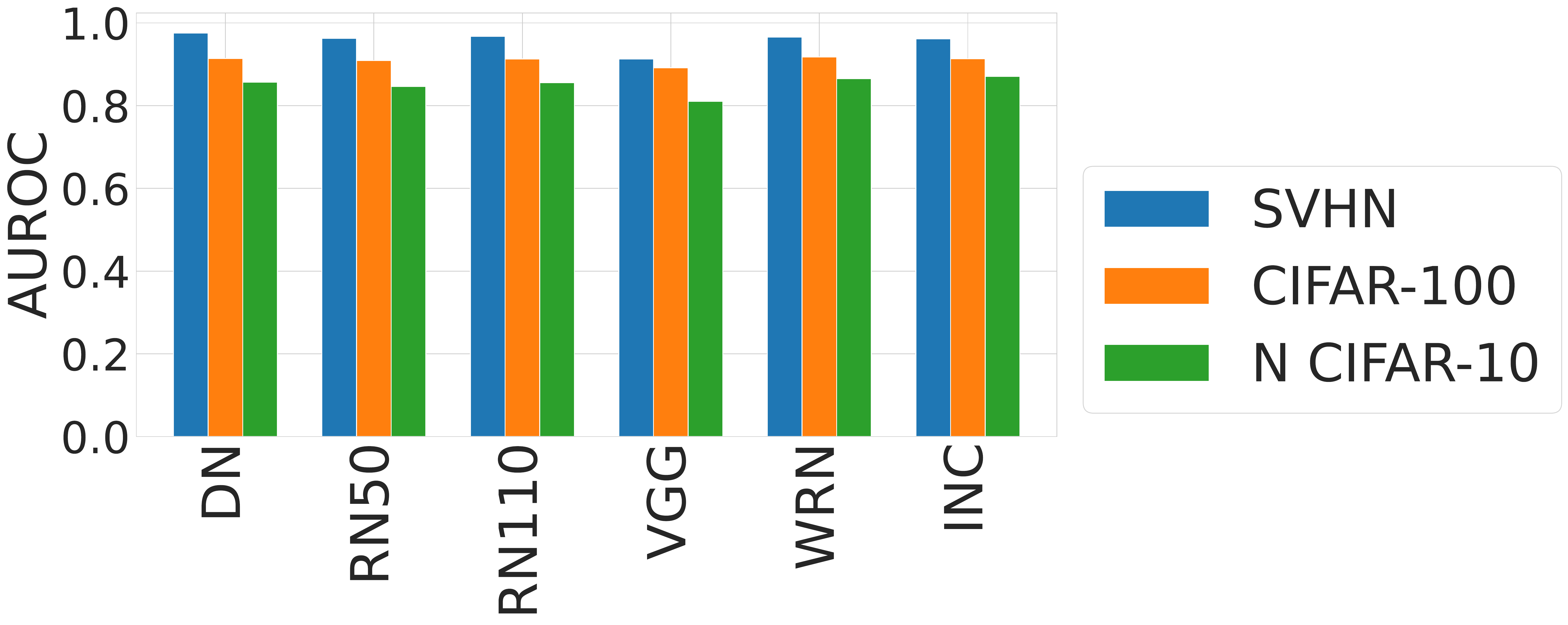}
        \vspace{-3mm}
        \label{subfig:auroc_c10_ensemble}
    \end{subfigure}
    
    \begin{subfigure}{0.24\linewidth}
        \centering
        \includegraphics[width=\linewidth]{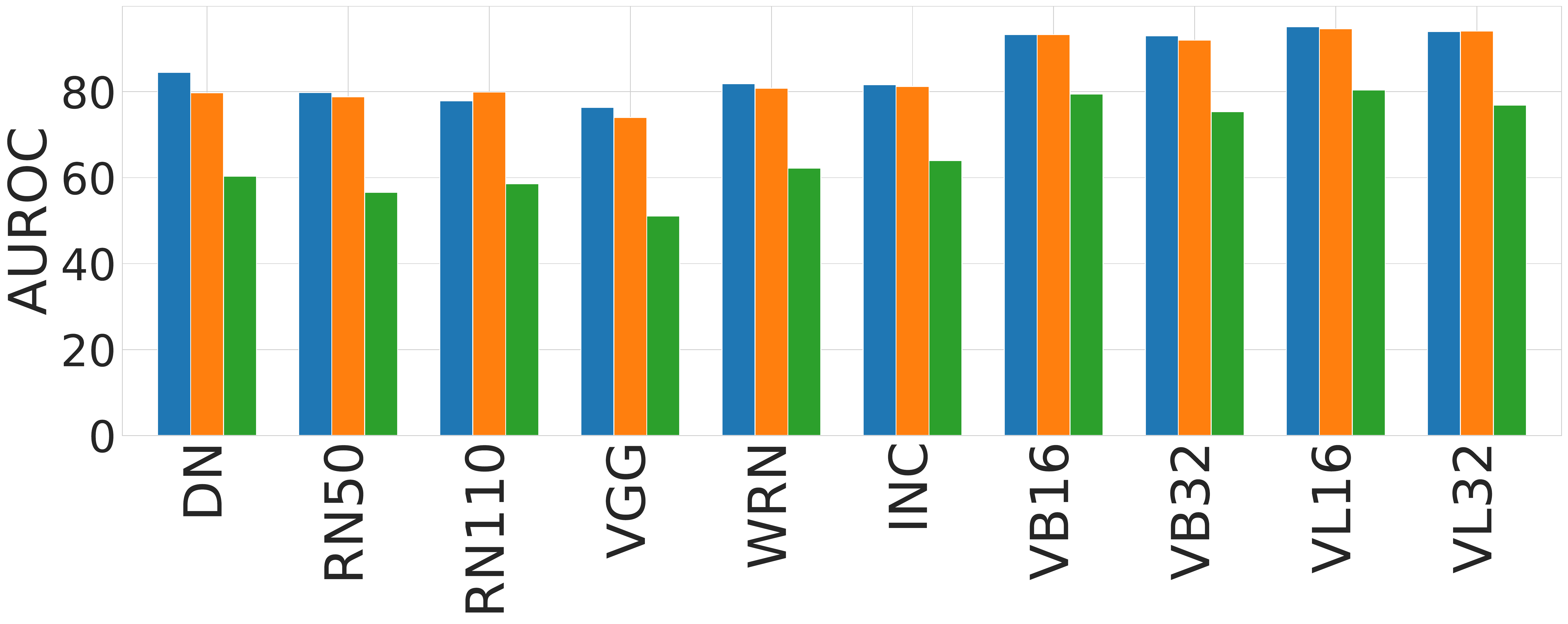}
        \caption{Entropy}
        \vspace{-2mm}
        \label{subfig:auroc_c100_softmax_entropy}
    \end{subfigure}
    \begin{subfigure}{0.24\linewidth}
        \centering
        \includegraphics[width=\linewidth]{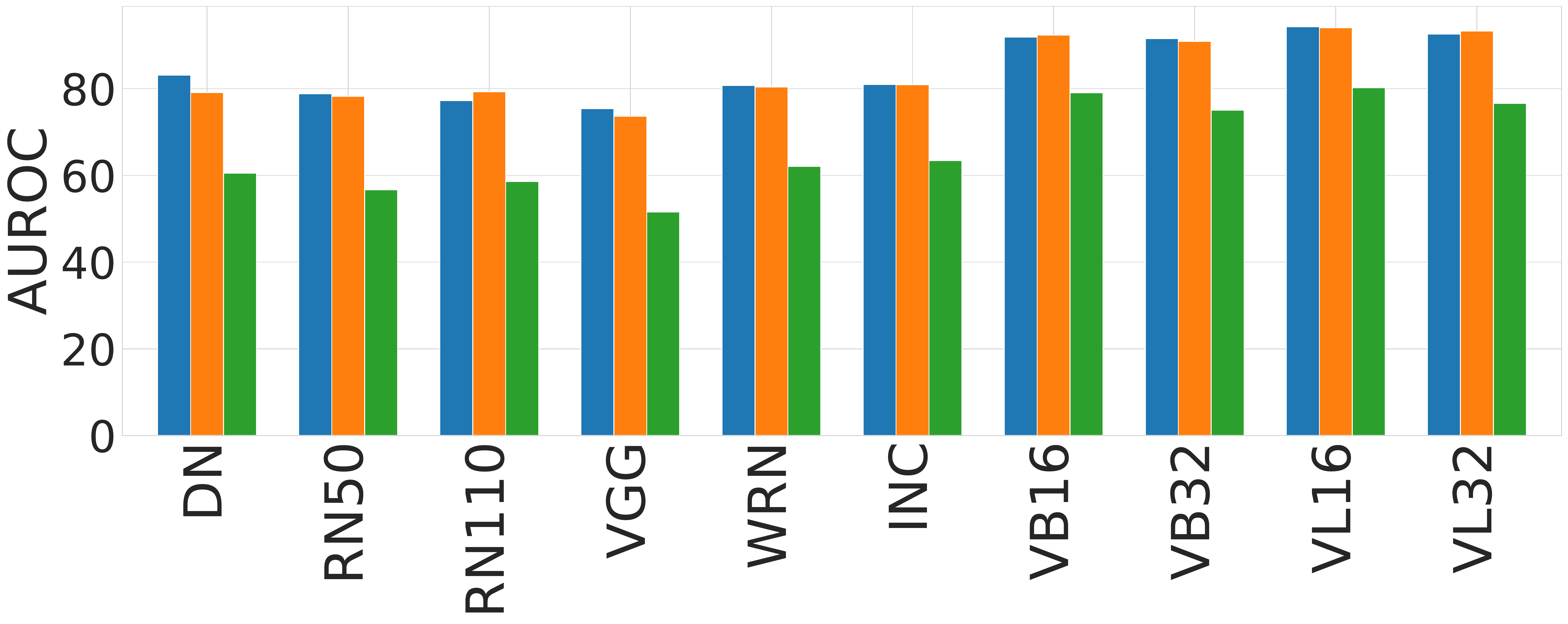}
        \caption{Confidence}
        \vspace{-2mm}
        \label{subfig:auroc_c100_softmax_confidence}
    \end{subfigure}
    \begin{subfigure}{0.24\linewidth}
        \centering
        \includegraphics[width=\linewidth]{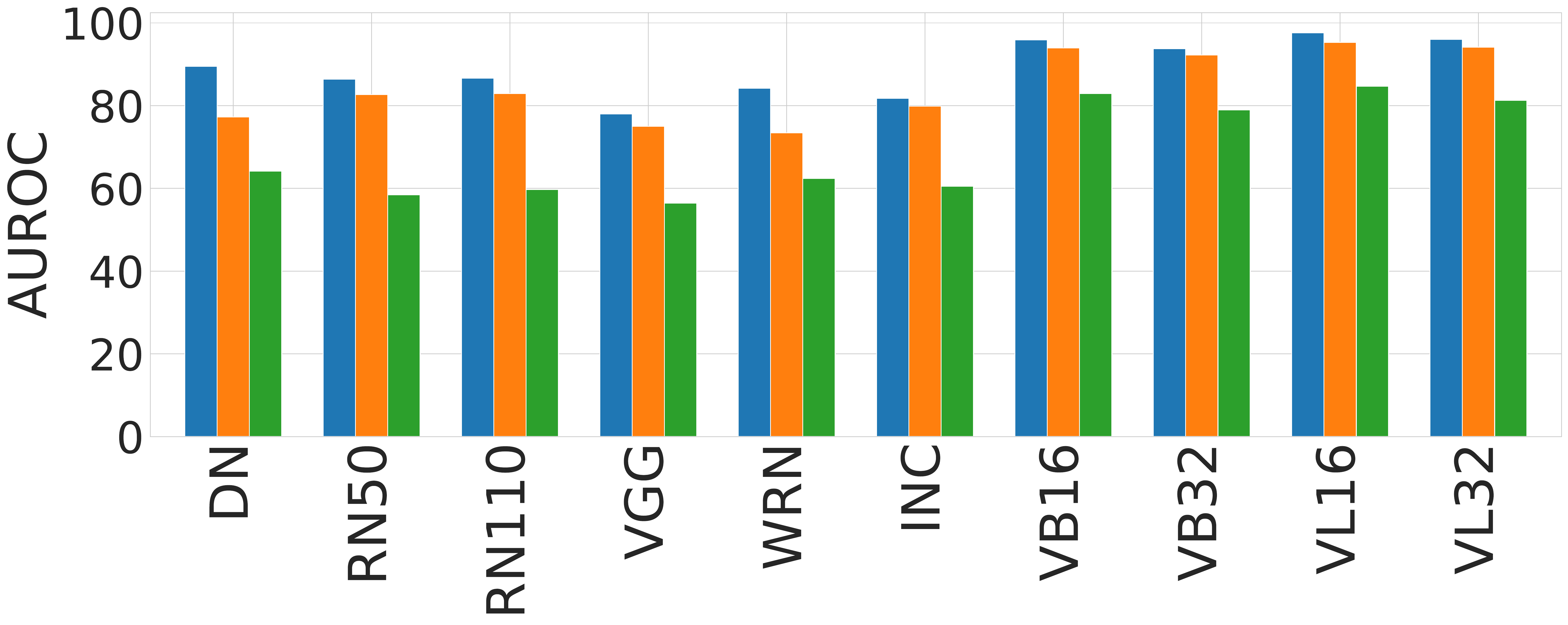}
        \caption{Mahalanobis}
        \vspace{-2mm}
        \label{subfig:auroc_c100_mahalanobis}
    \end{subfigure}
    \begin{subfigure}{0.24\linewidth}
        \centering
        \includegraphics[width=\linewidth]{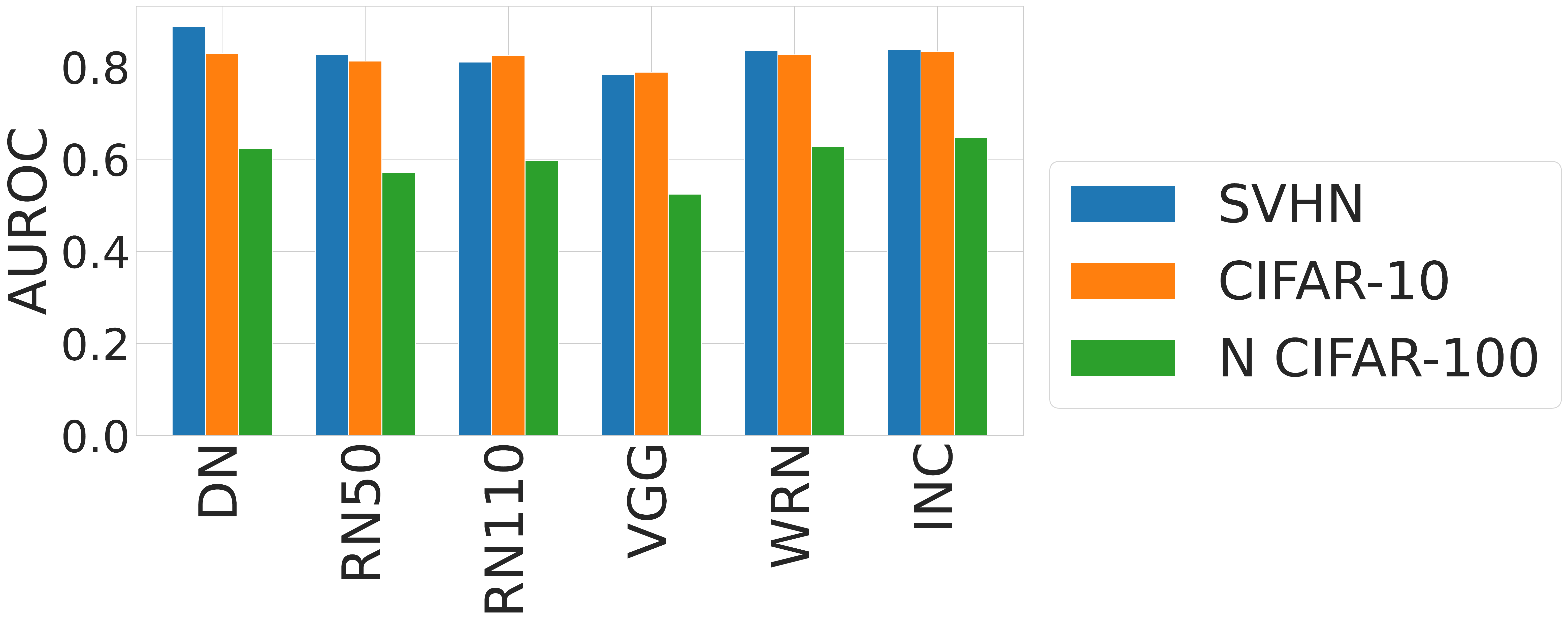}
        \caption{Ensemble}
        \vspace{-2mm}
        \label{subfig:auroc_c100_ensemble}
    \end{subfigure}
    \vspace{1mm}
    \caption{AUROC \% for different models, DenseNet-121 (DN), ResNet-50 (RN50), ResNet-110 (RN110), VGG-16, Wide-ResNet-28-10 (WRN) and Inception-v3 (INC), ViT-B-16/32 (VB16/32) and ViT-L-16/32 (VL16/32) trained on CIFAR-10 (first row) and CIFAR-100 (second row) using SVHN, CIFAR-10/100 and Near OoD (N) CIFAR-10/100 as OoD datasets.}
    \label{fig:auroc_cifar10_100}
    \vspace{-4mm}
\end{figure*}

\textbf{Competitive Benchmarks} For comparison, we use current conventional benchmarks. In particular, we compare with MNIST vs Fashion-MNIST, CIFAR-10 vs SVHN/CIFAR-100, CIFAR-100 vs SVHN/CIFAR-10 and ImageNet vs ImageNet-O \cite{hendrycks2021natural}. We present the test set accuracies and AUROC scores for MNIST in \cref{table:mnist_ood_results} of the appendix, CIFAR-10/100 and ImageNet test accuracies in \cref{table:cifar10_100_imagenet_accuracy} of the appendix, corresponding AUROC scores for CIFAR-10/100 shown as bar plots in \cref{fig:auroc_cifar10_100} and finally, AUROC scores for ImageNet models in \cref{table:auroc_imagenet}. All related AUPRC scores are shown \cref{app:additional_results}.

\begin{table}[!t]
    \centering
    \scriptsize
    \resizebox{0.5\linewidth}{!}
    {
    \begin{tabular}{ccccc}
    \toprule
    \textbf{Model} & \textbf{Im-O (SE)} & \textbf{Im-O (SC)} & \textbf{N-Im (Ours) (SE)} & \textbf{N-Im (Ours) (SC)} \\
    \midrule
    ViT-B-16 & $89.15$ & $88.13$ & $\mathbf{80.63}$ & $\mathbf{79.72}$ \\
    ViT-B-32 & $84.94$ & $82.96$ & $\mathbf{76.98}$ & $\mathbf{75.43}$ \\
    ViT-L-16 & $91.36$ & $90.69$ & $\mathbf{84.70}$ & $\mathbf{82.68}$ \\
    ViT-L-32 & $90.51$ & $88.91$ & $\mathbf{81.65}$ & $\mathbf{80.06}$ \\ 
    \bottomrule
    \end{tabular}
    }
    \vspace{1mm}
    \caption{AUROC \% of ViT on ImageNet vs ImageNet-O (Im-O) and Near OoD ImageNet (N-Im) with softmax entropy (SE) and confidence (SC).}
    \label{table:auroc_imagenet}
    \vspace{-1mm}
\end{table}

\begin{figure}[!t]
    \centering
    \includegraphics[width=0.4\linewidth]{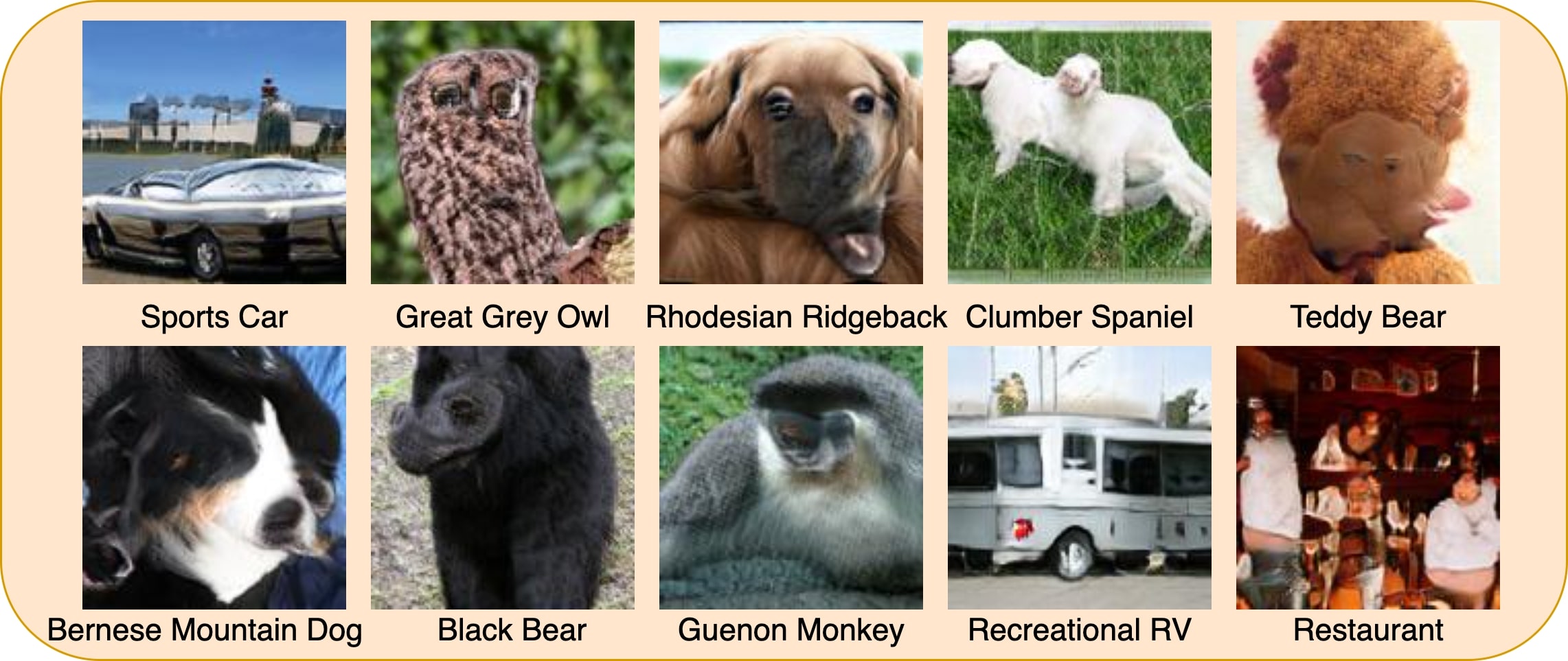}
    \vspace{-1mm}
    \caption{ImageNet Near OoD with class names}
    \label{fig:morphed_samples_with_classnames}
    \vspace{-2mm}
\end{figure}

\textbf{Observations} Our observations are as follows:
\vspace{-1mm}
\begin{enumerate}[leftmargin=*]
    \itemsep-1mm
    \item \emph{AUROC \& AUPRC for all model architectures across all training datasets and all baselines are significantly lower for our Near OoD samples as compared to conventionally used OoD datasets}. This brings into question, the performance of baselines which are considered to be robust to OoD inputs, as evidenced from their performance on standard benchmarks. It also provides evidence in favour of more challenging OoD detection benchmarks for evaluation which are far from saturation. 
    \item \emph{The order of performance is preserved.} If a model $M_1$ outperforms a model $M_2$ on our Near OoD benchmark, it broadly outperforms $M_2$ on all conventionally used real-world benchmarks and vice-versa.
\end{enumerate}
\vspace{-1mm}
\textbf{Discussion \& Visualisation} Firstly, as mentioned before, conventional OoD benchmarks might become redundant in future for the best OoD detection methods. Our Near OoD datasets are much more effective at measuring performance as all the baselines consistently underperform on our samples. Secondly, even though the Near OoD samples might not look like real-world objects, the fact that the ordering of performance is preserved between our benchmarks and conventional ones implies that they can be used to estimate real-world OoD detection performance. We visualise Near OoD samples from 10 ImageNet classes in \cref{fig:morphed_samples_with_classnames} and observe that Near OoD images contain patches from original classes in an odd order which makes the images unrecognizable, while still preserving close perceptual proximity to original classes. We find this to be true in general for other classes too. In fact, we find that it is not necessary for an OoD image to represent a real-world object as long as it captures certain desirable properties. For near OoD, the desirable property is to be semantically dissimilar while lying in the close perceptual vicinity of the training distribution.

\subsection{Outlier Exposure on Near OoD Datasets}
\label{sec:exp_outlier_exposure}
\begin{table}[!t]
\centering
\scriptsize
\resizebox{0.8\linewidth}{!}
{
\begin{tabular}{ccccccccccc}
\toprule
\textbf{Model} & \textbf{Outlier Dataset} & \textbf{Test Accuracy} & \multicolumn{4}{c}{\textbf{AUROC}} & \multicolumn{4}{c}{\textbf{AUPRC}} \\
\cmidrule{4-11}
& & & \textbf{SVHN} & \textbf{C10} & \textbf{N C100} & \textbf{Tiny-ImageNet} & \textbf{SVHN} & \textbf{C10} & \textbf{N C100} & \textbf{Tiny-ImageNet} \\
\midrule
\multirow{4}{*}{ResNet-50} & None & $79.52$ & $80.97$ & $78.98$ & $56.38$ & $79.52$ & $88.97$ & $74.92$ & $65.97$ & $76.57$ \\
                           & SVHN & $78.79$ & $-$ & $79.95$ & $60.45$ & $80.02$ & $-$ & $78.82$ & $69.31$ & $77.82$ \\
                           & C10 & $78.98$ & $82.97$ & $-$ & $62.60$ & $83.11$ & $90.54$ & $-$ & $70.75$ & $78.73$ \\
                           & N C100 & $78.82$ & $\mathbf{87.02}$ & $\mathbf{81.65}$ & $-$ & $\mathbf{84.40}$ & $\mathbf{94.51}$ & $\mathbf{82.81}$ & $-$ & $\mathbf{80.70}$\\
\midrule
\multirow{4}{*}{Wide-ResNet-28-10} & None & $80.46$ & $81.46$ & $80.54$ & $62.69$ & $81.84$ & $90.52$ & $76.42$ & $68.91$ & $78.49$ \\
                                   & SVHN & $80.13$ & $-$ & $82.97$ & $64.44$ & $82.77$ & $-$ & $77.32$ & $69.95$ & $79.68$\\
                                   & C10 & $79.7$ & $84.45$ & $-$ & $64.52$ & $82.98$ & $91.96$ & $-$ & $73.55$ & $80.61$\\
                                   & N C100 & $79.92$ & $\mathbf{87.23}$ & $\mathbf{84.31}$ & $-$ & $\mathbf{85.63}$ & $\mathbf{92.88}$ & $\mathbf{79.22}$ & $-$ & $\mathbf{81.42}$ \\

\bottomrule
\end{tabular}
}
\vspace{1mm}
\caption{AUROC and AUPRC scores obtained by performing outlier exposure \cite{hendrycks2018deep} on models trained on CIFAR-100 (C100). Models tuned using Near OoD CIFAR-100 (N C100) consistently obtain the highest AUROC and AUPRC scores.}
\label{table:outlier_exposure}
\end{table}

\begin{table}[!t]
\centering
\scriptsize
\resizebox{0.6\linewidth}{!}
{
\begin{tabular}{ccccccccc}
\toprule
\textbf{Model} & \textbf{Im-val} & \textbf{Im-A} & \textbf{Im-v2} & \multicolumn{2}{c}{\textbf{Im-C}} & \textbf{Im-R} & \textbf{Im-Sketch} & \textbf{S Im (Ours)} \\
&  \multicolumn{3}{|c|}{\textbf{\textit{ECE \%}}} & \textbf{\textit{Avg ECE \%}} & \textbf{\textit{Max ECE \%}} & \multicolumn{3}{|c|}{\textbf{\textit{ECE \%}}} \\
\midrule
ViT-B-16 & $3.62$ & $14.16$ & $7.43$ & $11.18$ & $18.61$ & $5.38$ & $15.44$ & $\mathbf{26.15}$ \\
ViT-B-32 & $3.70$ & $23.13$ & $8.15$ & $11.32$ & $19.21$ & $7.69$ & $17.64$ & $\mathbf{29.15}$ \\
ViT-L-16 & $2.35$ & $12.67$ & $7.30$ & $9.42$ & $13.44$ & $4.79$ & $14.97$ & $\mathbf{22.76}$ \\
ViT-L-32 & $2.51$ & $13.20$ & $7.62$ & $11.03$ & $15.76$ & $4.85$ & $15.12$ & $\mathbf{23.54}$ \\
\bottomrule
\end{tabular}
}
\vspace{1mm}
\caption{ECE \% on standard ImageNet (Im) shifts compared to our Shifted ImageNet (S Im).}
\label{table:ece_imagenet_c_shifted_imagenet}
\vspace{-5mm}
\end{table}

\begin{figure}[!t]
    \centering
    \begin{subfigure}{0.15\linewidth}
        \centering
        \includegraphics[width=\linewidth]{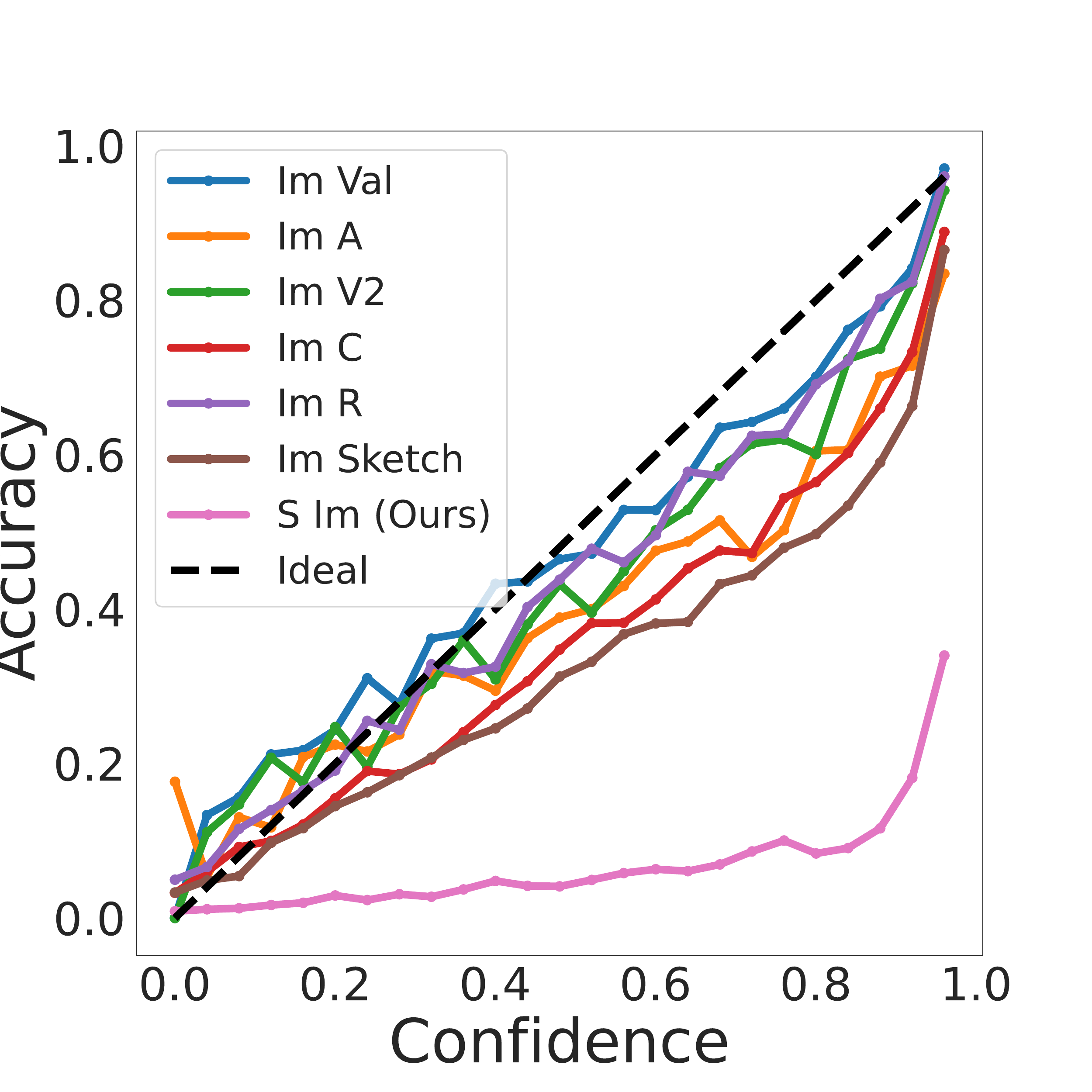}
        \caption{VB16}
        \label{subfig:vitb16_reliability_plot}
    \end{subfigure}
    \begin{subfigure}{0.15\linewidth}
        \centering
        \includegraphics[width=\linewidth]{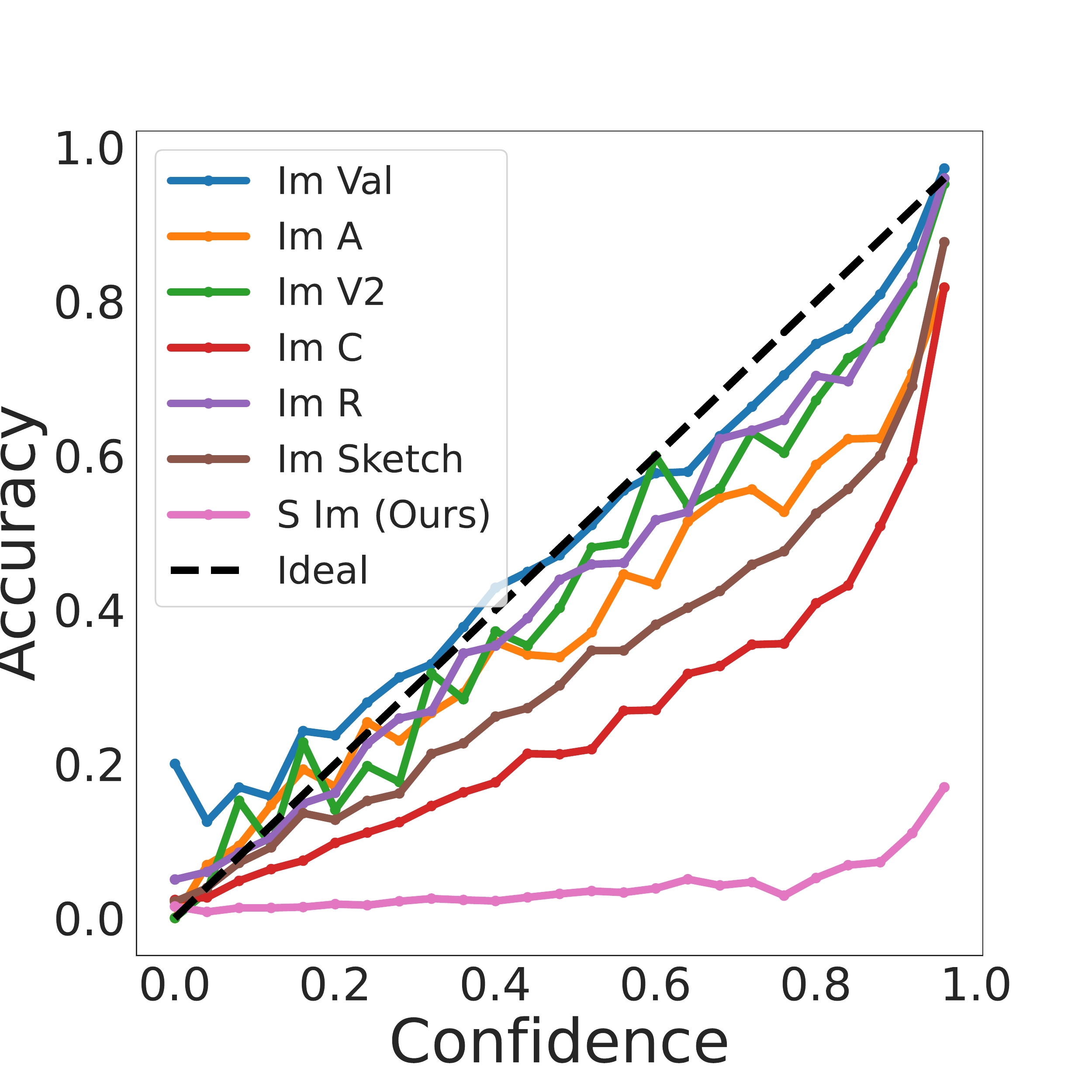}
        \caption{VB32}
        \label{subfig:vitb32_reliability_plot}
    \end{subfigure}
    \begin{subfigure}{0.15\linewidth}
        \centering
        \includegraphics[width=\linewidth]{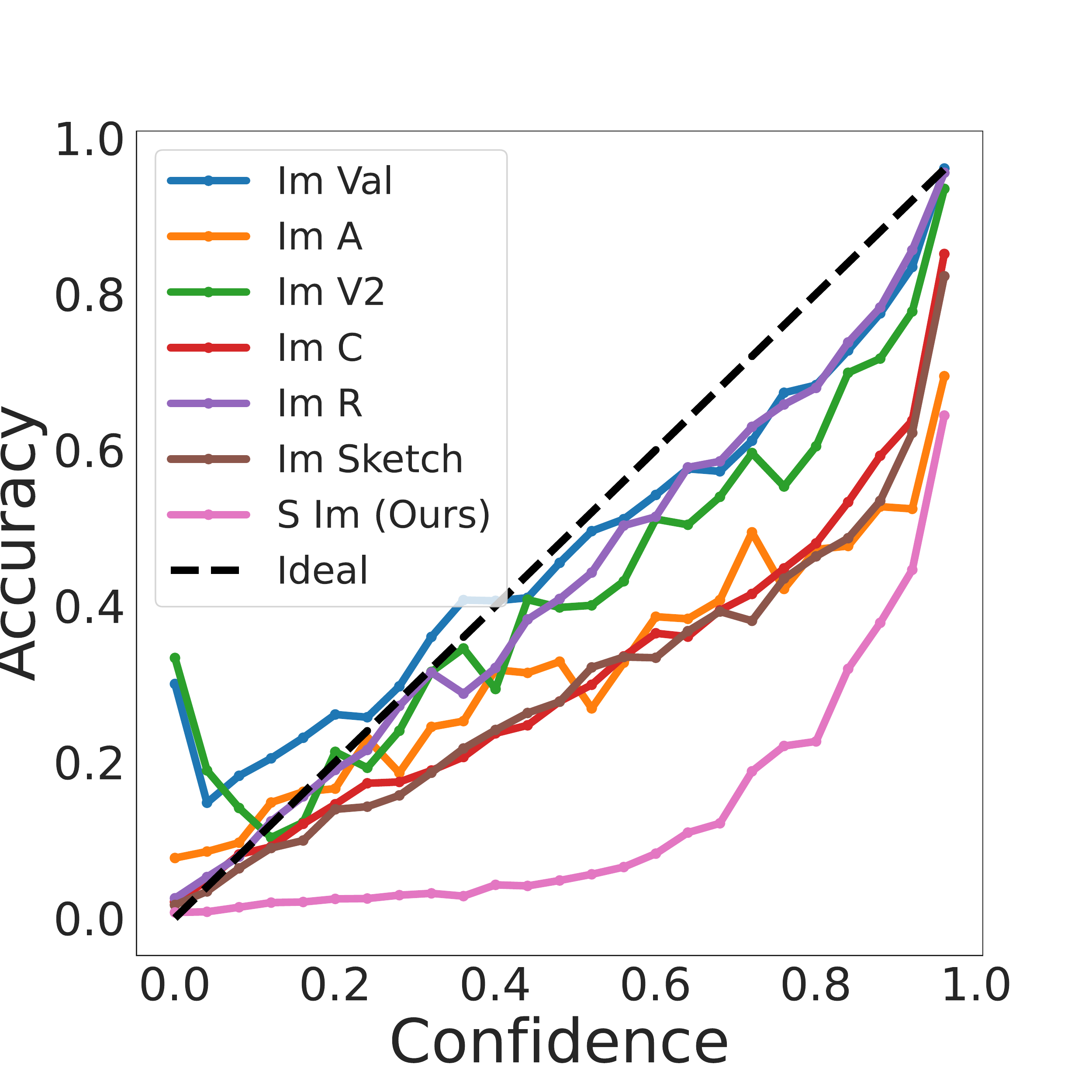}
        \caption{VL16}
        \label{subfig:vitl16_reliability_plot}
    \end{subfigure}
    \begin{subfigure}{0.15\linewidth}
        \centering
        \includegraphics[width=\linewidth]{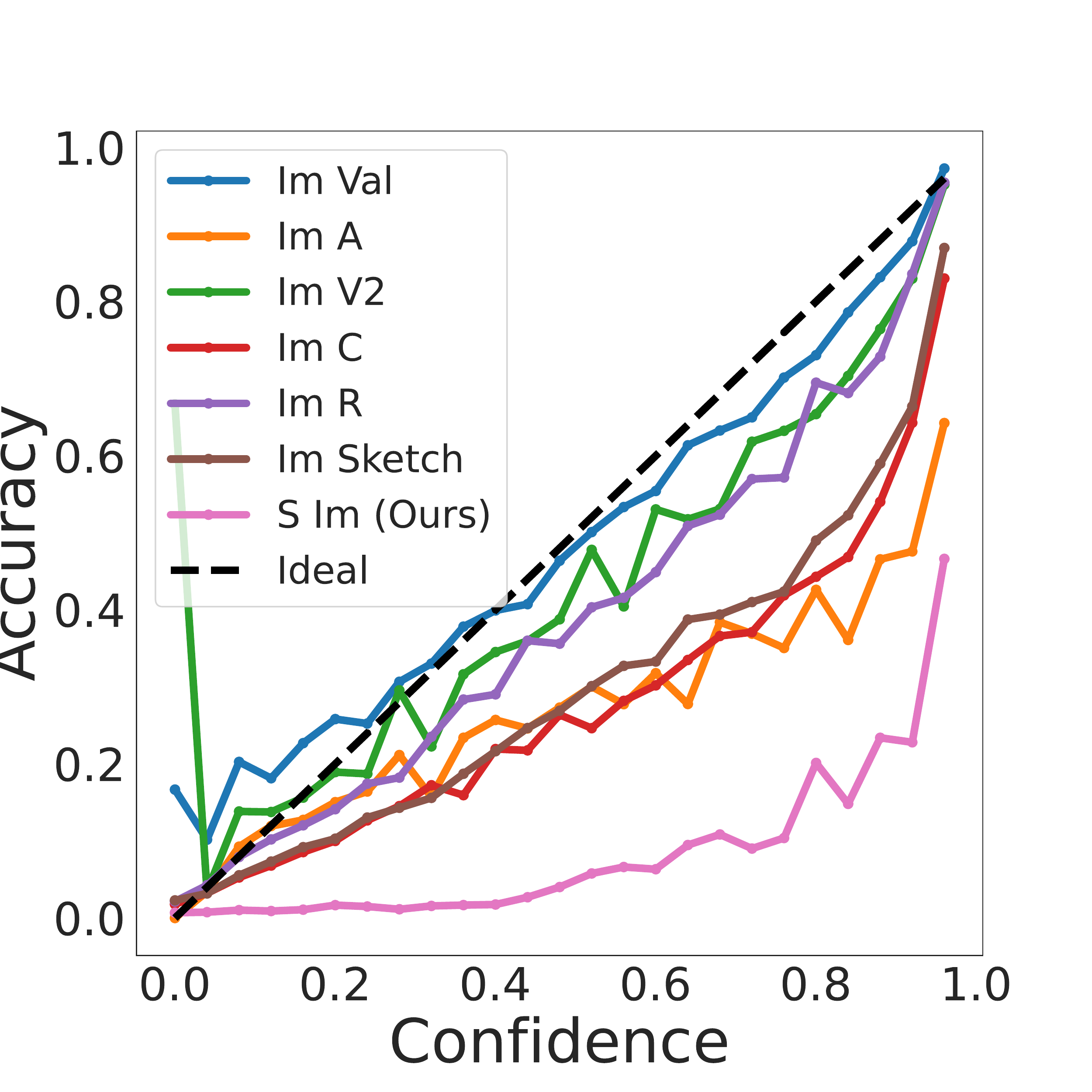}
        \caption{VL32}
        \label{subfig:vitl32_reliability_plot}
    \end{subfigure}
    \caption{Reliability plots for ImageNet shifts}
    \label{fig:reliability_plots_shifted_imagenet}
    \vspace{-1mm}
\end{figure}

In \cite{hendrycks2018deep}, exposure to outliers during training was proposed as a way to improve model performance on OoD datasets. In outlier exposure, models are trained on two datasets: i) the training set on which the loss is the usual cross-entropy loss and ii) the outlier dataset on which the loss is the cross-entropy between the softmax distribution and a uniform distribution over class labels. The assumption is that exposure to good outlier datasets will make the model detect any unseen outlier datasets as well. In this experiment, we want to see how models can improve on OoD detection performance once exposed to our generated near OoD  samples as outliers.

To do this, we train a ResNet-50 and a Wide-ResNet-28-10 on CIFAR-100 using SVHN, CIFAR-10 and Near OoD CIFAR-100 as outlier datasets. Training details can be found in \cref{app:additional_training_details}. For both models, we also compare with a baseline with no exposure to outliers. Finally, we use Tiny-ImageNet as an independent OoD dataset which is not used for outlier exposure. We present the test accuracy, AUROC and AUPRC scores for all models in \cref{table:outlier_exposure}. \emph{Again, we observe a clear ordering in performance improvement where models trained using Near OoD CIFAR-100 as outliers outperform models trained with CIFAR-100 as outliers, which in turn outperform models trained using SVHN as outliers. All models outperform the ones trained without any outliers.}

The above observation provides additional evidence to support the use of Near OoD samples, not just to benchmark OoD detection baselines, but also to improve them through outlier exposure. Furthermore, it corroborates our previous observation that even an image which does not represent any real world object can be very useful if it captures desirable properties in terms of semantic and perceptual similarity.

\subsection{Evaluating Shifted Datasets}
\label{sec:exp_shifted_datasets}

In this experiment, we evaluate our method of learning distribution shifts by comparing the shifted datasets with other well-known synthetic and real-world shifts. For CIFAR-10, we compare with CIFAR-10-C \cite{hendrycks2019benchmarking} and for ImageNet, we compare with synthetic shifts: ImageNet-C, ImageNet-R (renditions) \cite{hendrycks2020many}, ImageNet-Sketch \cite{wang2019learning} and real-world shifts: ImageNet-A \cite{hendrycks2021natural} and ImageNet-V2 \cite{recht2019imagenet}. For CIFAR-10-C and ImageNet-C, we use corrupted images at the highest intensity 5.

We report the Expected Calibration Error (ECE) in \cref{table:ece_cifar10_c_shifted_cifar10} of the appendix for CIFAR-10 models and \cref{table:ece_imagenet_c_shifted_imagenet} for ImageNet. Detailed results for each corruption type can be found in \cref{app:additional_results}. For each model, the ECE for every competitive dataset is starkly lower than the ECE obtained on our Shifted datasets. To better understand the effect of learned shifts, we compute the VGG LPIPS between the ImageNet val set and corruptions taken from ImageNet-C and from our method. Note that this is different from the FID analysis in \cref{fig:shifted_morphed_perceptual_meaning} as unlike FID, here, we are computing difference between individual pairs of images. We ensure that the L2 norm of the difference between normal and corrupted images is same (set to 50) for all corruption types. Results are in \cref{fig:lpips} in the appendix. The high LPIPS for our shift suggests that the corruptions learnt are transformations which decreases perceptual similarity from the perspective of a neural net. At the same time, minimizing an ensemble's MI in $\mathcal{L}_{\mathrm{Shift}}$ encourages corruptions which make classifiers confident on their predictions. Hence, we have high ECE scores indicating miscalibrated models. This is further corroborated from the reliability plots for Shifted-ImageNet vs all other shifts in \cref{fig:reliability_plots_shifted_imagenet}, where for Shifted-ImageNet, models consistently have lower accuracy as compared confidence, i.e., they are overconfident and miscalibrated.

\subsection{Evaluation of OoD detection in real life}
Having evaluated our benchmarks, in order to use them in real-life applications, we need to verify if we can compare baselines sensibly using our benchmarks. Indeed that is the very purpose of a benchmark. We make the observation here that in all our experiments across datasets, the ordering of performance between baselines is broadly consistent between our benchmarks and the conventional ones. \emph{Models which perform well on our benchmarks also perform well on conventional benchmarks and vice versa.} We show this particularly for ImageNet in a radar plot (\cref{fig:radar} in the appendix). Assuming that conventional OoD detection benchmarks generalise to and are indicative of real-world performance, we can then use the above observation as validation for the proposed benchmarks. Interestingly, this also shows that we might not need an OoD dataset in the first place to evaluate a model's OoD detection performance. We could reliably estimate the OoD detection performance of any model just from the training set by following our proposed method.

\section{Conclusion}
Reliably evaluating the behaviour of models in unknown scenarios is an open problem and is extremely difficult because of infinitely many situations that could potentially fulfill our notion of ``unknown" with respect to in-distribution. 
To our knowledge, our work is the very first step in the direction where we do not advocate the naive approach of testing on arbitrarily chosen ``other" datasets. Rather we propose to learn the distribution of samples satisfying constraints mimicking our notion of what is out-of-distribution. Through numerous experiments, we show that our generated samples provide for a more challenging and reliable benchmark for even the current state-of-the-art OoD detection baselines.

\section{Acknowledgements}
Jishnu started this project as part of his internship at Meta AI (June 2021 - Sept 2021) and then continued working on it after returning to Oxford University. He was also employed part-time by Meta (Nov 2021 - Jan 2022) to work on this project. This work is also supported by the UKRI grant: Turing AI Fellowship EP/W002981/1 and EPSRC/MURI grant: EP/N019474/1. We would like to thank the Royal Academy of Engineering, FiveAI and Meta AI.

\bibliographystyle{unsrtnat}
\bibliography{references}

\newpage
\appendix

\appendix

\section{Additional Training Details}
\label{app:additional_training_details}

In this section, we provide training details for all models used to report results in \cref{sec:experiments} of the main paper.

\subsection{Classifier Training}
\label{sec:classifier_training_details}

\textbf{MNIST models}: For experiments on MNIST, we use 4 convolutional architectures: LeNet, AlexNet, VGG-11 and ResNet-18. Each model has been trained on a single 12 GB TITAN Xp GPU for 100 epochs using SGD as the optimizer with a momentum of 0.9. The initial learning rate used was 0.1 and there are learning rate drops by a factor of 10 at training epochs 40 and 60. The training batch size used for MNIST is 256.

\textbf{CIFAR-10/100 models}: All convolutional classifiers: DenseNet-121, ResNet-50/110, VGG-16,  are trained using the Pytorch framework with a single 12 GB TITAN Xp GPU. To train models on CIFAR-10/100, we use the SGD optimiser with a momentum of 0.9 and a weight decay of $5e^{-4}$. We train each model for 350 epochs using 0.1 as the learning rate and a learning rate drop by a factor of 10 at training epochs 150 and 250. We use a training batch size of 128 and augment the training set using random crops and random horizontal flips.

For Vision Transformer (ViT) models trained on CIFAR-10/100, we use 4 12 GB TITAN Xp GPUs to train a single model. We train 4 different ViT models: ViT-B-16/32 and ViT-L-16/32, using an image size of $224 \times 224$ and other conventional augmentations including random crop and random horizontal flips. All the ViTs are pretrained on ImageNet-21K. We use a SGD with a momentum of 0.9 and a learning rate of $3e^{-2}$ with a cosine learning rate decay. We use 500 warmup steps for each model and train them for a maximum of 10000 steps. We use a training batch size of 256 for the ViT models.

\textbf{ImageNet models}: For ImageNet, we use pretrained Vision Transformers for all evaluation purposes.\footnote{See \texttt{github.com/rwightman/pytorch-image-models} for details.}

\textbf{Classifier Suite for computing $\mathcal{L}_{\mathrm{MI}}$} Note that the in order to compute $\mathcal{L}_{\mathrm{MI}}$, we use a single ensemble containing models, each with a different architecture. For MNIST experiments, we use 4 different models: LeNet, AlexNet, VGG-11 and ResNet-18 (one model of each architecture) as the ensemble to compute mutual information over. Similarly, for CIFAR-10/100, we use 6 models with 6 different architectures: DenseNet-121, ResNet-50/110, VGG-16, Wide-ResNet-28-10 and Inception-v3. Finally, for ImageNet, we use a set of pretrained classifiers from the Pytorch \texttt{torchvision.models} library. In particular, we get ResNet-18, MobileNet-v3-Large and EfficientNet-B0. Note that the use of ensembles with different architectures is to encourage higher variability in predictions and representations within the ensemble, thereby encouraging higher mutual information for predictions. Ensembles used for the evaluation of generated samples all have the same architecture. All the classifiers used for computing $\mathcal{L}_{\mathrm{MI}}$ are trained using the same dataset-specific settings as mentioned above.

\subsection{Training Pix-2-Pix GAN}

In order to train a Pix-2-Pix GAN, we use $\mathcal{L}_{\mathrm{Shift}}$ defined in \cref{eq:shift_loss} as the loss function for the generator of the GAN and there is no change to the loss of the discriminator. However, note that the target image for the Pix-2-Pix discriminator is the same as the input. Thus the loss of the discriminator can be given as:
\begin{equation}
\begin{split}
    \mathcal{L}_{\mathrm{D}_{\mathrm{Pix-2-Pix}}} = & \mathbb{E}_{\mathbf{x}}\left[\log (D(\mathbf{x}, \mathbf{x}))\right] - \\
    & \mathbb{E}_{\mathbf{x}, \mathbf{z}}\left[\log (1-D(\mathbf{x}, G(\mathbf{x}, \mathbf{z})))\right]
\end{split}
\end{equation}
For $\mathcal{L}_{\mathrm{MI}}$, in \cref{eq:shift_loss}, we use the method specified above. We use a single 12 GB TITAN Xp GPU to train the Pix-2-Pix model on CIFAR-10 and 8 such GPUs to train on ImageNet. We use a training batch size of 256 and train the model for 100 epochs using Adam as the optimizer, a learning rate of 0.0002 and beta values 0.5 and 0.999. All other training settings are the same as specified in the original Pix-2-Pix paper \cite{isola2017image}.

\subsection{Training GAN}

For generating Near OoD samples (i.e., Near OoD), we use a DCGAN for MNIST and a BigGAN for CIFAR-10/100 and ImageNet. We use a single 12 GB TITAN Xp GPU to train DCGAN for MNIST and BigGAN for CIFAR-10/100. However, we use 8 such GPUs to train a single BigGAN on ImageNet. The loss function for the discriminator of the GAN undergoes no change and is shown as follows:
\begin{equation}
    \begin{split}
        \mathcal{L}_{\mathrm{D}_{\mathrm{GAN}}} & = \mathbb{E}_{\mathbf{x}}\left[\log D(\mathbf{x})\right] - \mathbb{E}_{\mathbf{z}}\left[\log (1-D(G(z)))\right]
    \end{split}
\end{equation}
The loss function for the generator is $\mathcal{L}_{\text{Near OoD}}$ as shown in \cref{eq:morph_loss}. The $\mathcal{L}_{\mathrm{MI}}$ in $\mathcal{L}_{\text{Near OoD}}$ is computed as described above. We train all GANs for 100 epochs and all other training details for the GANs are exactly the same as set out in their respective repositories.\footnote{See \texttt{github.com/ajbrock/BigGAN-PyTorch} for details on training BigGAN.}

\subsection{Outlier Exposure}

In the outlier exposure experiment, we train a ResNet-50 and a Wide-ResNet-28-10 on CIFAR-100 using the standard training procedure set out in \cref{sec:classifier_training_details}. However, in the loss, following \cite{hendrycks2018deep}, in addition to the cross-entropy term, we also have an additional regulariser which computes the cross-entropy of the output with a uniform distribution for outlier samples.

\section{Additional Results}
\label{app:additional_results}

In this section, we present additional results to support the results in the main paper. 

\begin{figure}[!t]
    \centering
    \includegraphics[width=0.5\linewidth]{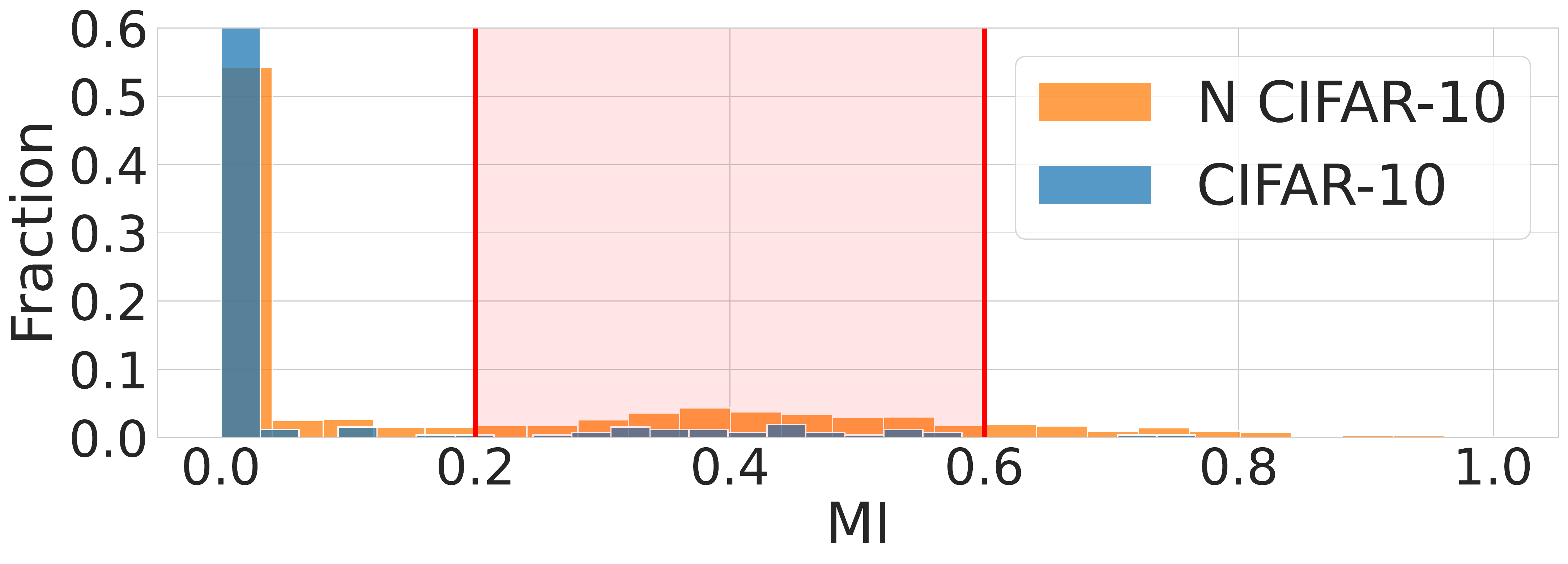}
    \caption{MI of the ensemble for Near OoD (N) CIFAR-10 and real CIFAR-10 samples. We use this to find plausible thresholds of MI.}
    \label{fig:cifar10_mi}
\end{figure}

\textbf{MI overlap}: In \cref{fig:cifar10_mi}, we show the mutual information of the training ensemble on real CIFAR-10 samples along with Near OoD generated samples on CIFAR-10. We choose samples which minimise MI overlap between real and generated samples without having a very high MI as that leads to generated samples losing their perceptual similarity with iD. For CIFAR-10, we choose $[0.2, 0.6]$ as the MI interval for generated samples.

\begin{figure*}[!t]
    \centering
    \begin{subfigure}{0.24\linewidth}
        \centering
        \includegraphics[width=\linewidth]{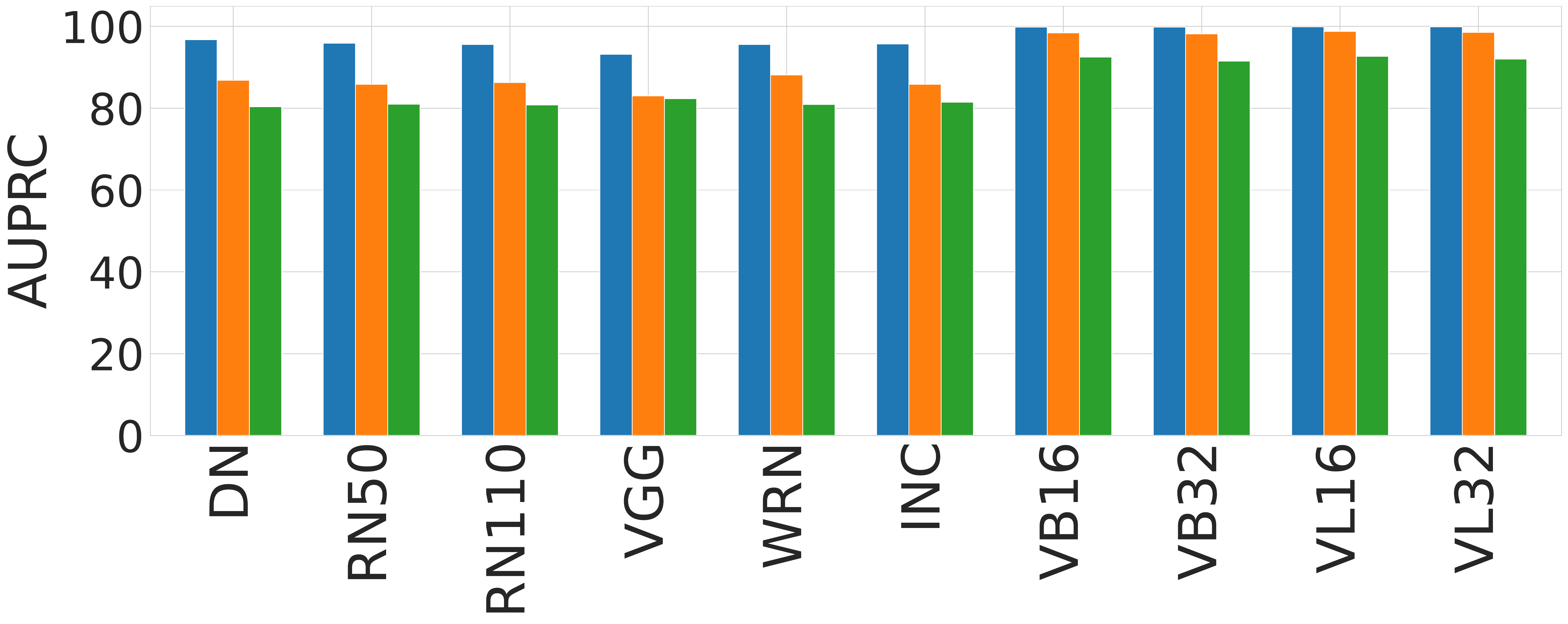}
        \label{subfig:auprc_c10_softmax_entropy}
    \end{subfigure}
    \begin{subfigure}{0.24\linewidth}
        \centering
        \includegraphics[width=\linewidth]{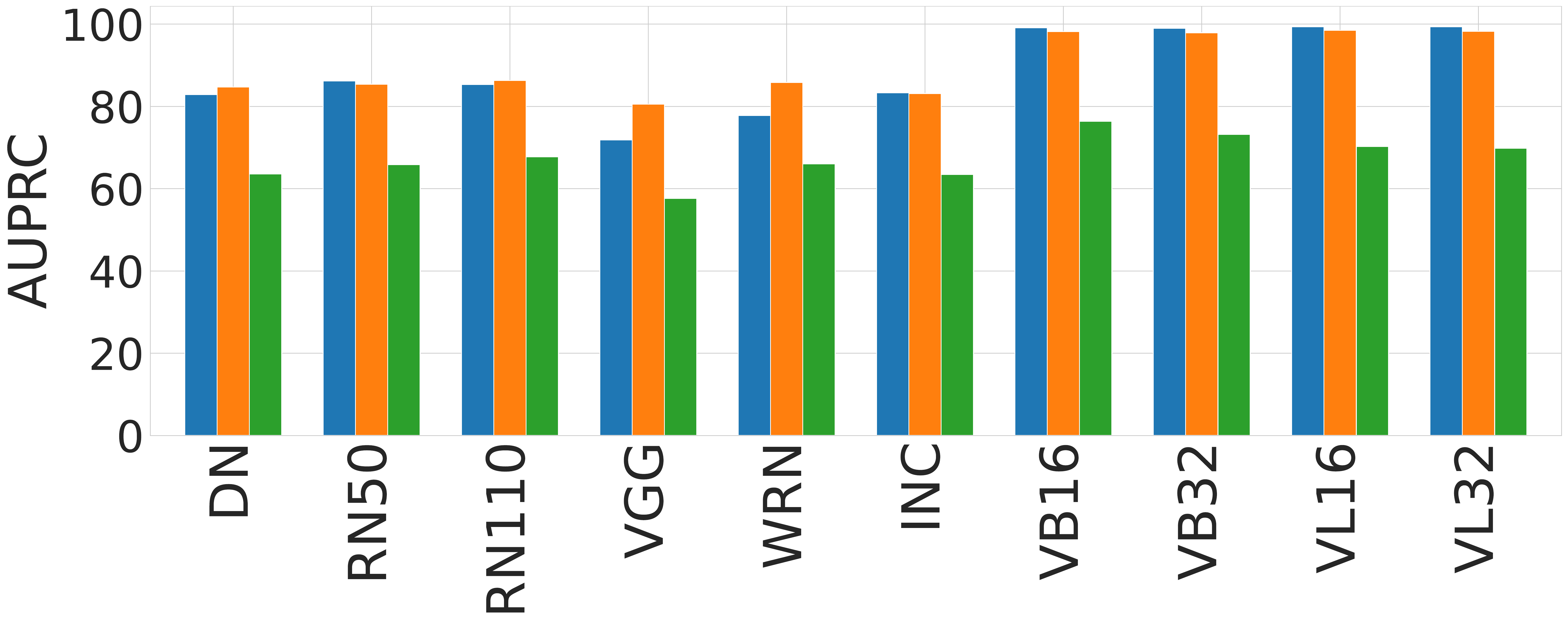}
        \label{subfig:auprc_c10_softmax_confidence}
    \end{subfigure}
    \begin{subfigure}{0.24\linewidth}
        \centering
        \includegraphics[width=\linewidth]{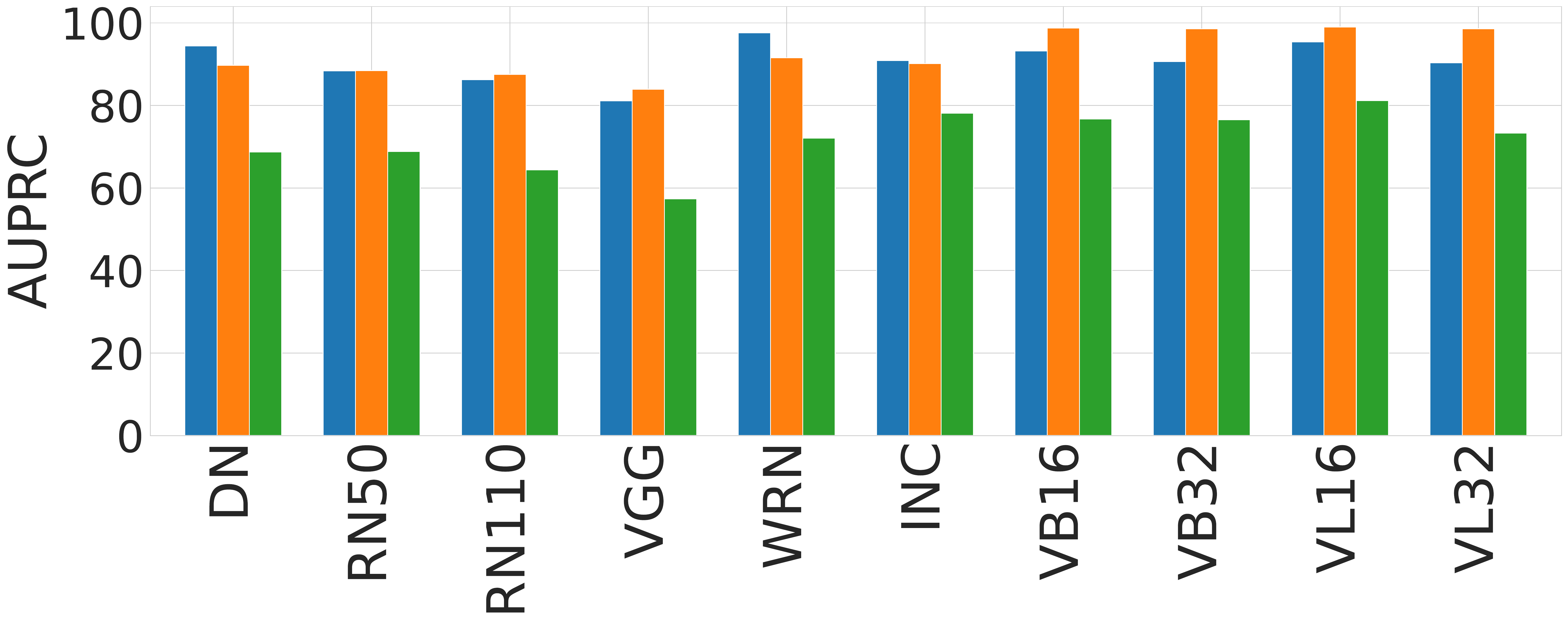}
        \label{subfig:auprc_c10_mahalanobis}
    \end{subfigure}
    \begin{subfigure}{0.24\linewidth}
        \centering
        \includegraphics[width=\linewidth]{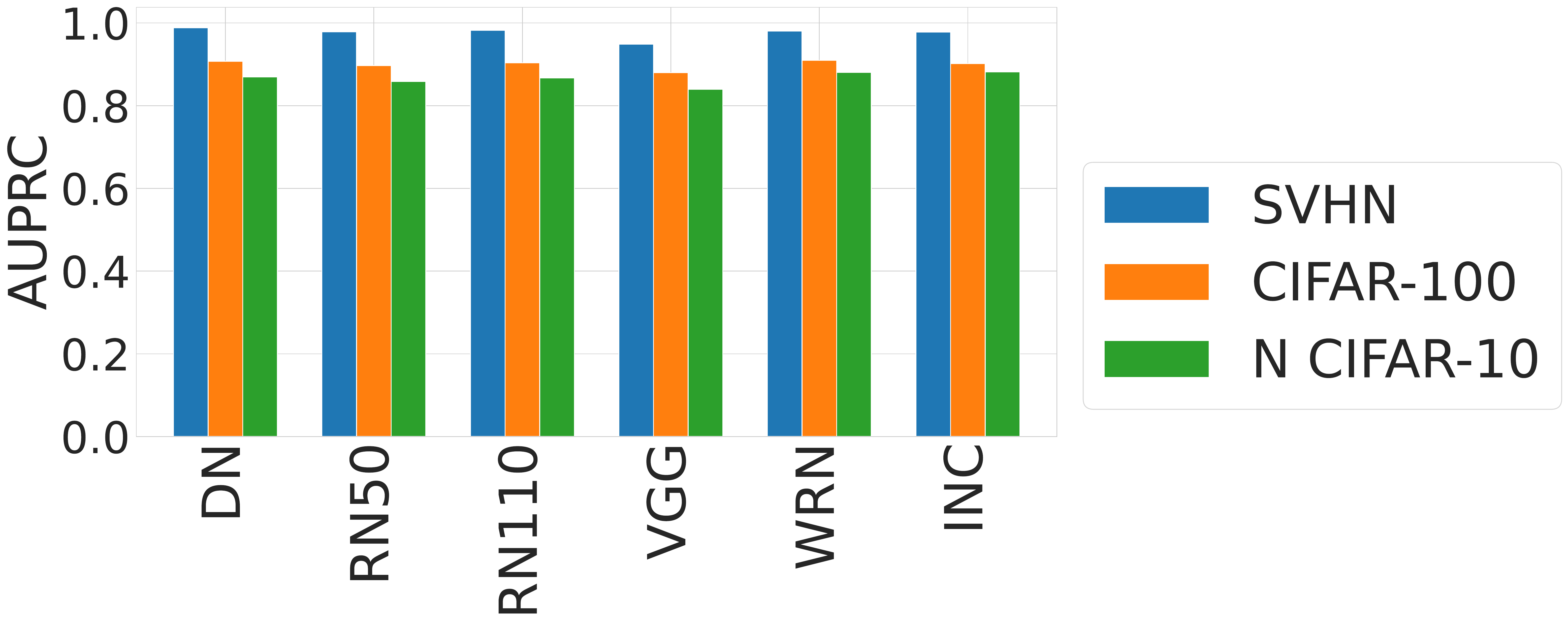}
        \label{subfig:auprc_c10_ensemble}
    \end{subfigure}
    
    \begin{subfigure}{0.24\linewidth}
        \centering
        \includegraphics[width=\linewidth]{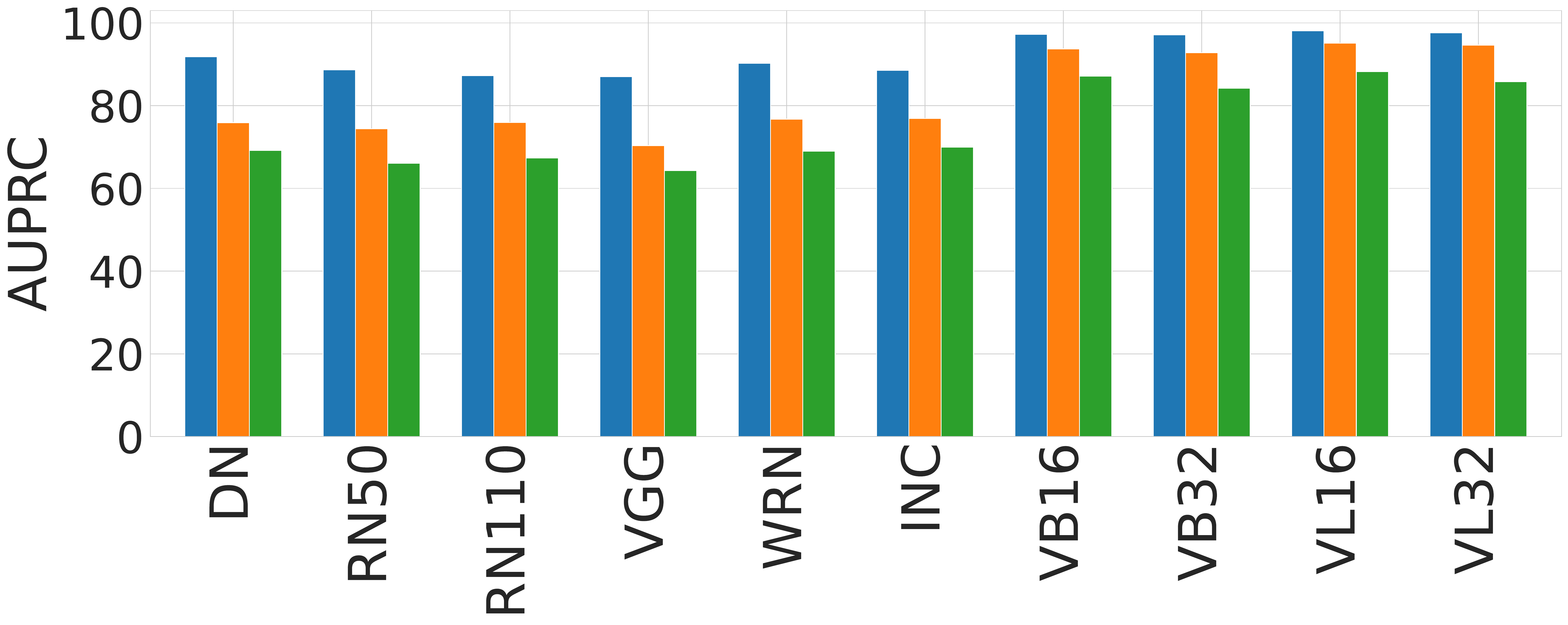}
        \caption{Entropy}
        \label{subfig:auprc_c100_softmax_entropy}
    \end{subfigure}
    \begin{subfigure}{0.24\linewidth}
        \centering
        \includegraphics[width=\linewidth]{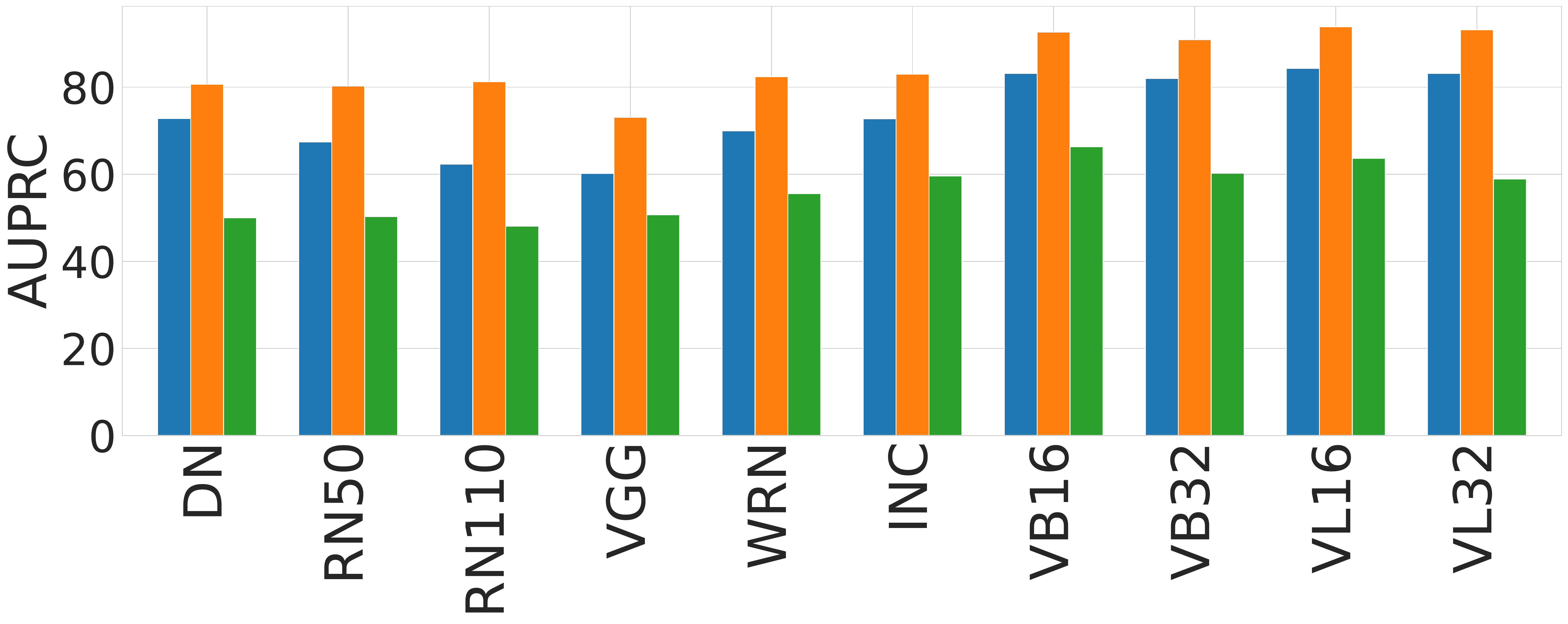}
        \caption{Confidence}
        \label{subfig:auprc_c100_softmax_confidence}
    \end{subfigure}
    \begin{subfigure}{0.24\linewidth}
        \centering
        \includegraphics[width=\linewidth]{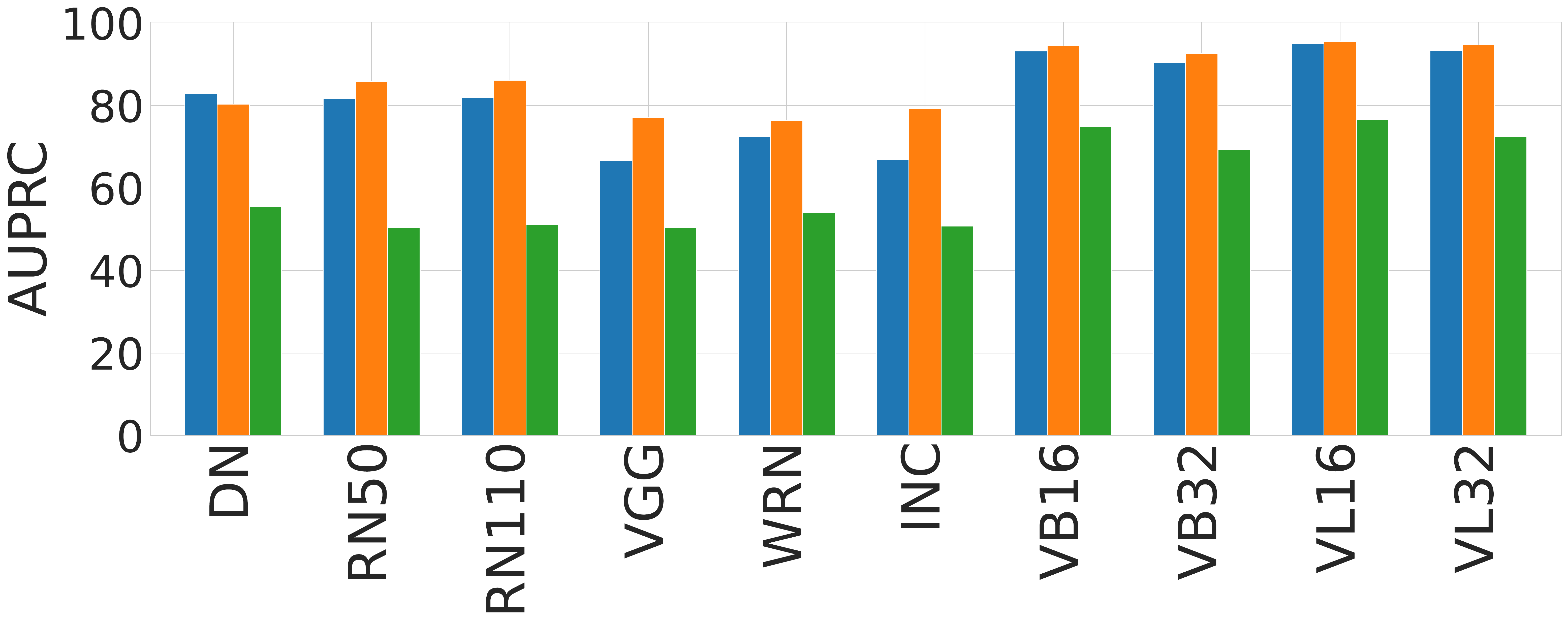}
        \caption{Mahalanobis}
        \label{subfig:auprc_c100_mahalanobis}
    \end{subfigure}
    \begin{subfigure}{0.24\linewidth}
        \centering
        \includegraphics[width=\linewidth]{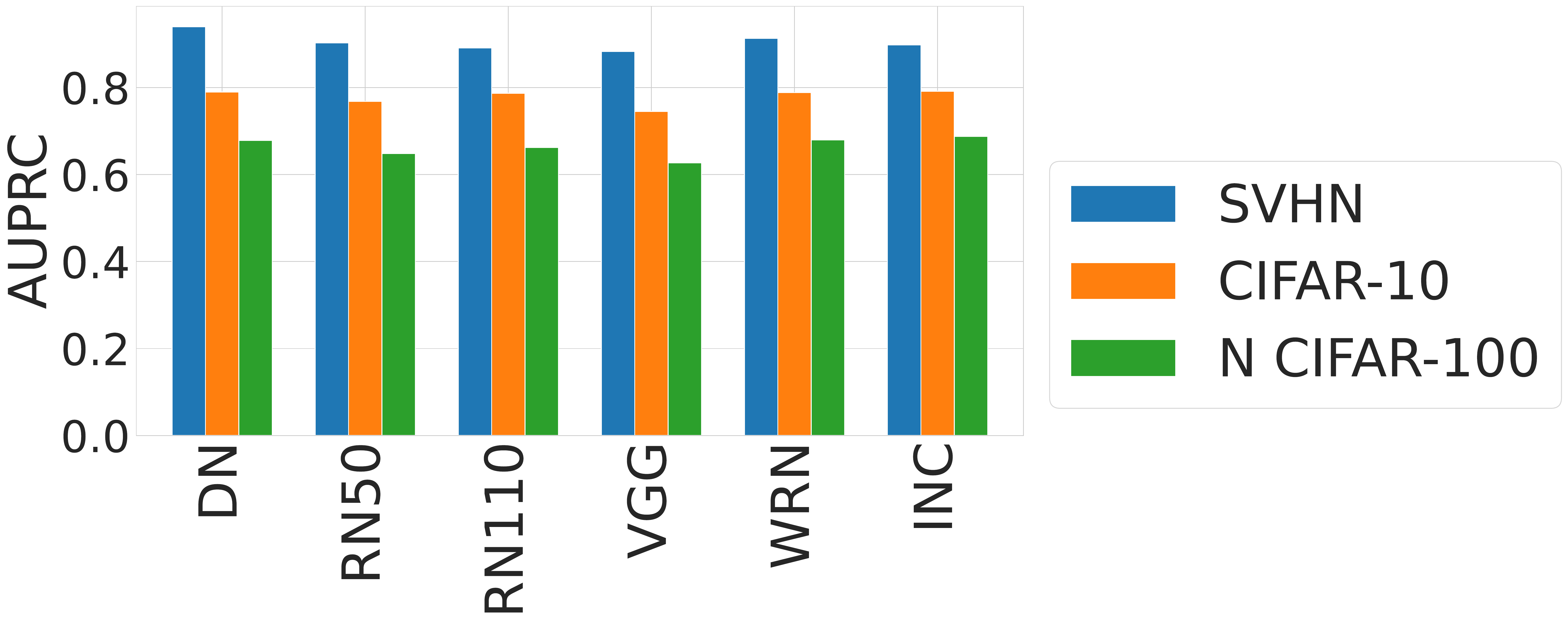}
        \caption{Ensemble}
        \label{subfig:auprc_c100_ensemble}
    \end{subfigure}
    \caption{AUPRC \% for different models, DenseNet-121 (DN), ResNet-50 (RN50), ResNet-110 (RN110), VGG-16, Wide-ResNet-28-10 (WRN) and Inception-v3 (INC), ViT-B-16/32 (VB16/32) and ViT-L-16/32 (VL16/32) trained on CIFAR-10 (first row) and CIFAR-100 (second row) using SVHN, CIFAR-10/100 and Near OoD (N) CIFAR-10/100 as OoD datasets and softmax entropy, confidence, Mahalanobis distance and deep ensemble baselines.}
    \label{fig:auprc_cifar10_100}
\end{figure*}

\textbf{OoD Detection on Near OoD Datasets} In \cref{table:mnist_ood_results}, we present test set accuracy and AUROC scores of models trained on MNIST on the MNIST vs Fashion-MNIST and MNIST vs Near OoD MNIST. In \cref{table:cifar10_100_imagenet_accuracy}, we report the CIFAR-10/100 and ImageNet test set accuracy of all the models we use to evaluate our benchmark. In \cref{table:auroc_cifar10} and \cref{table:auroc_cifar100}, we report the AUROC scores of 6 convolutional models: DenseNet-121, ResNet-50/110, VGG-16, Wide-ResNet-28-10 and Inception-v3 and 4 Vision Transformer models: ViT-B-16/32, ViT-L-16/32 trained on CIFAR-10 and CIFAR-100 respectively. The uncertainty computation method here uses the softmax entropy, softmax confidence and Mahalanobis distance computed from a single deterministic model. We also compute the AUROC scores for a deep ensemble of size 5, using the 6 convolutional architectures and report the correspoding results in \cref{table:auroc_ensemble_cifar10} and \cref{table:auroc_ensemble_cifar100} for models trained on CIFAR-10 and CIFAR-100 respectively. For CIFAR-10, we use SVHN, CIFAR-100 and Near OoD CIFAR-10 as OoD sets and for CIFAR-100, we use SVHN, CIFAR-10 and Near OoD CIFAR-100 as OoD sets. The corresponding AUPRC scores for all models trained on CIFAR-10 and CIFAR-100 are shown in \cref{table:auprc_cifar10} and \cref{table:auprc_cifar100} for deterministic models and \cref{table:auprc_ensemble_cifar10} and \cref{table:auprc_ensemble_cifar100} for deep ensembles respectively. In addition, we also show the AUPRC scores as plots for deterministic models, deep ensembles and Vision Transformers in \cref{fig:auprc_cifar10_100}. Finally, we report AUPRC scores for Vision Transformer models trained on ImageNet using ImageNet-O and Near OoD ImageNet as OoD datasets in \cref{table:auprc_imagenet}.

\begin{table}[!t]
\centering
\scriptsize
\resizebox{0.7\linewidth}{!}
{
\begin{tabular}{cccccc}
\toprule
\textbf{Model} & \textbf{Test Accuracy} & \multicolumn{4}{c}{\textbf{AUROC \%}} \\
\cmidrule{3-6}
& & \textbf{F-MNIST (SE)} & \textbf{F-MNIST (SC)} &  \textbf{N-MNIST (Ours) (SE)} & \textbf{N-MNIST (Ours) (SC)} \\
\midrule
LeNet & $98.97\pm0.02$ & $98.87\pm0.05$ & $98.80\pm0.05$ & $\mathbf{65.29\pm0.12}$ & $\mathbf{64.14\pm0.11}$ \\
AlexNet & $99.04\pm0.03$ & $99.10\pm0.05$ & $99.07\pm0.05$ & $\mathbf{70.31\pm0.15}$ & $\mathbf{69.64\pm0.13}$ \\
VGG-11 & $99.35\pm0.02$ & $99.20\pm0.04$ & $99.17\pm0.03$ & $\mathbf{72.11\pm0.13}$ & $\mathbf{71.85\pm0.13}$ \\
ResNet-18 & $99.54\pm0.02$ & $99.16\pm0.03$ & $99.14\pm0.04$ & $\mathbf{73.15\pm0.13}$ & $\mathbf{72.81\pm0.12}$ \\
\bottomrule
\end{tabular}
}
\vspace{1mm}
\caption{AUROC \% on MNIST using softmax entropy (SE) and softmax confidence (SC) with Fashion(F)-MNIST and Near OoD(N)-MNIST as OoD.}
\label{table:mnist_ood_results}
\end{table}

\begin{table}[!t]
  \centering
    \scriptsize
    \resizebox{0.5\linewidth}{!}
    {
    \begin{tabular}{cccc}
    \toprule
    \textbf{Model} & \multicolumn{3}{c}{\textbf{Test/Val Set Accuracy}} \\
    \cmidrule{2-4}
    & \textbf{CIFAR-10} &  \textbf{CIFAR-100} & \textbf{ImageNet}\\
    \midrule
    DenseNet-121 & $95.66\pm0.05$ & $80.08\pm0.15$ & $-$\\
    ResNet-50 & $95.34\pm0.05$ & $78.26\pm0.33$ & $-$ \\
    ResNet-110 & $95.50\pm0.12$ & $79.50\pm0.27$ & $-$ \\
    VGG-16 & $93.81\pm0.09$ & $74.33\pm0.18$ & $-$ \\
    Wide-ResNet-28-10 & $96.33\pm0.07$ & $80.60\pm0.11$ & $-$ \\
    Inception-v3 & $95.25\pm0.10$ & $78.04\pm0.14$ & $-$ \\
    \midrule
    ViT-B-16 & $99.12\pm0.03$ & $92.73\pm0.04$ & $85.02$\\
    ViT-B-32 & $98.73\pm0.01$ & $92.14\pm0.02$ & $84.72$\\
    ViT-L-16 & $99.18\pm0.01$ & $93.73\pm0.04$ & $86.55$\\
    ViT-L-32 & $99.02\pm0.01$ & $93.29\pm0.04$ & $85.47$\\
    \bottomrule
    \end{tabular}}
    \vspace{1mm}
    \caption{CIFAR-10/100 test and ImageNet val accuracy for CNNs and ViTs used in our evaluation.}
    \label{table:cifar10_100_imagenet_accuracy}
\end{table}

\begin{figure}[!t]
    \centering
    \begin{subfigure}{0.45\linewidth}
    \centering
    \includegraphics[width=\linewidth]{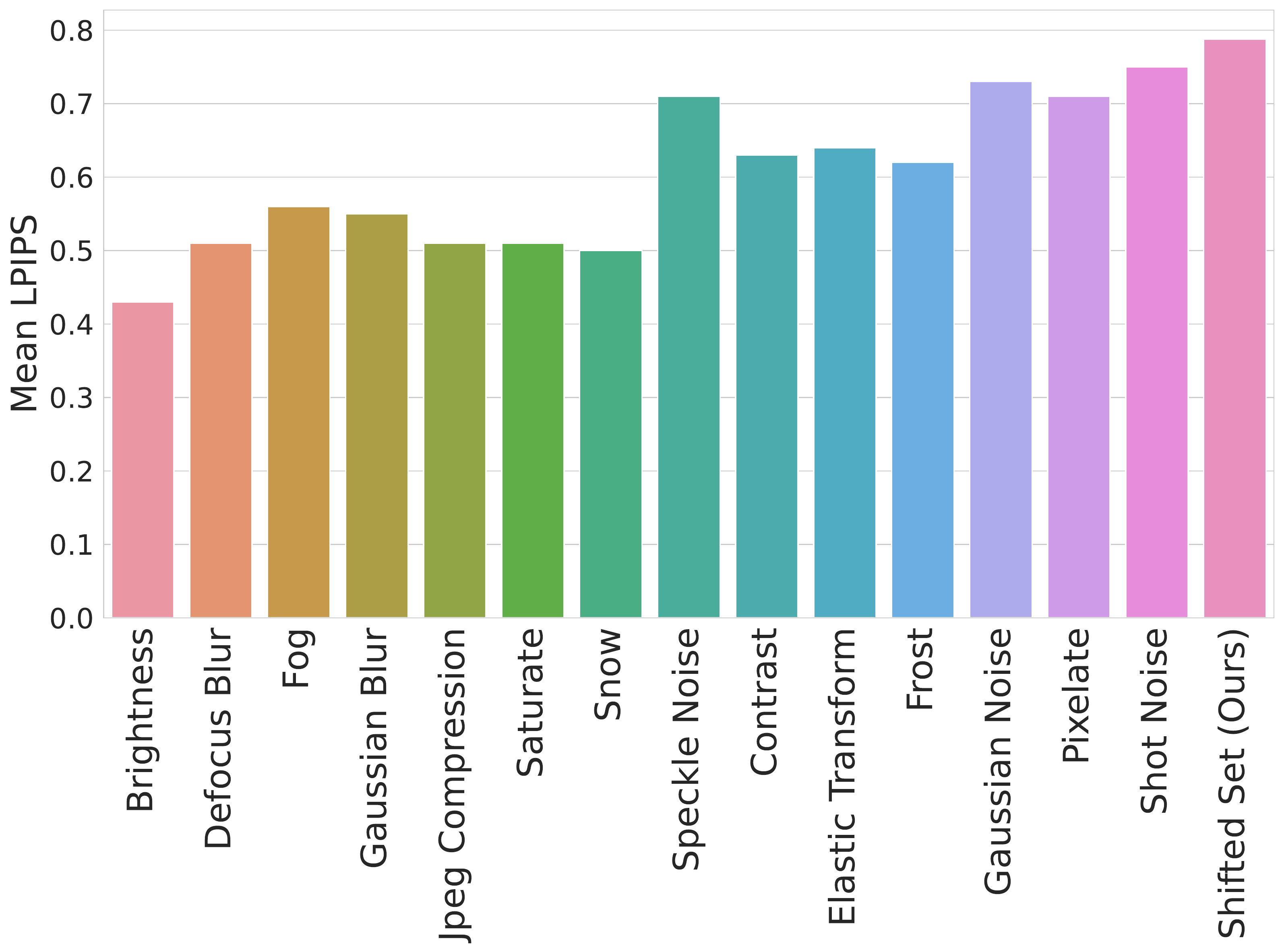}
    \caption{}
    \label{fig:lpips}
    \end{subfigure}
    \begin{subfigure}{0.3\linewidth}
    \centering
    \includegraphics[width=\linewidth]{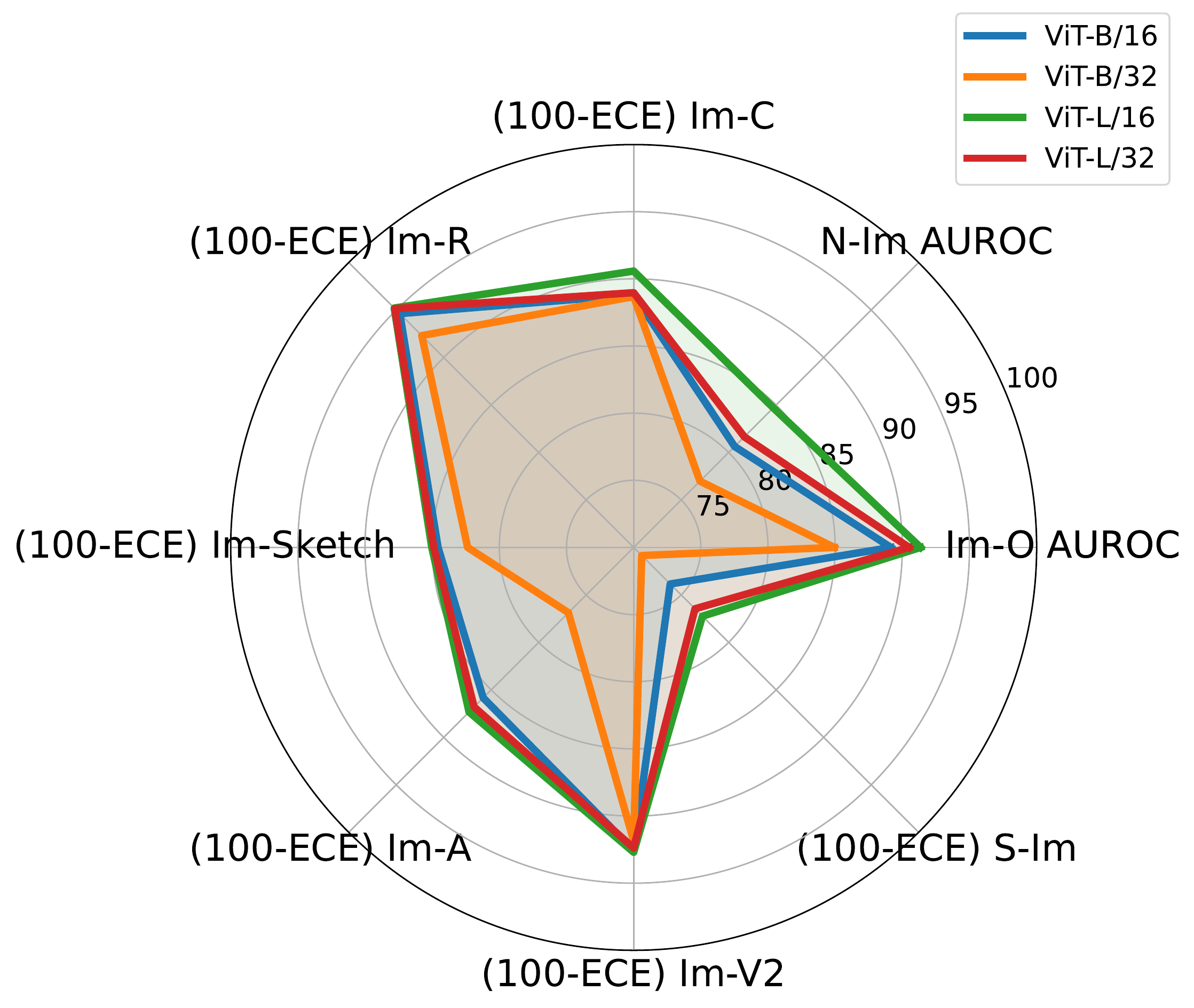}
    \caption{}
    \label{fig:radar}
    \end{subfigure}
    \caption{(a) average LPIPS between ImageNet val and corrupted samples, (b) performance of the 4 ViT architectures on all ImageNet benchmarks: ImageNet-C (Im-C), ImageNet-R (Im-R), ImageNet-O (Im-O), ImageNet-A(Im-A), ImageNet-V2(Im-V2), ImageNet-Sketch (Im-Sketch) and our benchmarks Near OoD ImageNet (N-Im) and Shifted ImageNet (S-Im).}
\end{figure}

\begin{figure*}[!t]
    \centering
    \begin{subfigure}{0.16\linewidth}
        \centering
        \includegraphics[width=\linewidth]{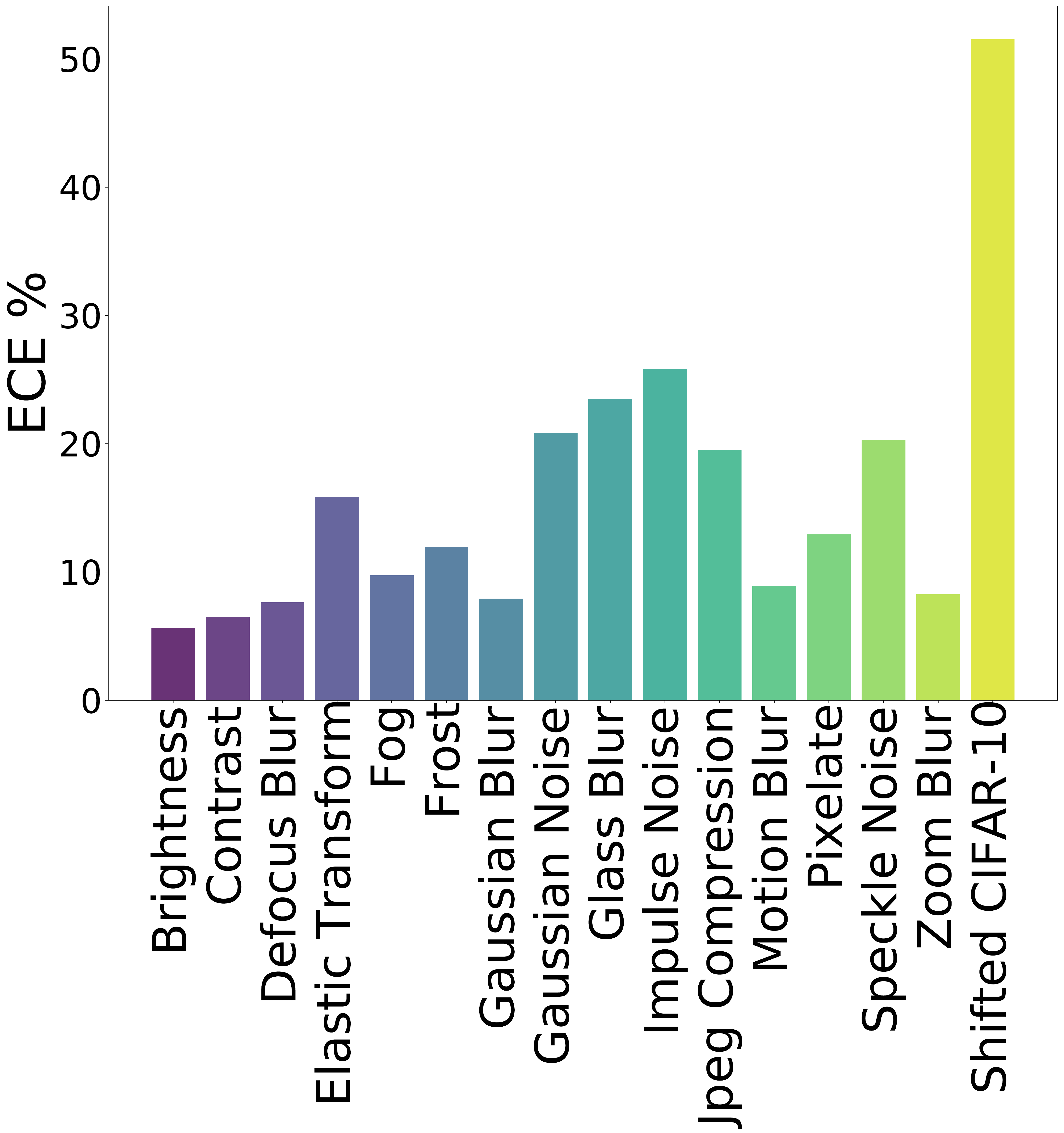}
        \caption{DN-121}
        \label{subfig:ece_cifar10c_densenet121}
    \end{subfigure}
    \begin{subfigure}{0.16\linewidth}
        \centering
        \includegraphics[width=\linewidth]{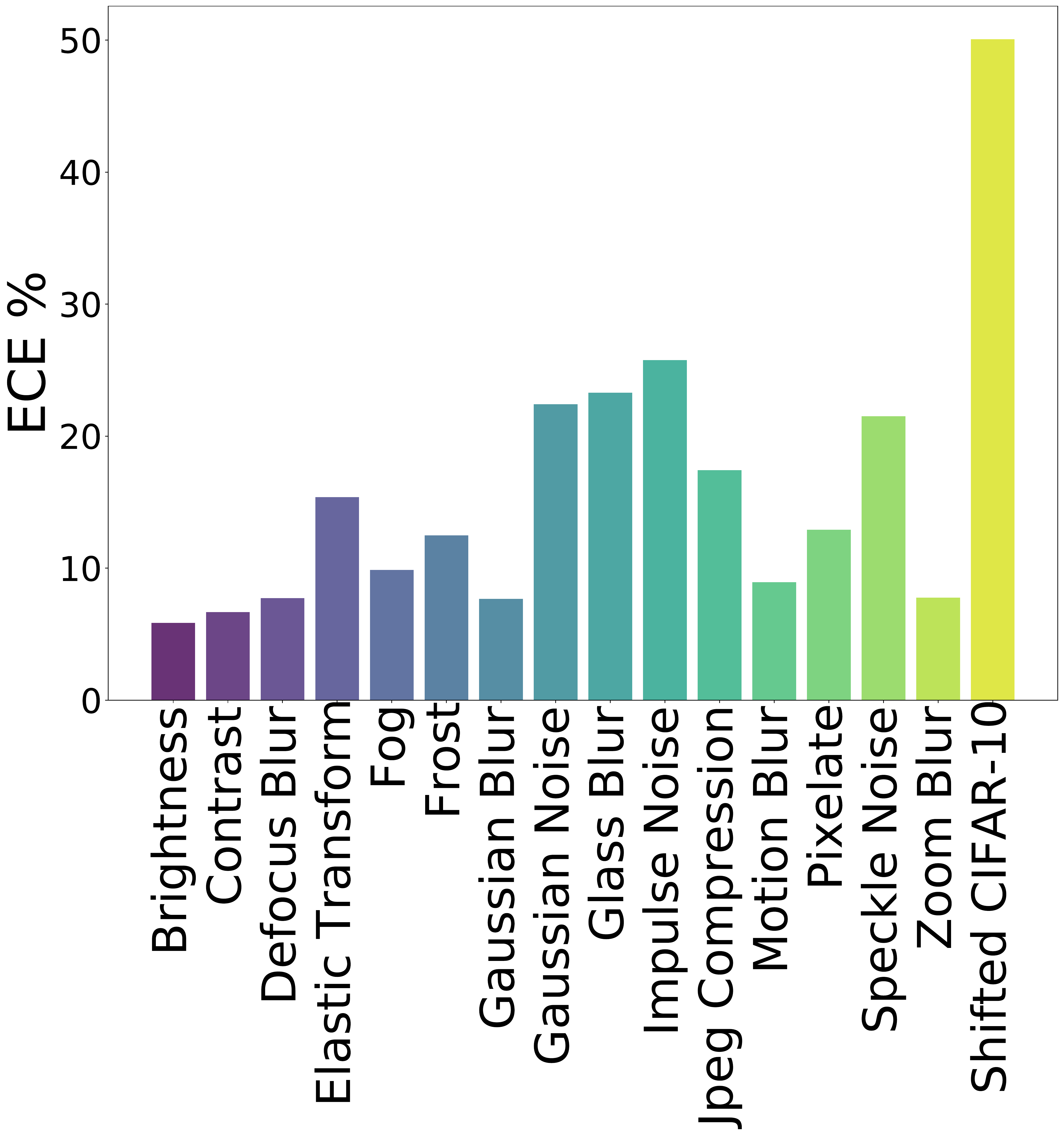}
        \caption{RN-50}
        \label{subfig:ece_cifar10c_resnet50}
    \end{subfigure}
    \begin{subfigure}{0.16\linewidth}
        \centering
        \includegraphics[width=\linewidth]{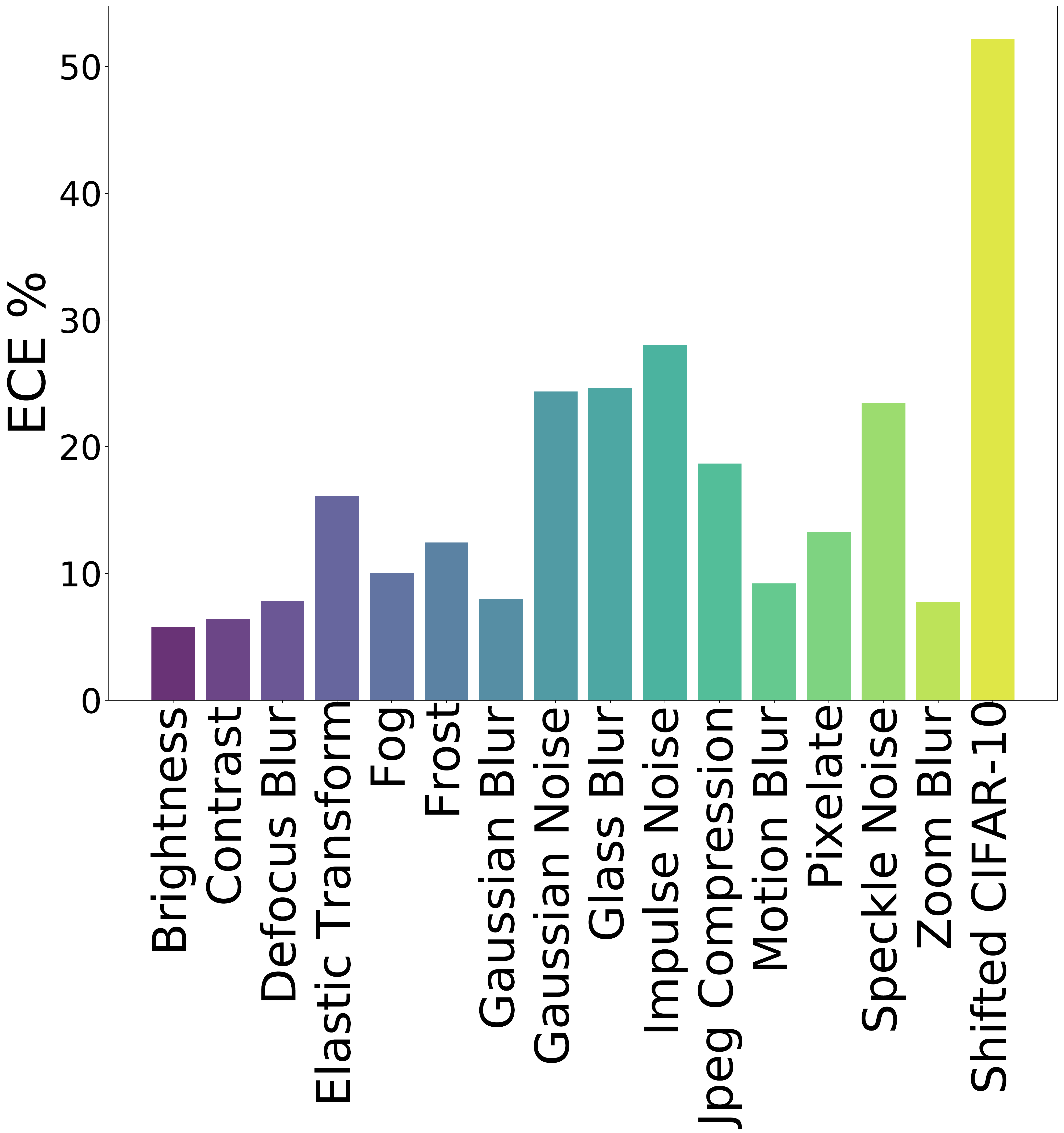}
        \caption{RN-110}
        \label{subfig:ece_cifar10c_resnet110}
    \end{subfigure}
    \begin{subfigure}{0.16\linewidth}
        \centering
        \includegraphics[width=\linewidth]{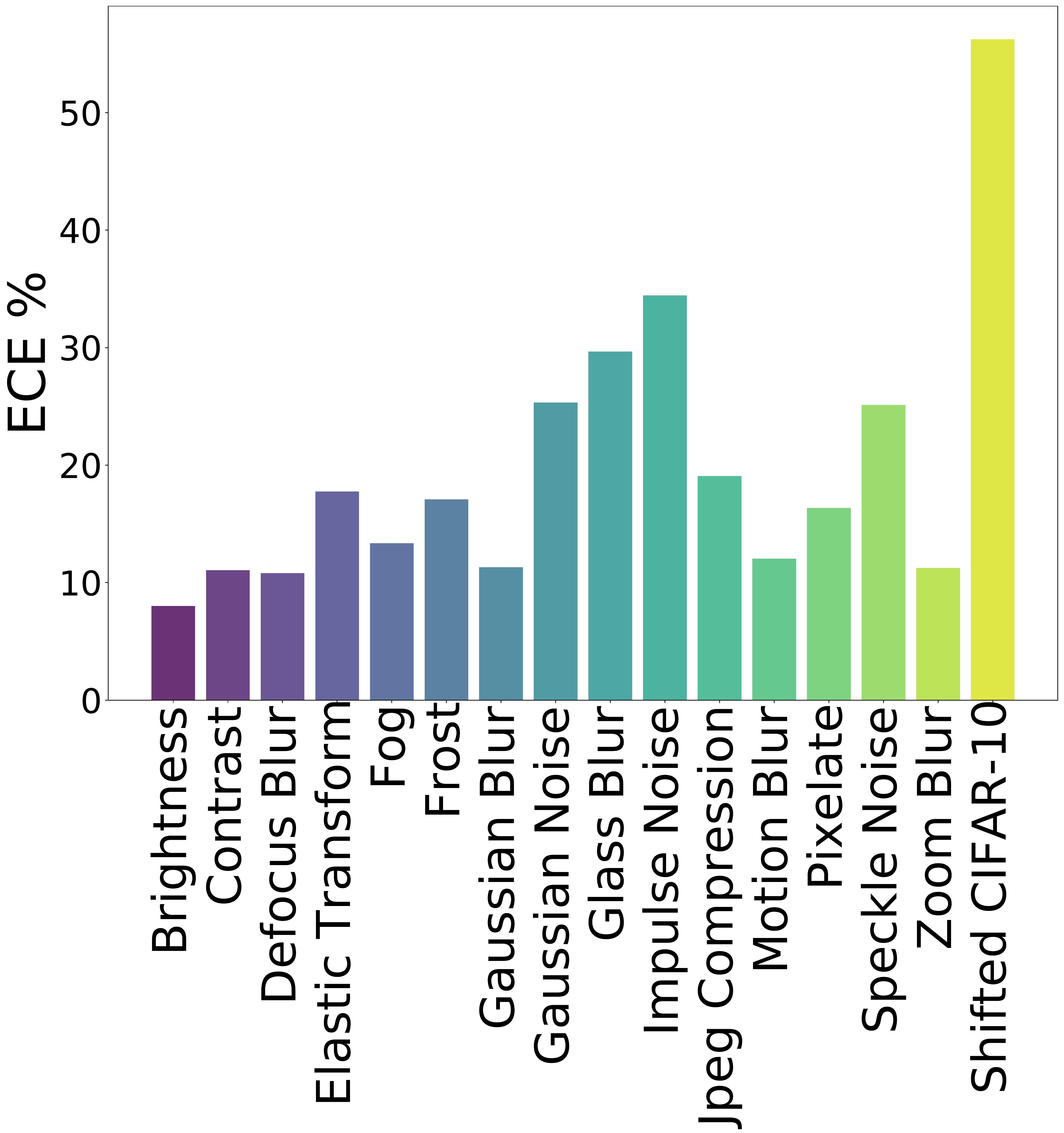}
        \caption{VGG-16}
        \label{subfig:ece_cifar10c_vgg16}
    \end{subfigure}
    \begin{subfigure}{0.16\linewidth}
        \centering
        \includegraphics[width=\linewidth]{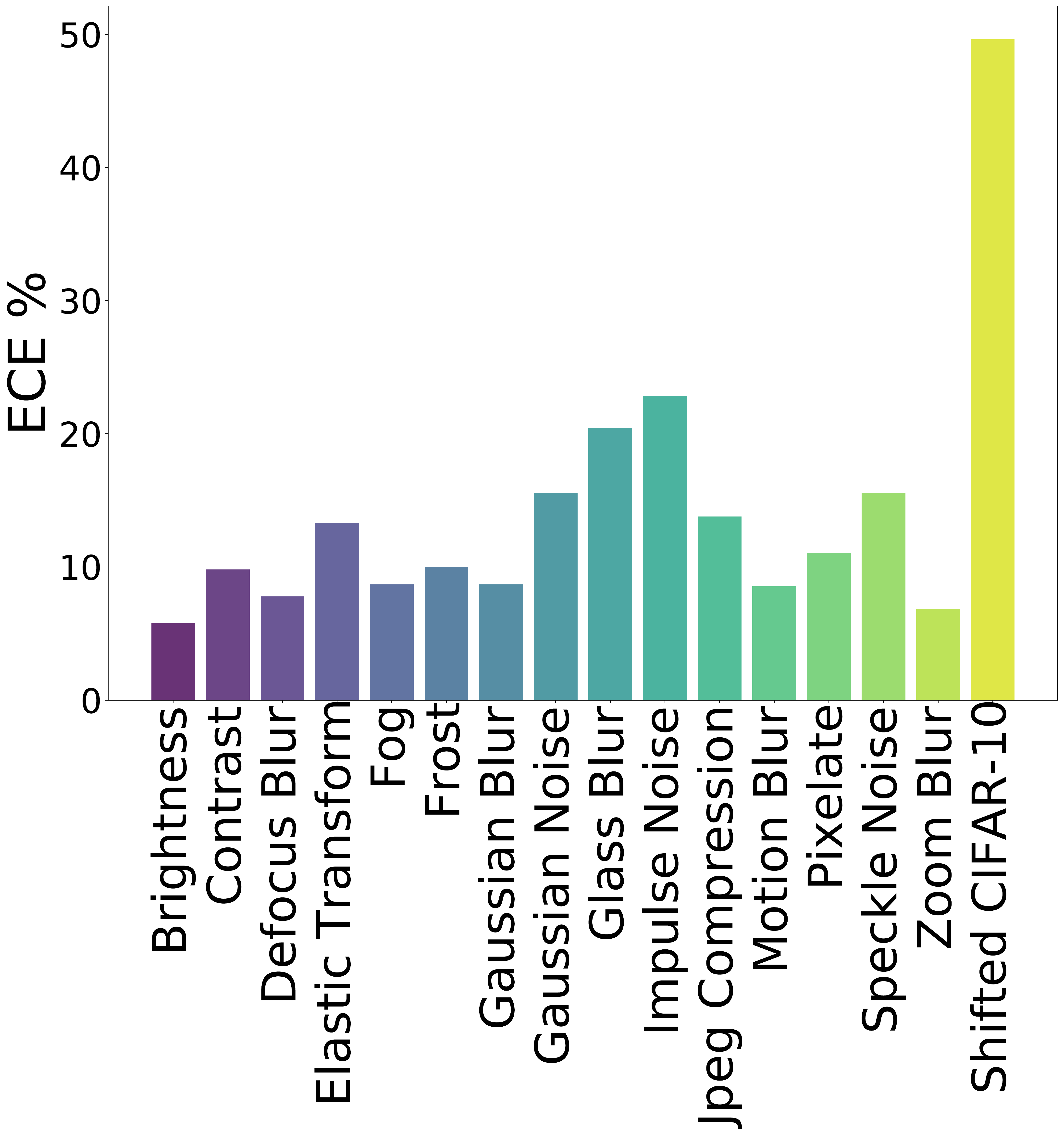}
        \caption{WRN}
        \label{subfig:ece_cifar10c_wide_resnet}
    \end{subfigure}
    \begin{subfigure}{0.16\linewidth}
        \centering
        \includegraphics[width=\linewidth]{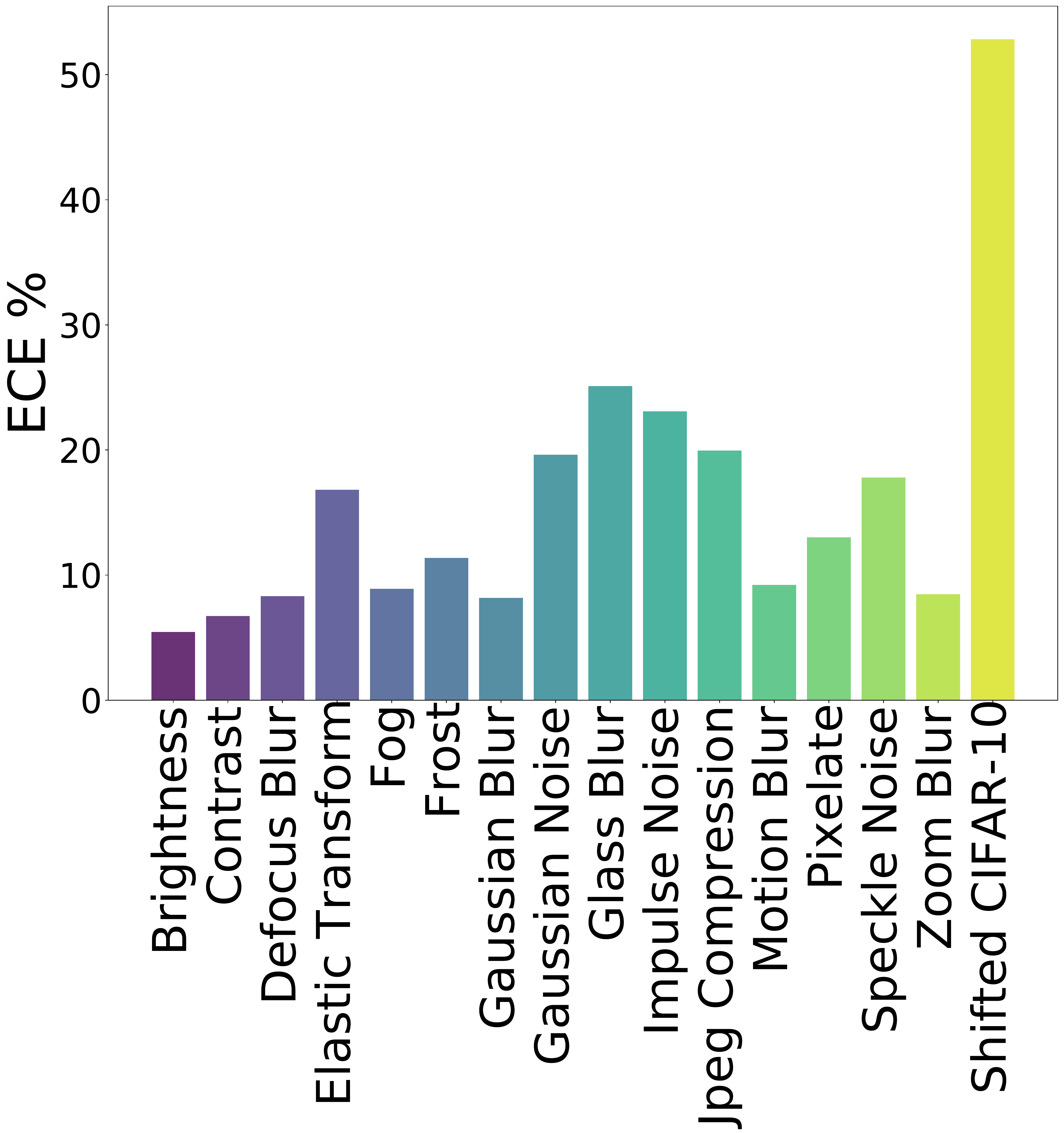}
        \caption{INC-v3}
        \label{subfig:ece_cifar10c_inception_v3}
    \end{subfigure}

    \begin{subfigure}{0.16\linewidth}
        \centering
        \includegraphics[width=\linewidth]{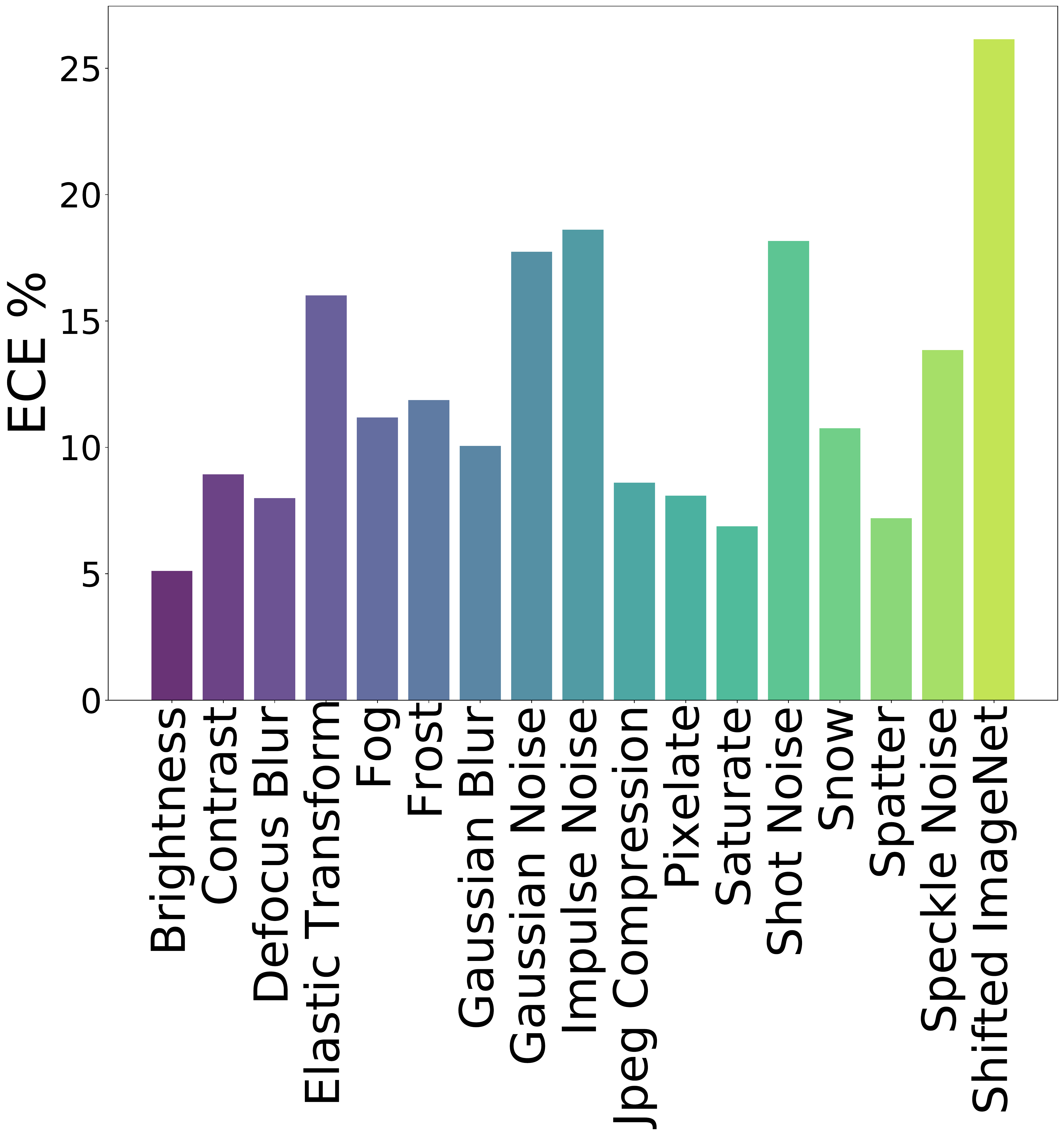}
        \caption{VB-16}
        \label{subfig:ece_imagenet_c_vb16}
    \end{subfigure}
    \begin{subfigure}{0.16\linewidth}
        \centering
        \includegraphics[width=\linewidth]{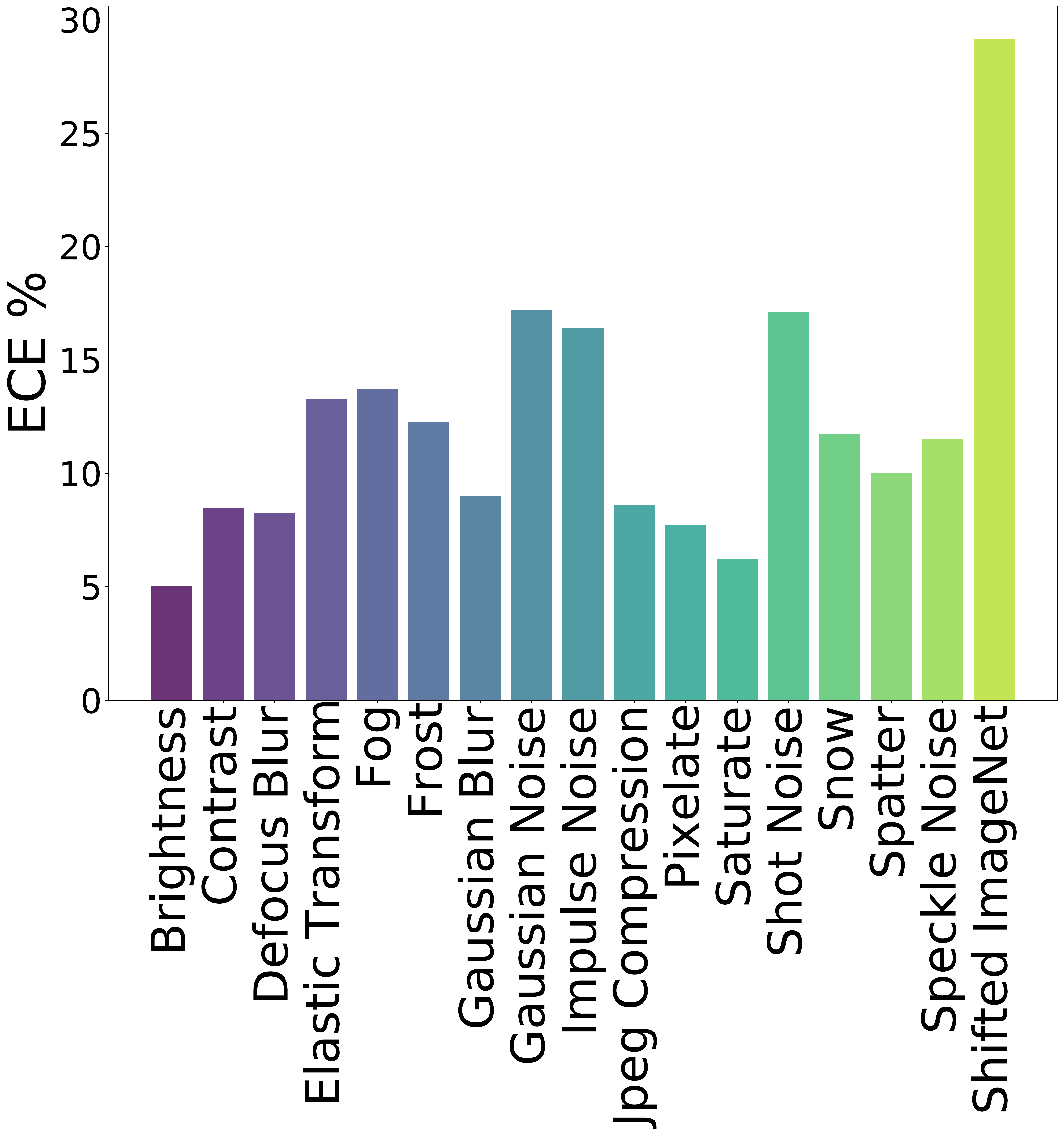}
        \caption{VB-32}
        \label{subfig:ece_imagenet_c_vb32}
    \end{subfigure}
    \begin{subfigure}{0.16\linewidth}
        \centering
        \includegraphics[width=\linewidth]{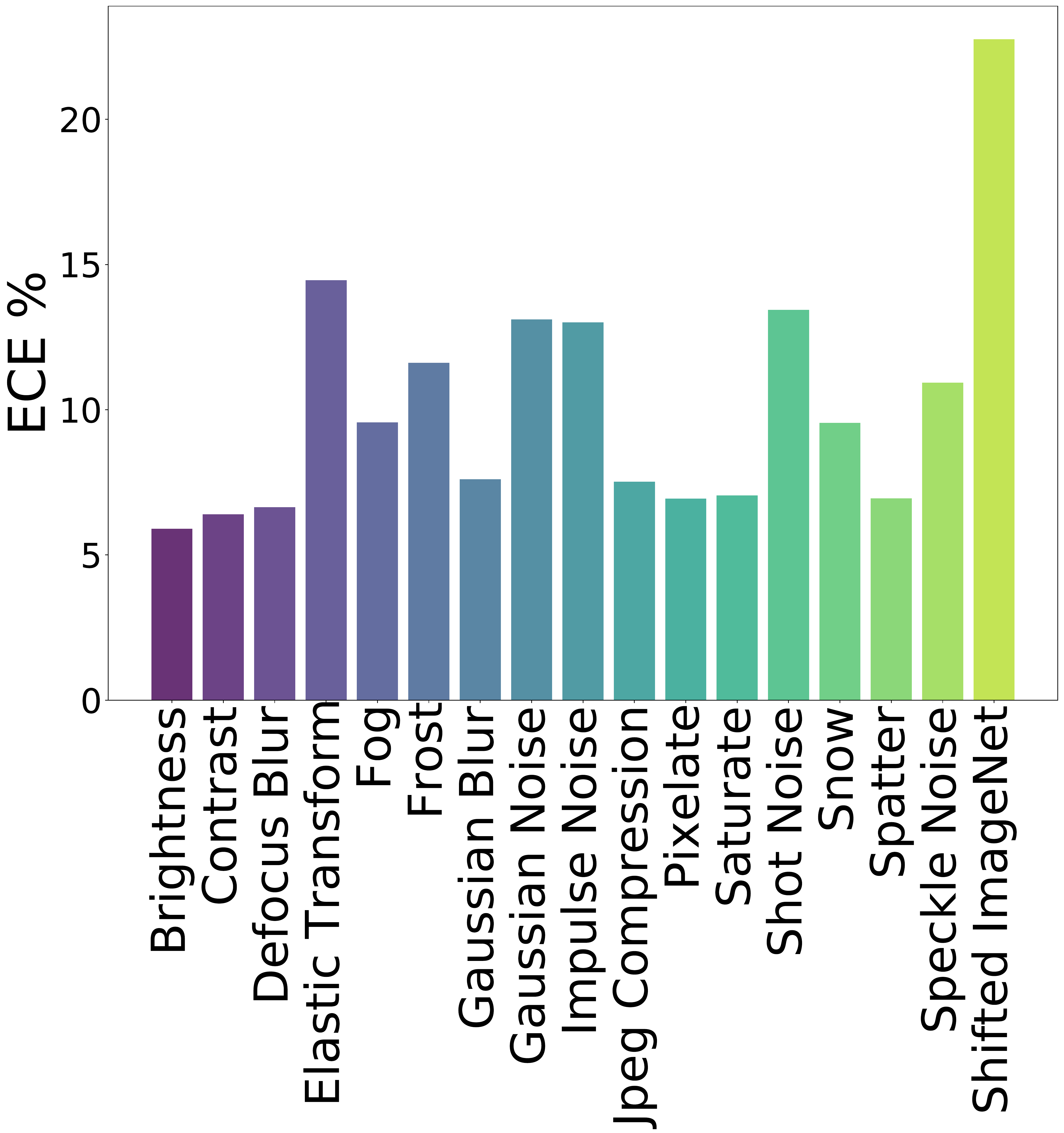}
        \caption{VL-16}
        \label{subfig:ece_imagenet_c_vl16}
    \end{subfigure}
    \begin{subfigure}{0.16\linewidth}
        \centering
        \includegraphics[width=\linewidth]{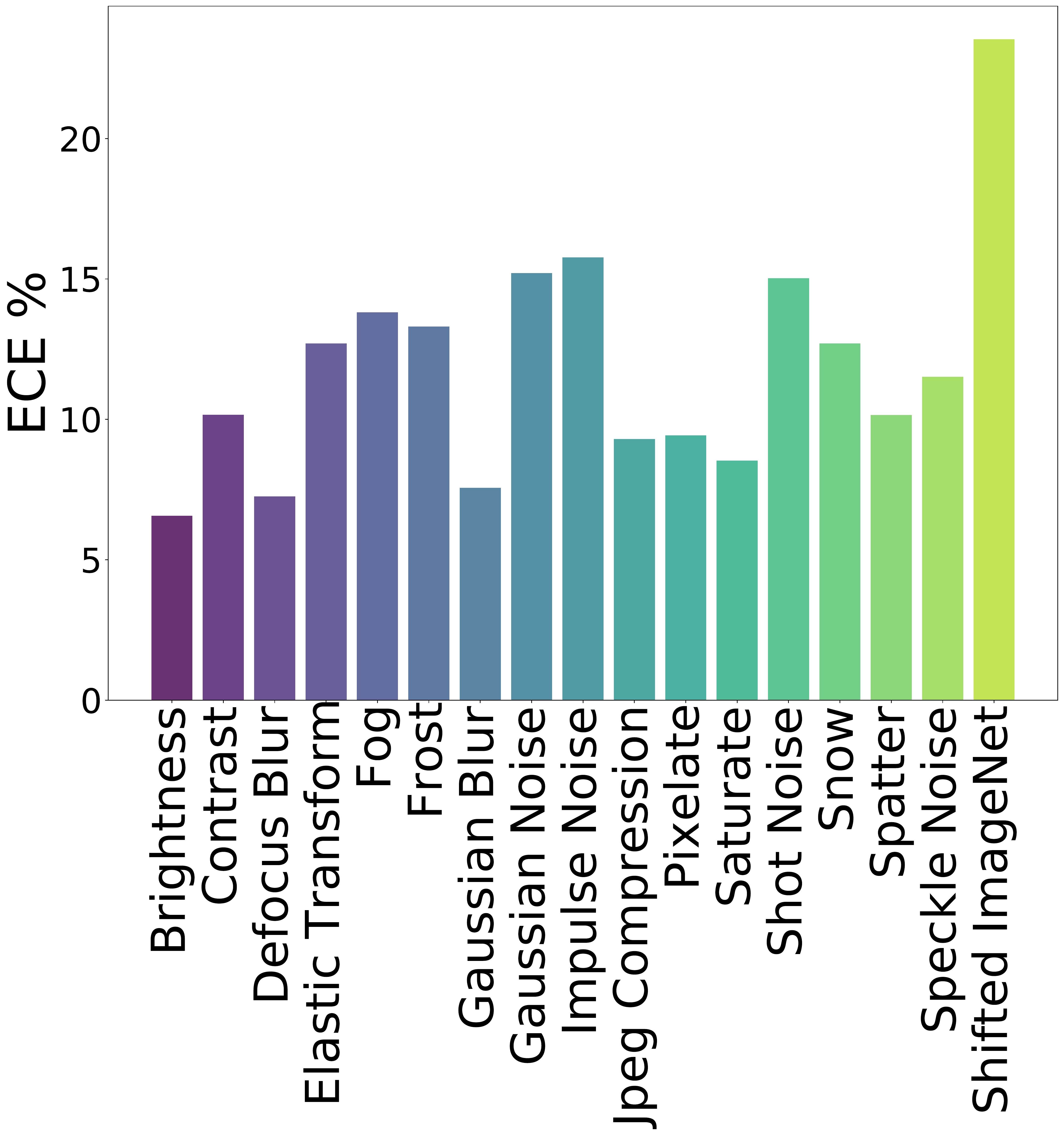}
        \caption{VL-32}
        \label{subfig:ece_imagenet_c_vl32}
    \end{subfigure}
    \caption{ECE\% of models on CIFAR-10-C (top row) and ImageNet-C (bottom row) for different corruption types.}
    \label{fig:ece_cifar10_c_imagenet_c_corruption_types}
\end{figure*}

\textbf{Evaluation of Shifted Datasets} We present the ECE\% of 6 architectures: DenseNet-121, ResNet-50/110, VGG-16, Wide-ResNet-28-10 and Inception-v3, all trained on CIFAR-10 on CIFAR-10-C where we use 15 different corruption types at the highest intensity (i.e., 5) and compare with the ECE\% of shifted CIFAR-10. Similarly, we present the ECE\% of 4 ViTs: ViT-B-16, ViT-B-32, ViT-L-16 and ViT-L-32, trained on ImageNet, evaluated on ImageNet-C. The results are presented in \cref{fig:ece_cifar10_c_imagenet_c_corruption_types}. Clearly, the ECE of shifted CIFAR-10 and shifted ImageNet is significantly higher than all corruption types.

In \cref{fig:lpips}, we present the LPIPS between images of the ImageNet val set with those corrupted using ImageNet-C corruptions and Shifted ImageNet, where the sample pairs have the same L2 difference in image space. We note that for Shifted ImageNet, the LPIPS is the highest compared to other corruption types. In \cref{fig:radar}, we present the radar plot of performance for all ViT models on ImageNet benchmarks. We note that the order of performance remains exactly the same for our benchmarks compared to real-world ImageNet benchmarks.

\section{Qualitative Examples of Generated Samples}
\label{app:qualitative}

\begin{figure*}[!t]
    \centering
    \begin{subfigure}{0.49\linewidth}
        \centering
        \includegraphics[width=\linewidth]{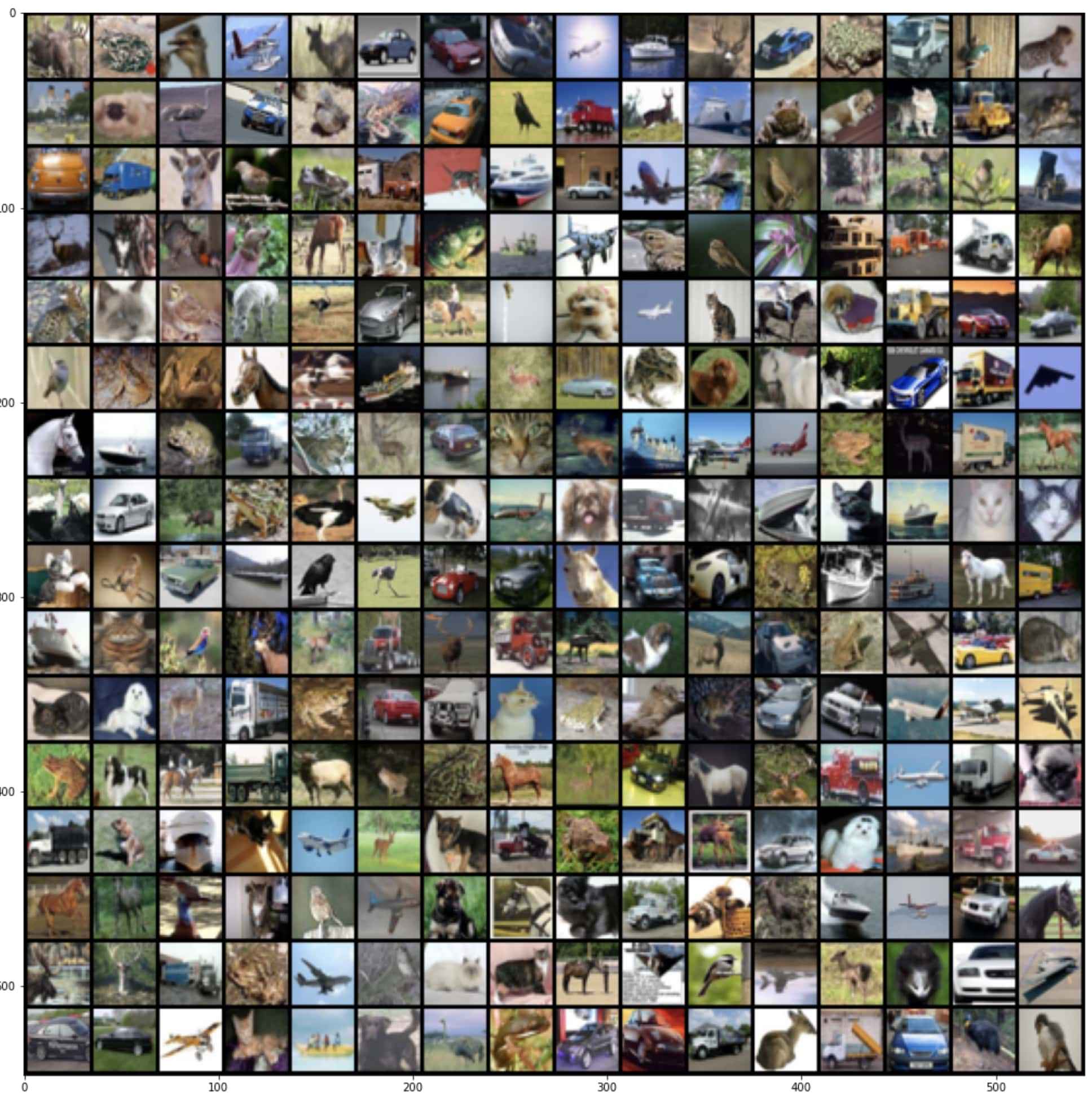}
        \caption{CIFAR-10 Real}
        \label{subfig:cifar10_real_app}
    \end{subfigure}
    \begin{subfigure}{0.49\linewidth}
        \centering
        \includegraphics[width=\linewidth]{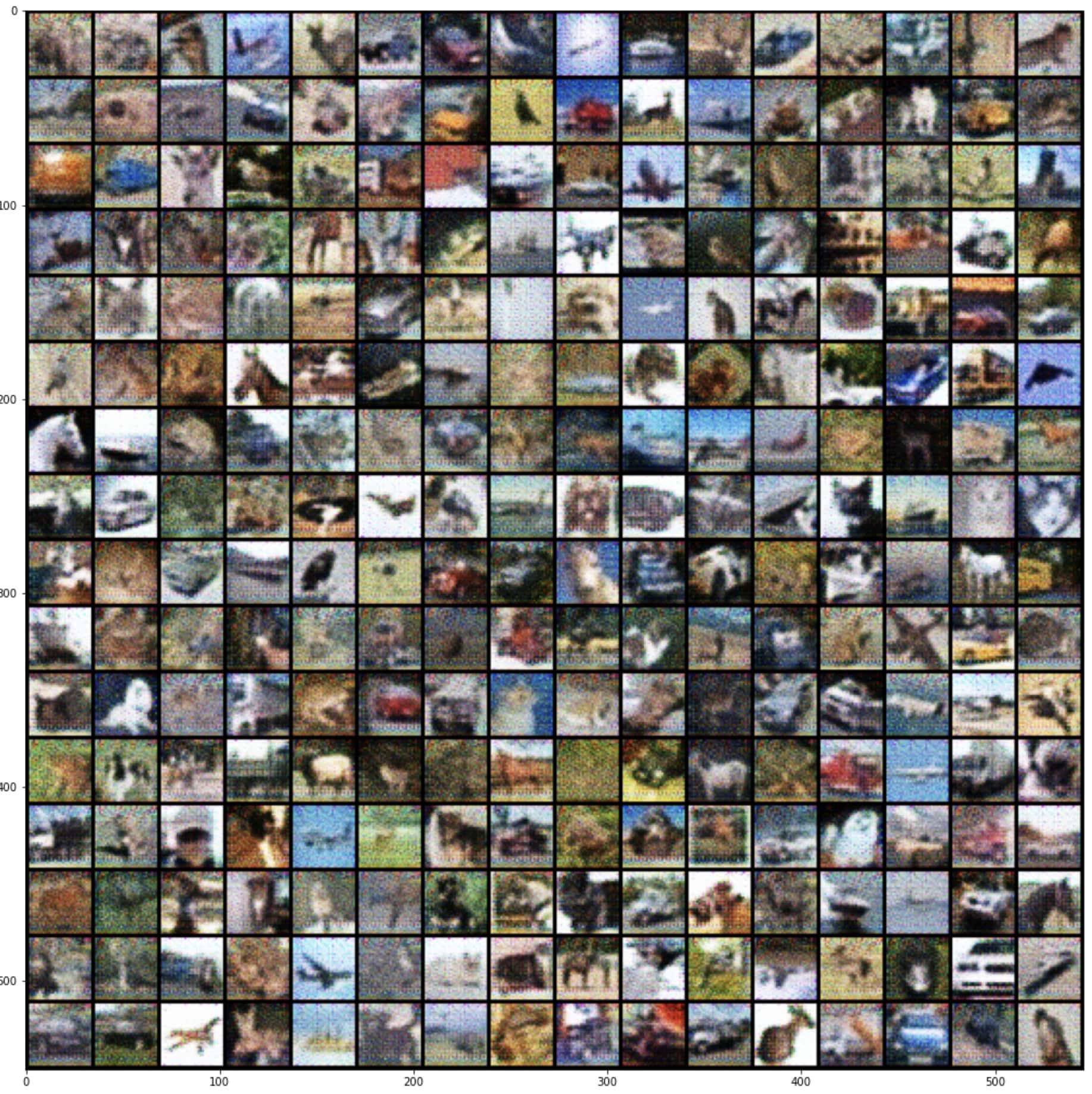}
        \caption{CIFAR-10 Shifted}
        \label{subfig:cifar10_shifted_app}
    \end{subfigure}

    \begin{subfigure}{0.49\linewidth}
        \centering
        \includegraphics[width=\linewidth]{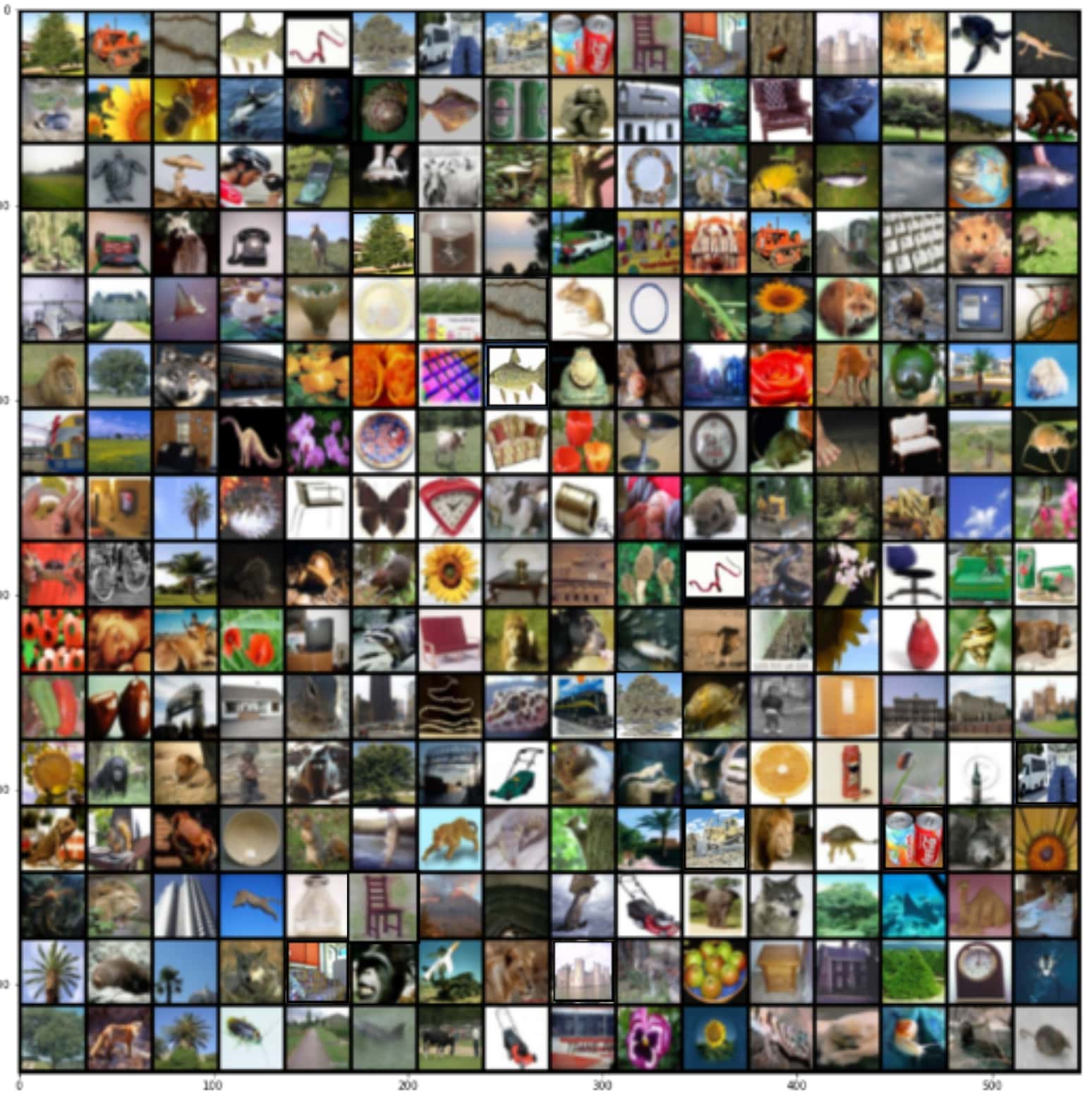}
        \caption{CIFAR-100 Real}
        \label{subfig:cifar100_real_app}
    \end{subfigure}
    \begin{subfigure}{0.49\linewidth}
        \centering
        \includegraphics[width=\linewidth]{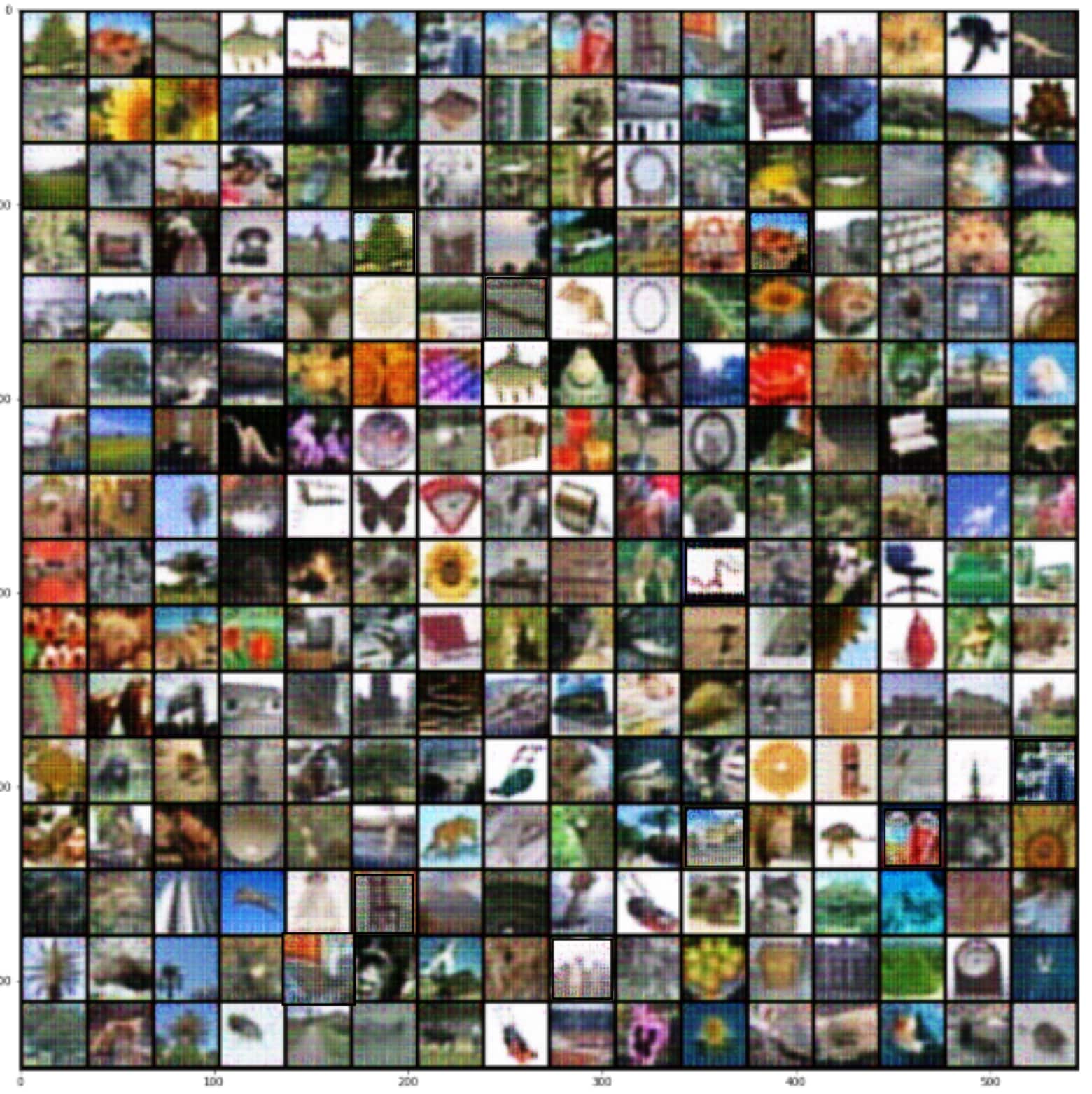}
        \caption{CIFAR-100 Shifted}
        \label{subfig:cifar100_shifted_app}
    \end{subfigure}

    \caption{Additional qualitative samples for CIFAR-10 and CIFAR-100 datasets. Left column shows real samples, and the right column shows corresponding shifted/transformed samples.}
    \label{fig:additional_qualitative_shifted}
\end{figure*}

\begin{figure*}[!t]
    \centering
    \begin{subfigure}{0.6\linewidth}
        \centering
        \includegraphics[width=\linewidth]{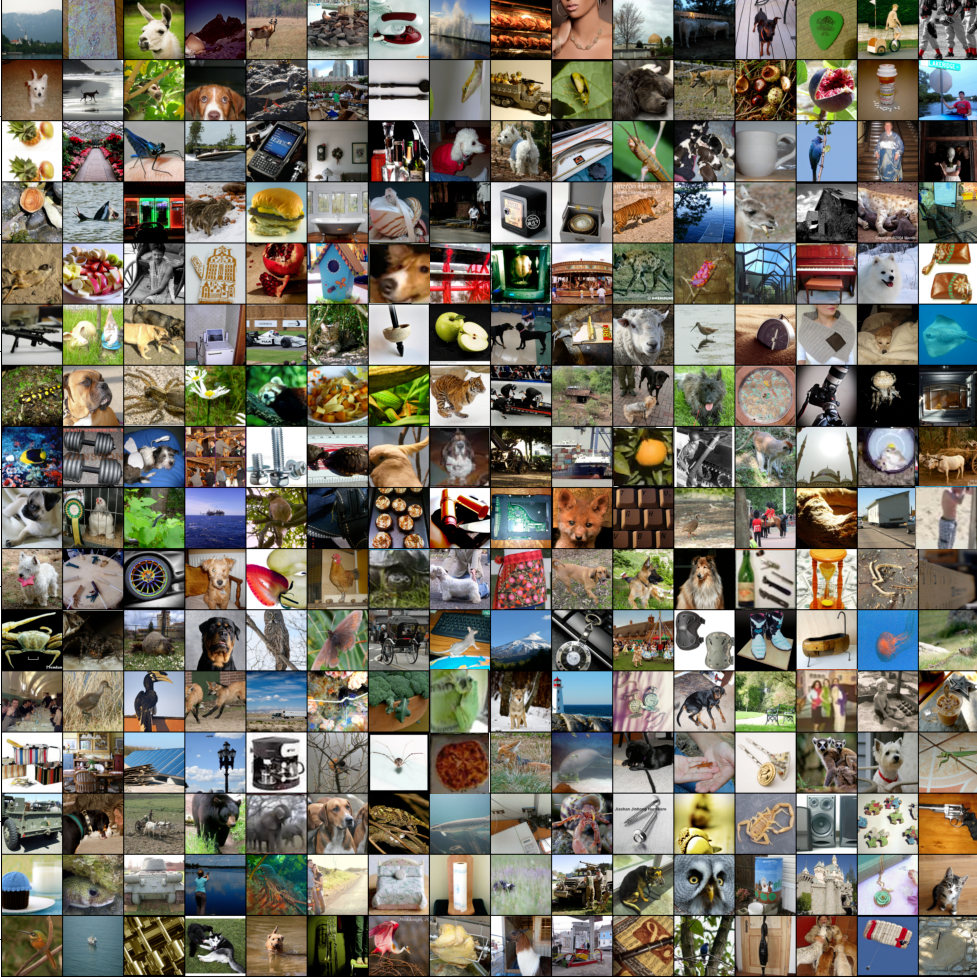}
        \caption{ImageNet Real}
        \label{subfig:imagenet_real_app}
    \end{subfigure}
    \begin{subfigure}{0.6\linewidth}
        \centering
        \includegraphics[width=\linewidth]{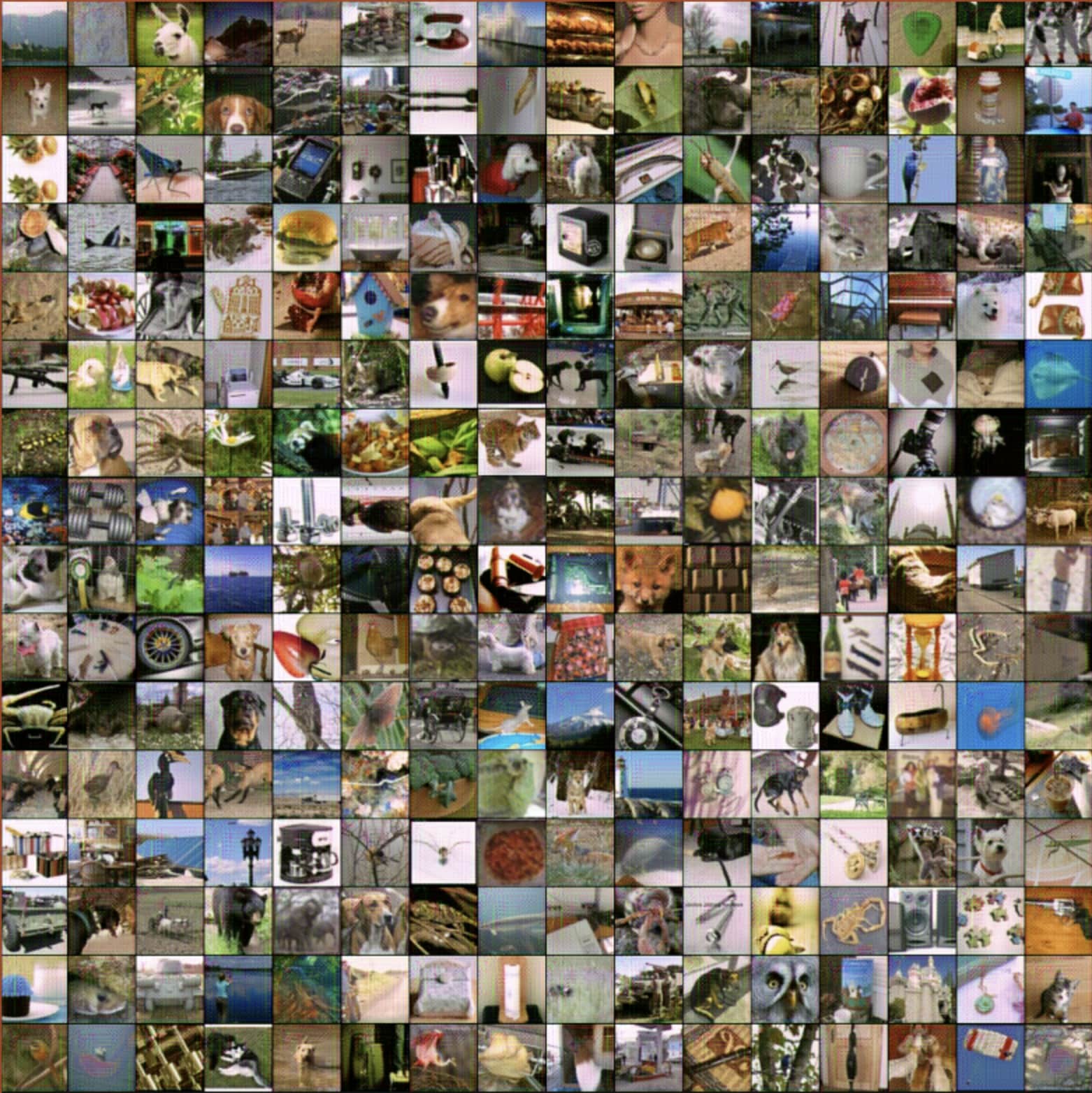}
        \caption{ImageNet Shifted}
        \label{subfig:imagenet_shifted_app}
    \end{subfigure}
    \caption{Additional qualitative samples for ImageNet datasets. Top shows real samples, and bottom shows corresponding shifted/transformed samples.}
    \label{fig:additional_qualitative_shifted_imagenet}
\end{figure*}

\begin{figure*}[!t]
    \centering
    \begin{subfigure}{0.49\linewidth}
        \centering
        \includegraphics[width=\linewidth]{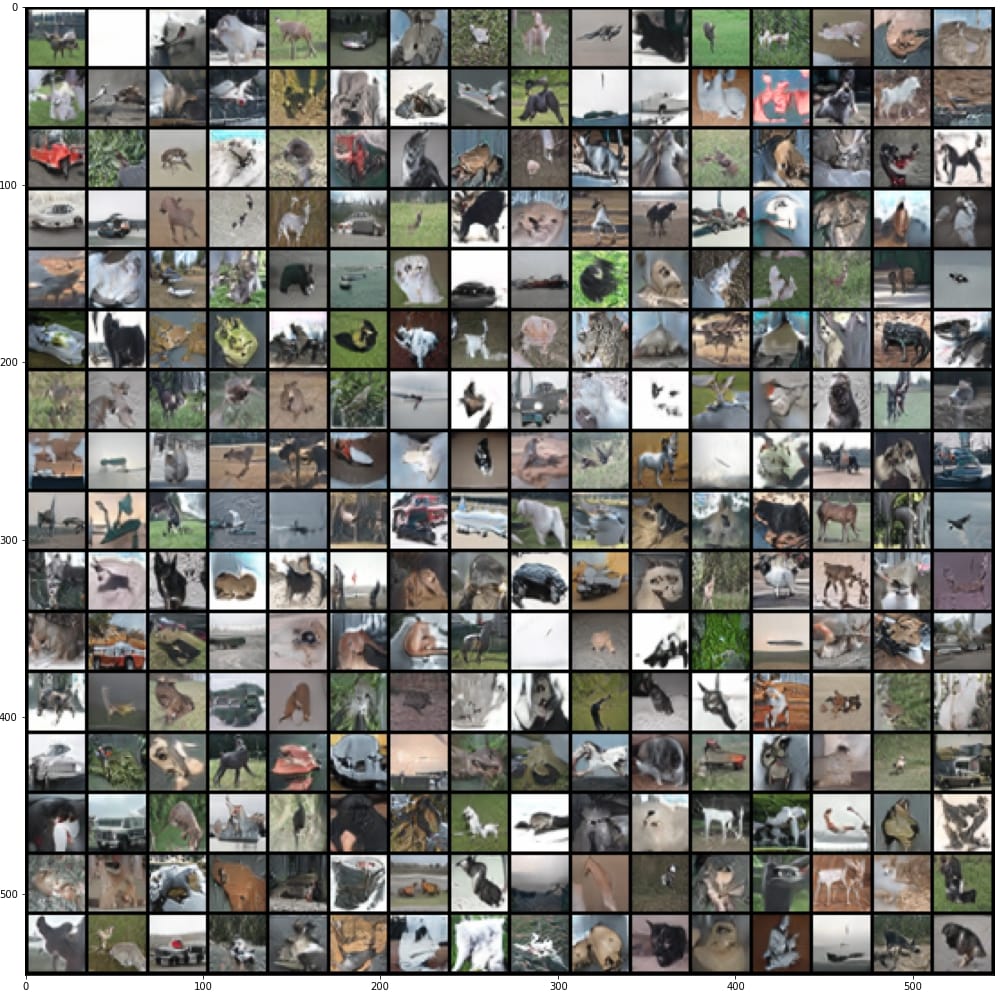}
        \caption{Near OoD CIFAR-10}
        \label{subfig:cifar10_morphed_app}
    \end{subfigure}
    \begin{subfigure}{0.49\linewidth}
        \centering
        \includegraphics[width=\linewidth]{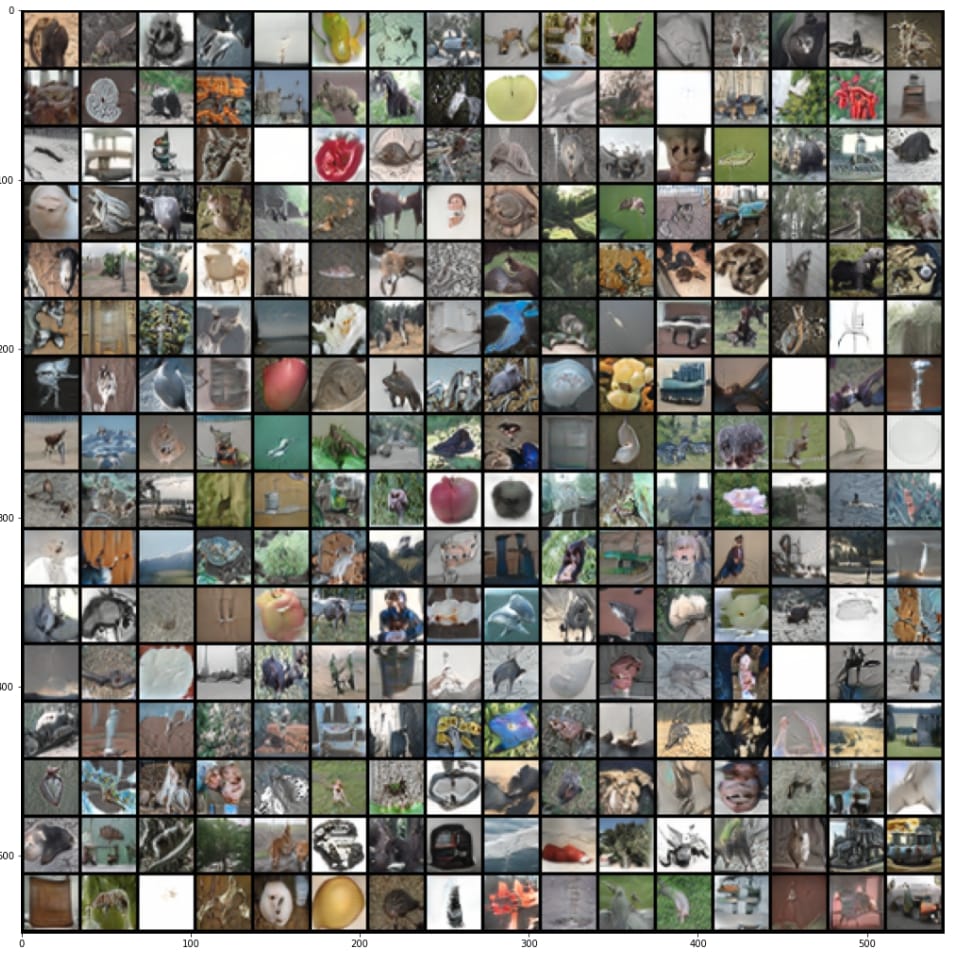}
        \caption{Near OoD CIFAR-100}
        \label{subfig:cifar100_morphed_app}
    \end{subfigure}
    \begin{subfigure}{0.75\linewidth}
        \centering
        \includegraphics[width=\linewidth]{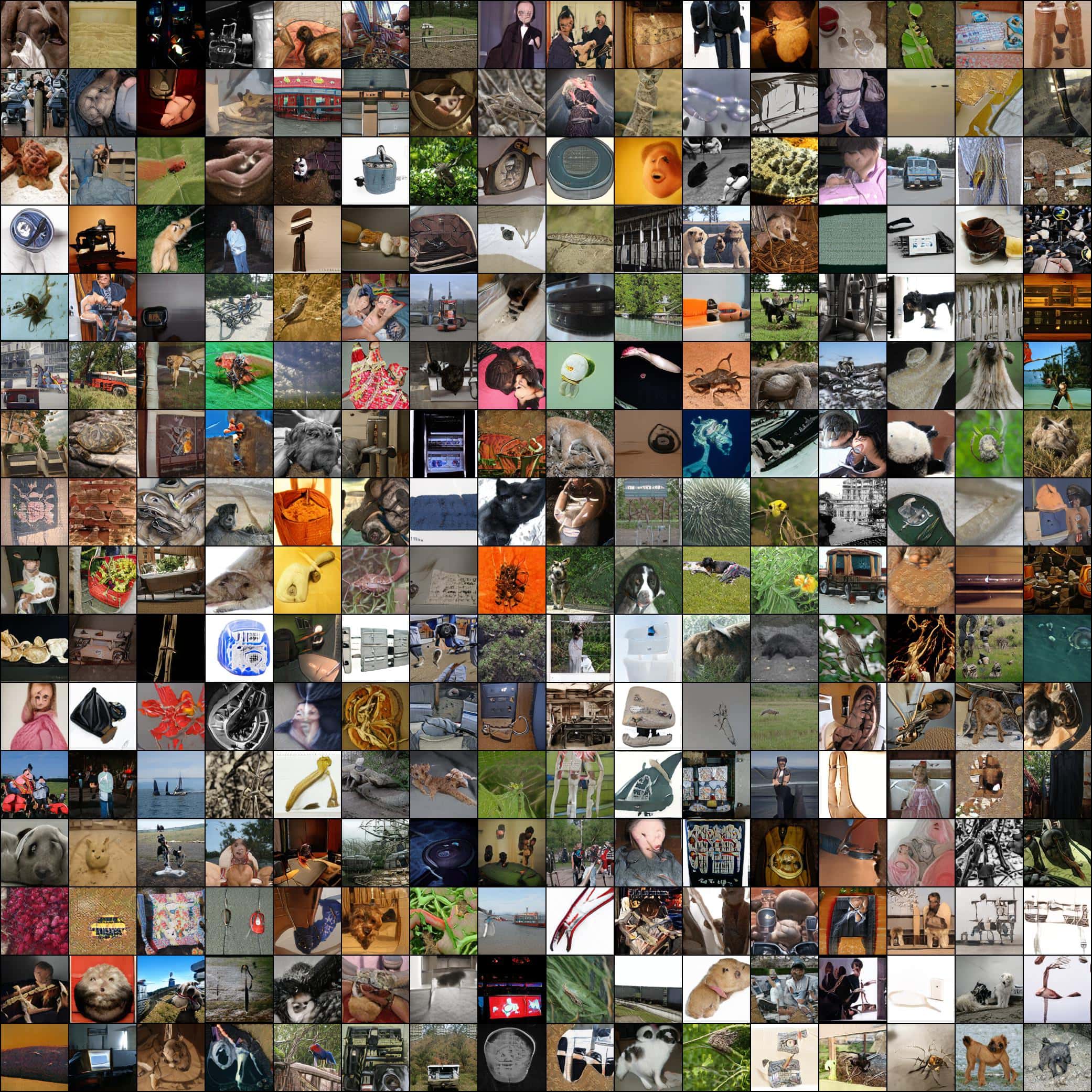}
        \caption{Near OoD ImageNet}
        \label{subfig:imagenet_morphed_app}
    \end{subfigure}
    \caption{Additional qualitative samples for CIFAR-10, CIFAR-100 and ImageNet datasets. Samples show Near OoD images.}
    \label{fig:additional_qualitative_morphed}
\end{figure*}

In \cref{fig:additional_qualitative_shifted}, \cref{fig:additional_qualitative_shifted_imagenet} and \cref{fig:additional_qualitative_morphed}, we present additional qualitative samples of shifted and near OoD examples respectively for both CIFAR-10, CIFAR-100 and ImageNet. In \cref{fig:additional_qualitative_shifted}, on the left column, we show real samples from CIFAR-10 and CIFAR-100. On the right column, we show corresponding shifted samples. Similar examples for ImageNet can be found in \cref{fig:additional_qualitative_shifted_imagenet}. Finally, in \cref{fig:additional_qualitative_morphed}, we show examples of near OoD samples for both CIFAR-10, CIFAR-100 and ImageNet.

\begin{table*}[!t]
\centering
\scriptsize
\begin{tabular}{ccccc}
\toprule
\textbf{Model} & \multicolumn{4}{c}{\textbf{AUPRC}} \\
& \textbf{Im-O (SE)} & \textbf{Im-O (SC)} & \textbf{N-Im (Ours) (SE)} & \textbf{N-Im(Ours) (SC)} \\
\midrule
ViT-B-16 & $73.23$ & $72.57$ & $67.10$ & $62.08$ \\
ViT-B-32 & $69.39$ & $67.48$ & $55.32$ & $52.24$ \\
ViT-L-16 & $80.46$ & $79.39$ & $70.17$ & $69.58$ \\
ViT-L-32 & $76.15$ & $75.32$ & $68.56$ & $66.93$ \\ 
\bottomrule
\end{tabular}
\caption{AUPRC of Vision-Transformer models trained on ImageNet using ImageNet-O \cite{hendrycks2021natural} (Im-O) and Near OoD ImageNet (N-Im) as OoD datasets and with softmax entropy (SE) and confidence (SC) as uncertainty/confidence.}
\label{table:auprc_imagenet}
\end{table*}

\begin{table*}[!t]
    \begin{minipage}{.49\linewidth}
      \centering
        \scriptsize
            \begin{tabular}{cccc}
            \toprule
            \textbf{Model} & \multicolumn{3}{c}{\textbf{AUROC}} \\
            & \textbf{SVHN} &  \textbf{CIFAR-100} & \textbf{N CIFAR-10} \\
            \midrule
            DenseNet-121 & $97.52$ & $91.42$ & $\mathbf{85.67}$ \\
            ResNet-50 & $96.24$ & $90.89$ & $\mathbf{84.66}$ \\
            ResNet-110 & $96.75$ & $91.3$ & $\mathbf{85.55}$ \\
            VGG-16 & $91.26$ & $89.16$ & $\mathbf{81.07}$ \\
            Wide-ResNet-28-10 & $96.59$ & $91.78$ & $\mathbf{86.54}$ \\
            Inception-v3 & $96.12$ & $91.31$ & $\mathbf{87.07}$ \\
            \bottomrule
            \end{tabular}
            \caption{AUROC of ensemble models trained on CIFAR-10 using predictive entropy on SVHN, CIFAR-100 and Near OoD CIFAR-10 (N CIFAR-10).}
            \label{table:auroc_ensemble_cifar10}
    \end{minipage} \hfill
    \begin{minipage}{.49\linewidth}
        \centering
        \scriptsize
            \begin{tabular}{cccc}
            \toprule
            \textbf{Model} & \multicolumn{3}{c}{\textbf{AUPRC}} \\
            & \textbf{SVHN} &  \textbf{CIFAR-10} & \textbf{N CIFAR-100} \\
            \midrule
            DenseNet-121 & $98.83$ & $90.72$ & $\mathbf{86.92}$ \\
            ResNet-50 & $97.87$ & $89.7$ & $\mathbf{85.83}$ \\
            ResNet-110 & $98.19$ & $90.35$ & $\mathbf{86.7}$ \\
            VGG-16 & $94.89$ & $87.97$ & $\mathbf{83.93}$ \\
            Wide-ResNet-28-10 & $98.06$ & $90.94$ & $\mathbf{88.03}$ \\
            Inception-v3 & $97.79$ & $90.14$ & $\mathbf{88.14}$ \\
            \bottomrule
            \end{tabular}
            \caption{AUPRC of ensemble models trained on CIFAR-10 using predictive entropy on SVHN, CIFAR-100 and Near OoD CIFAR-10 (N CIFAR-10).}
            \label{table:auprc_ensemble_cifar10}
    \end{minipage} 
\end{table*}

\begin{table*}[!t]
    \begin{minipage}{.49\linewidth}
      \centering
        \scriptsize
            \begin{tabular}{cccc}
            \toprule
            \textbf{Model} & \multicolumn{3}{c}{\textbf{AUROC}} \\
            & \textbf{SVHN} &  \textbf{CIFAR-10} & \textbf{N CIFAR-100} \\
            \midrule
            DenseNet-121 & $88.75$ & $82.92$ & $\mathbf{62.33}$ \\
            ResNet-50 & $82.66$ & $81.28$ & $\mathbf{57.18}$ \\
            ResNet-110 & $81.07$ & $82.55$ & $\mathbf{59.7}$ \\
            VGG-16 & $78.3$ & $78.87$ & $\mathbf{52.42}$ \\
            Wide-ResNet-28-10 & $83.62$ & $82.65$ & $\mathbf{62.83}$ \\
            Inception-v3 & $83.89$ & $83.3$ & $\mathbf{64.66}$ \\
            \bottomrule
            \end{tabular}
            \caption{AUROC of ensemble models trained on CIFAR-100 using predictive entropy on SVHN, CIFAR-10 and Near OoD CIFAR-100.}
            \label{table:auroc_ensemble_cifar100}
    \end{minipage} \hfill
    \begin{minipage}{.49\linewidth}
        \centering
        \scriptsize
            \begin{tabular}{cccc}
            \toprule
            \textbf{Model} & \multicolumn{3}{c}{AUPRC} \\
            & \textbf{SVHN} &  \textbf{CIFAR-10} & \textbf{N CIFAR-100} \\
            \midrule
            DenseNet-121 & $93.97$ & $78.96$ & $\mathbf{67.82}$ \\
            ResNet-50 & $90.24$ & $76.78$ & $\mathbf{64.85}$ \\
            ResNet-110 & $89.08$ & $78.67$ & $\mathbf{66.21}$ \\
            VGG-16 & $88.27$ & $74.51$ & $\mathbf{62.7}$ \\
            Wide-ResNet-28-10 & $91.27$ & $78.81$ & $\mathbf{67.94}$ \\
            Inception-v3 & $89.81$ & $79.14$ & $\mathbf{68.76}$ \\
            \bottomrule
            \end{tabular}
            \caption{AUPRC of ensemble models trained on CIFAR-100 using predictive entropy on SVHN, CIFAR-10 and Near OoD CIFAR-100.}
            \label{table:auprc_ensemble_cifar100}
    \end{minipage} 
\end{table*}

\begin{table*}[!t]
\centering
\scriptsize
\begin{tabular}{cccc}
\toprule
\textbf{Model} & \multicolumn{2}{c}{\textbf{CIFAR-10-C}} & \textbf{Shifted CIFAR-10 (Ours)} \\
& \textbf{\textit{Avg ECE \%}} & \textbf{\textit{Max ECE \%}} & \textbf{\textit{ECE \%}} \\
\midrule
DenseNet121 & $13.69\pm0.17$ & $25.86\pm0.40$ & $\mathbf{51.55\pm0.33}$ \\
ResNet-50 & $13.71\pm0.48$ & $25.76 \pm 1.09$ & $\mathbf{50.07 \pm 1.24}$ \\
ResNet-110 & $14.40\pm0.28$ & $28.03 \pm 0.55$ & $\mathbf{52.16\pm0.66}$ \\
VGG-16 & $17.51\pm0.22$ & $34.45 \pm 0.40$ & $\mathbf{56.25\pm0.41}$ \\
Wide-ResNet-28-10 & $11.92\pm0.13$ & $22.87 \pm 0.21$ & $\mathbf{49.64 \pm 0.43}$ \\
Inception-v3 & $13.47\pm0.37$ & $25.10 \pm 0.71$ & $\mathbf{52.84 \pm 0.19}$ \\
\bottomrule
\end{tabular}
\caption{ECE \% on CIFAR-10-C compared to Shifted CIFAR-10.}
\label{table:ece_cifar10_c_shifted_cifar10}
\end{table*}

\begin{table*}[!t]
\centering
\scriptsize
\resizebox{\linewidth}{!}
{
\begin{tabular}{cccccccccc}
\toprule
\textbf{Model} & \multicolumn{3}{c}{\textbf{AUROC SVHN}} & \multicolumn{3}{c}{\textbf{AUROC CIFAR-100}} & \multicolumn{3}{c}{\textbf{AUROC Near OoD CIFAR-10}} \\
\cmidrule{2-10}
& \textit{\textbf{Entropy}} & \textit{\textbf{Confidence}} & \textit{\textbf{Mahalanobis}} & \textit{\textbf{Entropy}} & \textit{\textbf{Confidence}} & \textit{\textbf{Mahalanobis}} & \textit{\textbf{Entropy}} & \textit{\textbf{Confidence}} & \textit{\textbf{Mahalanobis}} \\
\midrule
DenseNet-121 & $93.12\pm1.13$ & $92.85\pm1.11$ & $96.22\pm0.30$ & $87.23\pm0.21$ & $87.17\pm0.22$ & $89.71\pm0.14$ & $78.81 \pm 0.36$ & $79.11\pm0.34$ & $79.75\pm0.39$ \\
ResNet-50 & $92.39\pm0.30$ & $92.17\pm0.30$ & $92.67\pm1.35$ & $86.92\pm0.53$ & $86.78\pm0.50$ & $88.40\pm0.33$ & $78.92\pm0.75$ & $79.09\pm0.72$ & $79.04\pm0.57$ \\
ResNet-110 & $91.63\pm1.82$ & $91.41\pm1.81$ & $91.94\pm1.56$ & $87.48\pm0.09$ & $87.35\pm0.09$ & $87.91\pm0.2$ & $78.08\pm0.49$ & $80.20\pm0.48$ & $78.14\pm0.50$ \\
VGG-16 & $86.70\pm1.05$ & $86.78\pm1.00$ & $90.93\pm0.81$ & $83.37\pm0.22$ & $83.30\pm0.21$ & $85.94\pm0.35$ & $73.43\pm0.55$ & $73.61\pm0.53$ & $75.46\pm1.12$ \\
Wide-ResNet-28-10 & $90.98\pm1.14$ & $90.89\pm1.09$ & $98.72\pm0.11$ & $88.60\pm0.06$ & $88.48\pm0.06$ & $91.15\pm0.02$ & $80.56 \pm 0.47$ & $81.73\pm0.46$ & $81.78\pm0.11$ \\
Inception-v3 & $91.94\pm0.54$ & $91.77\pm0.53$ & $93.49\pm0.79$ & $86.54\pm0.43$ & $86.42\pm0.42$ & $89.56\pm0.28$ & $80.27\pm0.39$ & $80.41\pm0.38$ & $83.76\pm0.43$ \\
\midrule
ViT-B-16 & $99.65\pm0.01$ & $99.49\pm0.01$ & $96.67\pm0.18$ & $98.33\pm0.03$ & $98.19\pm0.03$ & $98.87\pm0.00$ & $87.00\pm0.04$ & $87.08\pm.04$ & $86.65\pm0.22$ \\
ViT-B-32 & $99.65\pm0.01$ & $99.44\pm0.02$ & $95.35\pm0.21$ & $98.10\pm0.03$ & $97.93\pm0.03$ & $98.67\pm0.01$ & $85.33\pm0.12$ & $85.44\pm.12$ & $86.21\pm0.23$ \\
ViT-L-16 & $99.76\pm0.02$ & $99.64\pm0.01$ & $97.66\pm.42$ & $98.70\pm0.02$ & $98.61\pm0.01$ & $99.17\pm0.01$ & $85.93\pm0.28$ & $86.15\pm0.27$ & $89.47\pm0.25$ \\
ViT-L-32 & $99.78\pm0.01$ & $99.63\pm0.02$ & $95.63\pm0.09$ & $98.45\pm.02$ & $98.29\pm.02$ & $98.80\pm.02$ & $85.25\pm0.2$ & $85.38\pm0.20$ & $84.83\pm0.11$ \\
\bottomrule
\end{tabular}
}
\caption{AUROC of models trained on CIFAR-10 using softmax entropy (Entropy), softmax confidence (Confidence) and Mahalanobis distance on SVHN, CIFAR-100 and Near OoD CIFAR-10. Near OoD samples are far harder to detect given their consistently low AUROC scores.}
\label{table:auroc_cifar10}
\end{table*}

\begin{table*}[!t]
\centering
\scriptsize
\resizebox{\linewidth}{!}
{
\begin{tabular}{cccccccccc}
\toprule
\textbf{Model} & \multicolumn{3}{c}{\textbf{AUPRC SVHN}} & \multicolumn{3}{c}{\textbf{AUPRC CIFAR-100}} & \multicolumn{3}{c}{\textbf{AUPRC Near OoD CIFAR-10}} \\
\cmidrule{2-10}
& \textit{\textbf{Entropy}} & \textit{\textbf{Confidence}} & \textit{\textbf{Mahalanobis}} & \textit{\textbf{Entropy}} & \textit{\textbf{Confidence}} & \textit{\textbf{Mahalanobis}} & \textit{\textbf{Entropy}} & \textit{\textbf{Confidence}} & \textit{\textbf{Mahalanobis}} \\
\midrule
DenseNet-121 & $96.78\pm0.38$& $82.89\pm5.11$& $94.4\pm0.41$& $86.84\pm0.11$& $84.7\pm0.66$& $89.75\pm0.15$& $80.38\pm0.14$& $63.6\pm0.52$& $68.72\pm0.75$ \\
ResNet-50 & $95.88\pm0.13$& $86.19\pm1.44$& $88.41\pm2.29$& $85.85\pm0.39$& $85.39\pm0.92$& $88.45\pm0.42$& $80.97\pm0.33$& $65.81\pm1.92$& $68.86\pm1.09$ \\
ResNet-110 & $95.58\pm0.95$& $85.35\pm3.25$& $86.25\pm2.29$& $86.29\pm0.14$& $86.27\pm0.23$& $87.54\pm0.29$& $80.8\pm0.28$& $67.73\pm1.26$& $64.41\pm1.2$ \\
Wide-ResNet-28-10 & $95.58\pm0.59$& $77.81\pm2.88$& $97.6\pm0.15$& $88.15\pm0.08$& $85.83\pm0.16$& $91.55\pm0.04$& $80.9\pm0.22$& $66.03\pm0.87$& $72.06\pm0.2$ \\
Inception-v3 & $95.7\pm0.31$& $83.32\pm1.8$& $90.91\pm0.86$& $85.86\pm0.41$& $83.14\pm0.67$& $90.13\pm0.28$& $81.47\pm0.22$& $63.45\pm0.66$& $78.11\pm0.34$ \\
\midrule
ViT-B-16 & $99.86\pm0.0$& $99.08\pm0.02$& $93.2\pm0.37$& $98.41\pm0.03$& $98.18\pm0.03$& $98.77\pm0.01$& $92.54\pm0.03$& $76.39\pm0.19$& $76.72\pm0.43$ \\
ViT-B-32 & $99.86\pm0.01$& $99.0\pm0.04$& $90.64\pm0.42$& $98.19\pm0.03$& $97.88\pm0.02$& $98.57\pm0.01$& $91.53\pm0.08$& $73.21\pm0.15$& $76.56\pm0.6$ \\
ViT-L-16 & $99.9\pm0.01$& $99.36\pm0.04$& $95.42\pm0.85$& $98.8\pm0.01$& $98.49\pm0.02$& $99.02\pm0.01$& $92.68\pm0.12$& $70.26\pm0.77$& $81.19\pm0.29$ \\
ViT-L-32 & $99.91\pm0.0$& $99.33\pm0.02$& $90.36\pm0.18$& $98.55\pm0.02$& $98.25\pm0.04$& $98.6\pm0.03$& $92.02\pm0.08$& $69.81\pm0.59$& $73.29\pm0.14$ \\
\bottomrule
\end{tabular}
}
\caption{AUPRC of models trained on CIFAR-10 using softmax entropy (Entropy), softmax confidence (Confidence) and Mahalanobis distance on SVHN, CIFAR-100 and Near OoD CIFAR-10. Near OoD samples are far harder to detect given their consistently low AUPRC scores.}
\label{table:auprc_cifar10}
\end{table*}

\clearpage
\begin{table*}[!t]
\centering
\scriptsize
\resizebox{\linewidth}{!}
{
\begin{tabular}{cccccccccc}
\toprule
\textbf{Model} & \multicolumn{3}{c}{\textbf{AUROC SVHN}} & \multicolumn{3}{c}{\textbf{AUROC CIFAR-100}} & \multicolumn{3}{c}{\textbf{AUROC Near OoD CIFAR-100}} \\
\cmidrule{2-10}
& \textit{\textbf{Entropy}} & \textit{\textbf{Confidence}} & \textit{\textbf{Mahalanobis}} & \textit{\textbf{Entropy}} & \textit{\textbf{Confidence}} & \textit{\textbf{Mahalanobis}} & \textit{\textbf{Entropy}} & \textit{\textbf{Confidence}} & \textit{\textbf{Mahalanobis}} \\
\midrule
DenseNet-121 & $84.52\pm1.55$& $83.13\pm1.44$& $89.49\pm0.55$& $79.73\pm0.25$& $79.12\pm0.24$& $77.26\pm0.47$& $60.32\pm0.27$& $60.5\pm0.26$& $64.16\pm0.28$ \\
ResNet-50 & $79.77\pm0.69$& $78.82\pm0.71$& $86.41\pm0.11$& $78.82\pm0.08$& $78.26\pm0.09$& $82.68\pm0.18$& $56.58\pm0.26$& $56.67\pm0.26$& $58.48\pm0.67$ \\
ResNet-110 & $77.84\pm1.56$& $77.26\pm1.4$& $86.62\pm0.23$& $79.92\pm0.17$& $79.3\pm0.15$& $82.9\pm0.23$& $58.6\pm0.56$& $58.58\pm0.46$& $59.73\pm0.59$ \\
VGG-16 & $76.33\pm1.12$& $75.38\pm0.97$& $78.01\pm1.24$& $74.02\pm0.14$& $73.62\pm0.13$& $74.99\pm0.13$& $51.06\pm0.14$& $51.53\pm0.15$& $56.41\pm0.42$ \\
Wide-ResNet-28-10 & $81.85\pm0.79$& $80.71\pm0.7$& $84.18\pm1.01$& $80.82\pm0.11$& $80.41\pm0.12$& $73.42\pm0.14$& $62.19\pm0.17$& $62.05\pm0.14$& $62.38\pm0.1$ \\
Inception-v3 & $81.6\pm1.64$& $80.95\pm1.46$& $81.8\pm0.57$& $81.24\pm0.18$& $80.89\pm0.18$& $79.87\pm0.22$& $63.96\pm0.85$& $63.39\pm0.78$& $60.53\pm0.97$ \\
\midrule
ViT-B-16 & $93.31\pm0.21$& $91.92\pm0.19$& $95.91\pm0.03$& $93.29\pm0.04$& $92.35\pm0.05$& $93.95\pm0.03$& $79.47\pm0.06$& $79.04\pm0.06$& $82.91\pm0.07$ \\
ViT-B-32 & $92.98\pm0.13$& $91.56\pm0.11$& $93.78\pm0.21$& $91.97\pm0.2$& $90.94\pm0.21$& $92.22\pm0.19$& $75.36\pm0.16$& $75.05\pm0.15$& $78.97\pm0.24$ \\
ViT-L-16 & $95.11\pm0.16$& $94.29\pm0.15$& $97.6\pm0.04$& $94.62\pm0.08$& $94.04\pm0.09$& $95.31\pm0.09$& $80.36\pm0.08$& $80.23\pm0.1$& $84.72\pm0.21$ \\
ViT-L-32 & $94.01\pm0.07$& $92.62\pm0.06$& $96.01\pm0.12$& $94.09\pm0.07$& $93.28\pm0.06$& $94.15\pm0.06$& $76.87\pm0.12$& $76.64\pm0.12$& $81.29\pm0.14$ \\
\bottomrule
\end{tabular}
}
\caption{AUROC of models trained on CIFAR-100 using softmax entropy (Entropy), softmax confidence (Confidence) and Mahalanobis distance on SVHN, CIFAR-10 and Near OoD CIFAR-100. Near OoD samples are far harder to detect given their consistently low AUROC scores.}
\label{table:auroc_cifar100}
\end{table*}

\begin{table*}[!t]
\centering
\scriptsize
\resizebox{\linewidth}{!}
{
\begin{tabular}{cccccccccc}
\toprule
\textbf{Model} & \multicolumn{3}{c}{\textbf{AUPRC SVHN}} & \multicolumn{3}{c}{\textbf{AUPRC CIFAR-100}} & \multicolumn{3}{c}{\textbf{AUPRC Near OoD CIFAR-100}} \\
\cmidrule{2-10}
& \textit{\textbf{Entropy}} & \textit{\textbf{Confidence}} & \textit{\textbf{Mahalanobis}} & \textit{\textbf{Entropy}} & \textit{\textbf{Confidence}} & \textit{\textbf{Mahalanobis}} & \textit{\textbf{Entropy}} & \textit{\textbf{Confidence}} & \textit{\textbf{Mahalanobis}} \\
\midrule
DenseNet-121 & $91.84\pm0.93$& $72.78\pm2.29$& $82.85\pm0.73$& $75.89\pm0.3$& $80.67\pm0.38$& $80.31\pm0.26$& $69.17\pm0.13$& $50.01\pm0.6$& $55.53\pm0.3$ \\
ResNet-50 & $88.69\pm0.28$& $67.45\pm1.23$& $81.58\pm0.51$& $74.4\pm0.14$& $80.26\pm0.18$& $85.78\pm0.45$& $66.07\pm0.15$& $50.27\pm0.6$& $50.33\pm0.91$ \\
ResNet-110 & $87.29\pm1.01$& $62.34\pm2.97$& $81.9\pm0.78$& $75.94\pm0.17$& $81.25\pm0.23$& $86.1\pm0.55$& $67.37\pm0.35$& $48.07\pm1.02$& $51.08\pm0.99$ \\
VGG-16 & $87.04\pm0.78$& $60.21\pm1.61$& $66.7\pm1.54$& $70.35\pm0.18$& $73.07\pm0.23$& $77.02\pm0.25$& $64.3\pm0.07$& $50.72\pm0.17$& $50.32\pm0.39$ \\
Wide-ResNet-28-10 & $90.27\pm0.55$& $69.98\pm1.23$& $72.42\pm0.92$& $76.76\pm0.17$& $82.41\pm0.11$& $76.36\pm0.15$& $68.98\pm0.09$& $55.56\pm0.19$& $54.03\pm0.13$ \\
Inception-v3 & $88.54\pm1.19$& $72.75\pm2.25$& $66.81\pm2.55$& $76.89\pm0.3$& $82.99\pm0.1$& $79.31\pm0.35$& $69.98\pm0.34$& $59.6\pm1.21$& $50.77\pm1.07$ \\
\midrule
ViT-B-16 & $97.24\pm0.08$& $83.16\pm0.85$& $93.22\pm0.06$& $93.73\pm0.03$& $92.68\pm0.06$& $94.4\pm0.02$& $87.15\pm0.02$& $66.33\pm0.15$& $74.84\pm0.18$ \\
ViT-B-32 & $97.1\pm0.06$& $82.0\pm0.69$& $90.46\pm0.28$& $92.79\pm0.18$& $90.93\pm0.21$& $92.66\pm0.25$& $84.23\pm0.15$& $60.24\pm0.12$& $69.31\pm0.4$ \\
ViT-L-16 & $98.08\pm0.06$& $84.33\pm0.89$& $94.92\pm0.09$& $95.11\pm0.06$& $93.89\pm0.12$& $95.43\pm0.08$& $88.25\pm0.05$& $63.66\pm0.21$& $76.63\pm0.33$ \\
ViT-L-32 & $97.58\pm0.01$& $83.2\pm0.41$& $93.4\pm0.23$& $94.65\pm0.06$& $93.2\pm0.08$& $94.66\pm0.07$& $85.8\pm0.06$& $58.91\pm0.24$& $72.46\pm0.21$ \\
\bottomrule
\end{tabular}
}
\caption{AUPRC of models trained on CIFAR-100 using softmax entropy (Entropy), softmax confidence (Confidence) and Mahalanobis distance on SVHN, CIFAR-10 and Near OoD CIFAR-100. Near OoD samples are far harder to detect given their consistently low AUPRC scores.}
\label{table:auprc_cifar100}
\end{table*}

\end{document}